\documentclass{article}

\usepackage[preprint]{neurips_2026}

\usepackage[british]{babel}

\usepackage[utf8]{inputenc} %
\usepackage[T1]{fontenc}    %
\usepackage{hyperref}       %
\usepackage{url}            %
\usepackage{booktabs}       %
\usepackage{amsmath}        %
\usepackage{amsfonts}       %
\usepackage{nicefrac}       %
\usepackage{microtype}      %
\usepackage{xcolor}         %
\usepackage{graphicx}       %
\graphicspath{{plots-sep-17/}}   %
\usepackage{subcaption}     %
\usepackage{natbib}         %
\usepackage{multirow}
\usepackage{placeins}        %

\title{Beyond Accuracy: Robustness, Interpretability and Expressiveness of EEG Foundation Models}

\author{%
Urban Širca \\
Vrije Universiteit Amsterdam \\
\texttt{} \\
\And
Maryam Alimardani \\
Vrije Universiteit Amsterdam \\
\AND
Stefanos Zafeiriou \\
Imperial College London \\
\And
Konstantinos Barmpas \\
Imperial College London \\
}

\begin{document}

\maketitle

\begin{abstract}
EEG foundation models (EEG-FMs) have been evaluated predominantly on clean, in-distribution accuracy, demonstrating modest gains over supervised baselines and weak frozen representations. This study examines whether these conclusions hold beyond clean accuracy by evaluating six EEG-FMs and a supervised baseline across ten datasets along three layers of analysis: \emph{(i)} Robustness: we apply test-time perturbations including additive noise, random and region-based channel dropout and region-specific noise injection. Our analyses show that no single model dominates all failure modes. The most noise-robust model is among the most fragile under channel dropout and much of the dropout fragility disappears when channels are removed rather than zero-padded. \emph{(ii)} Interpretability: using attribution methods in EEG-FMs, we show that models broadly concentrate relevance on task-appropriate brain regions consistent with known neurophysiology. \emph{(iii)} Expressiveness: we demonstrate that the poor head-only performance previously attributed to low-quality pre-trained representations is largely explained by the pooling strategy and that EEG-FMs possess sufficient representational capacity when their token-level embeddings are preserved. Furthermore, with block-wise probing and attention analysis we show that late blocks are repurposed during fine-tuning, while early blocks already hold task-related information. Our results show that conclusions about EEG-FMs depend on evaluation choices and we recommend  that future evaluation of EEG-FMs should report robustness per perturbation type, produce attribution maps and examine multiple pooling strategies.
\end{abstract}

\section{Introduction}

Brain-Computer Interfaces (BCIs) translate neural activity into control signals for communication, rehabilitation and clinical monitoring \citep{mcfarland2011cacm_bci}. Electroencephalography (EEG) is the most practical neuroimaging modality for BCIs: it is non-invasive, portable, inexpensive and offers millisecond temporal resolution \citep{nicolasalonso2012bci_review}. However, EEG signals have a low signal-to-noise ratio due to muscular and environmental artefacts and vary across subjects, sessions and recording devices \citep{saha2020variabilityreview}.

To address these challenges, various methods and techniques have been deployed over the years for EEG decoding. Traditional BCI systems relied on hand-crafted features and machine learning algorithms to perform EEG classification \citep{pfurtscheller2001motorimagery, Classical, SignalProcessingMethods1, SignalProcessingMethods2, SignalProcessingMethods3, SignalProcessingMethods4}. The advent of deep learning reduced the reliance on manual feature selection \citep{lecun2015deeplearning, schirrmeister2017deepconvnet, lawhern2018eegnet, EEGInception, Barmpas_Scattering, EEGConformer}, with neural networks learning spatial and temporal representations end-to-end from raw or lightly preprocessed EEG data. Although these deep models achieve competitive performance across diverse BCI paradigms, they require large labelled datasets (a persistent bottleneck in EEG research) and often lose performance when applied across cohorts or recording conditions \citep{lotte2018bciupdate}. Inspired by the success of foundation models in natural language processing and computer vision \citep{brown2020fewshot, touvron2023llama}, researchers have begun pre-training large models on EEG data with the goal of learning general-purpose representations. These EEG foundation models (EEG-FMs) pre-train on large unlabelled corpora via self-supervision, aiming to generalise across users, tasks and conditions with minimal labelled fine-tuning data. 

Although several EEG-FMs (e.g.,~\citealp{jiang2024lbm_iclr, wang2024cbramod, ouahidi2025reve, xiao2025brainomni}) have been proposed that claim competitive downstream task performance, multiple independent studies converge on the same finding: \textit{EEG-FM downstream task performance gains over supervised baselines are modest.} 
Under subject-independent cross-validation, \citet{lee2025capabilities} found that EEG-FMs achieve marginal advantage over supervised baselines such as EEGNet, a finding later corroborated through systematic fine-tuning \citep{lee2025finetuninginsights} and few-shot calibration \citep{sirca2026peft}. Another large-scale study by \citet{liu2025eegfmbenchmark}, covering twelve models, reached a similar conclusion: \textit{specialist architectures trained from scratch sometimes outperformed EEG-FMs. Furthermore, linear probing of pre-trained models underperformed full fine-tuning across all models}.

These benchmarks, thorough as they are, share a common limitation: they evaluate EEG-FMs on clean, in-distribution data only. In practice, EEG recordings are inherently noisy due to electrode impedance drifts, gel dry-out, subject movements, and channel failures, leading to substantial variability across sessions and users~\citep{saha2020variabilityreview}. Consequently, models that perform well on curated laboratory data may significantly degrade under these conditions, limiting their value in clinical or real-world BCI settings~\citep{zhou2023robustxaisurvey, kuruppu2026fmreview}. Several works \citep{barmpas2024causalbci, dynaminc_conv} formalise this concern from a causal perspective, showing that data distribution shifts, arising from inter-subject variability but also sensor noise or electrode displacement, are among the main factors limiting generalisation in BCI systems. Yet, such conditions remain largely unexamined in current EEG-FM benchmarks. Among other concerns, \citet{kuruppu2026fmreview} highlights three gaps relevant to our study: \emph{(i)} robustness of EEG-FMs under simulated signal corruption has not been demonstrated, \emph{(ii)} interpretability remains unaddressed despite being necessary in high-risk domains such as healthcare and \emph{(iii)} head-only probing of pre-trained EEG-FMs consistently underperforms fine-tuning across all reviewed models, raising questions about the quality of pre-trained representations. 

The goal of this work is to test whether the conclusions currently drawn about EEG-FMs hold once evaluation extends beyond clean accuracy. We adopted the benchmarking protocol of \citet{lee2025capabilities} and extended the model set with state-of-the-art EEG-FMs: REVE \citep{ouahidi2025reve}, BrainOmni \citep{xiao2025brainomni}, NeuroRVQ \citep{barmpas2025neurorvq}, LaBraM \citep{jiang2024lbm_iclr}, CBraMod \citep{wang2024cbramod} and BIOT \citep{yang2023biot}, and EEGNet \citep{lawhern2018eegnet} as the baseline. We also added six benchmark datasets to the original set, totalling ten datasets spanning motor imagery, motor execution, event-related potentials, working memory, sleep staging, eyes-open versus eyes-closed classification and seizure detection. We conducted three layers of analysis to investigate the gaps highlighted by \citet{kuruppu2026fmreview}: \emph{(i)}~\textbf{Robustness:} using synthetic perturbations, we examined the performance of fine-tuned EEG-FMs to isolate distinct failure modes. \emph{(ii)}~\textbf{Interpretability:} using attribution maps as a plausibility test, we examined whether relevance falls on task-appropriate regions consistent with known neurophysiology. \emph{(iii)}~\textbf{Expressiveness:} using pooling ablations on frozen and fine-tuned backbones, we examined whether weak head-only performance reflects poor representation quality or the pooling strategy, and used block-wise probing on two best-performing models to examine where task-related information resides. Our contributions can be summarised as follows:

\begin{enumerate}
    \item \textbf{Robustness is perturbation-specific}. No single model is most robust across all tested perturbations. For example, the most noise-robust model is the most fragile under zero-padded channel dropout while all models (that accept variable input channels) are less fragile when channels are removed rather than zero-padded, pointing to the handling of zero-valued inputs rather than loss of task information.
    \item \textbf{Attribution maps pass a spatial plausibility test}. Attribution maps are broadly consistent with expected topographies, with model-dependent differences. This is necessary but not sufficient evidence of trustworthy decoding.
    \item \textbf{Poor head-only performance is substantially a pooling artefact}. Replacing mean pooling with token flattening improves head-only performance for all tested models. A parameter-matched control shows the gain is not explained by head size. Therefore, frozen EEG-FM representations retain more information than previous weak linear probing results suggested, and block-wise probing shows that early blocks already carry task information while fine-tuning mainly reshapes the later blocks.
\end{enumerate}

Based on this analysis, we recommend that future evaluation of EEG-FMs should report robustness per perturbation type (stating the exact testing configurations, such as zero-padded channels or removed), produce attribution maps for spatial plausibility evaluation and examine multiple pooling strategies before drawing conclusions about the representation quality of EEG-FMs.

\section{Related Work}
While clean accuracy has been extensively benchmarked for EEG-FMs, robustness, interpretability and expressiveness remain underexplored. This section reviews related work along these axes.

\subsection{Robustness and Expressiveness of EEG Foundation Models}

EEG signals are inherently noisy due to electrode artefacts and channel failures~\citep{saha2020variabilityreview}. While additive noise and channel dropout have been used as augmentation techniques during training~\citep{cheng2020subjectawarecontrastivelearningbiosignals}, to our knowledge no study has systematically evaluated existing EEG-FMs under test-time signal corruption. 
Separately, linear probing consistently underperforms full fine-tuning across all reviewed EEG-FMs, typically interpreted as evidence of low-quality representations~\citep{liu2025eegfmbenchmark, kuruppu2026fmreview, xiong2026eegfmbench}. However, this interpretation conflates representation quality with the pooling strategy used to aggregate tokens before the probe. The effect of pooling on frozen representations has been examined in other domains~\citep{rauch2025unmutepatchtokensrethinking} but not in EEG-FMs, where classifier heads have only been compared under full fine-tuning~\citep{xiong2026eegfmbench}. 

\subsection{Interpretability of EEG Foundation Models}
\label{related_work:interp}

Interpreting EEG-FMs requires faithful attribution methods. We review the main approaches used in EEG classification~\citep{zhou2023robustxaisurvey} and EEG-FMs~\citep{ma2026eegxplain, xiong2026eegfmbench, attnLRPcleverhans}, and the rules needed to extend them to transformer backbones. Gradient$\times$Input (G$\times$I)~\citep{deeplift, ancona2018inputxgradient} computes attribution by element-wise multiplying each input feature with its gradient of the output. Gradient-weighted Class Activation Mapping (GradCAM)~\citep{selvaraju2017gradcam} produces a coarse attribution map by computing a weighted gradient feature map to the last convolution layer in a CNN model, highlighting which spatial regions most influence the prediction. Layer-wise Relevance Propagation (LRP)~\citep{bach2015lrp} decomposes a model's output into per-input relevance scores by propagating relevance backward through every layer under conservation rules. Both LRP and G$\times$I pass standard sanity checks on EEG~\citep{ravindran2023eegxai_groundtruth}. GradCAM works primarily on CNNs and standard LRP rules are not directly applicable to the bilinear query-key interaction inside self-attention modules. For transformers, attention rollout~\citep{abnar2020rollout, chefer2021attnexplain} aggregates attention across layers but does not provide a faithful decomposition of the prediction and suffers from limited resolution. AttnLRP~\citep{achtibat2024attnlrp} resolves this by extending LRP with propagation rules for the operations present in self-attention, preserving the conservation property across the transformer. Beyond attribution methods, sparse autoencoders (SAEs) have been used to extract clinically grounded concepts from EEG-FM model embeddings~\citep{lehnschiøler2026mechanisticinterpretabilityeegfoundation}.

\section{Methods}
\label{sec:methods}

In this work, we benchmarked six EEG foundation models and one supervised baseline across ten datasets. Each model was tested on clean and perturbed data to assess performance and robustness, and with various attributions examined if learned representations align with expected brain regions.

\paragraph{Models.} The six EEG-FM models had a backbone parameter count spanning four orders of magnitude larger than the baseline model: BIOT \citep{yang2023biot} (3.19M), CBraMod \citep{wang2024cbramod} (4.88M), LaBraM \citep{jiang2024lbm_iclr} (5.82M), NeuroRVQ \citep{barmpas2025neurorvq} (5.87M), BrainOmni \citep{xiao2025brainomni} (37.76M, of which a 5.05M VQ tokeniser stayed frozen during fine-tuning) and REVE \citep{ouahidi2025reve} (69.19M). EEGNet \citep{lawhern2018eegnet} (3.4K parameters) served as the supervised baseline trained from scratch on each task. We evaluated every foundation model under two regimes: head-only probing on a frozen backbone and full fine-tuning of all parameters. The six foundation models were initialised from publicly available pre-trained checkpoints. Full architectural details, head layer specifications, and parameter breakdowns are provided in Appendix~\ref{app:models}. All models were trained following the protocol described in Appendix~\ref{app:training}, along with compute costs.

\paragraph{Datasets and Tasks.} We evaluated the models on ten datasets: a 4-class executed movement dataset \citep{schirrmeister2017deepconvnet} (\textit{Movement-HG}), two tasks from the Korean University OpenBMI \citep{lee2019openbmi} covering 2-class motor imagery (\textit{Motor-Imagery-KU}) and visual P300 event-related potentials (\textit{ERP}), three tasks from the PhysioNet EEG Motor dataset \citep{schalk2004bci2000, goldberger2000physionet} covering 4-class motor imagery (\textit{Motor-Imagery-PN}), 4-class motor execution (\textit{Movement-PN}), and eyes open versus closed (\textit{Eyes}), a digit span working memory task \citep{pavlov2022digitspan_dataset} (\textit{Memory}), a 6-class sleep staging dataset \citep{kemp2000sleepedf} (\textit{Sleep-EDF}), a 5-class clinical sleep staging dataset \citep{sleep2022hmc} (\textit{Sleep-HMC}), and a seizure detection task \citep{siena2020} (\textit{Seizure}). All recordings are common-average referenced and bandpass filtered following each model's preprocessing. Dataset and preprocessing details are listed in Appendix~\ref{app:datasets}.

\paragraph{Robustness to Perturbations.} All perturbations were applied at test time only to models trained on clean data, so any performance drop reflects the fragility of learned representations. We tested three settings (full details and parameters are given in Appendix~\ref{app:perturbations}): \emph{(i)} \textit{Additive noise.} Independent white and pink ($1/f$) noise was added to every channel at signal-to-noise ratios $\{10, 5, 0, -3, -5\}$\,dB, with an additional $-15$\,dB level for white noise. \emph{(ii)} \textit{Random channel dropout.} Channels were independently selected for zero-padding with per-channel probabilities $p \in \{0.1, 0.3, 0.5\}$. For the four models that accept a variable channel set (LaBraM, NeuroRVQ, REVE and BrainOmni), we also removed the dropped channels instead of zeroing them. \emph{(iii)} \textit{Region-based perturbations.} For each task, we defined a primary electrode region expected to carry the task signal and a control region with no expected contribution (assignments in Appendix~\ref{app:perturbations}). We tested both zero-padding and noise injection (with 5\,dB and $-3$\,dB) for each region. 

\paragraph{Interpretability Analyses.} 

To understand why models are robust or fragile and whether their learned features align with known neurophysiology, we adopted \textit{AttnLRP} as our primary attribution method (as motivated in Section~\ref{related_work:interp}) and used G$\times$I as an alternative for BrainOmni and NeuroRVQ where AttnLRP is numerically unstable (detailed in Appendix \ref{app:interpretability}). We supplement AttnLRP with raw \textit{attention maps} and \textit{linear probes} at each transformer block to locate where task-relevant information emerges, focusing on REVE and NeuroRVQ as the two best performing models in Section \ref{sec:clean}. 
Full implementation details and model-specific adaptations are described in Appendix~\ref{app:interpretability}. We show attribution maps for five benchmarks: Movement, Motor-Imagery, ERP, Eyes and Seizure.

\section{Results}

\subsection{Clean Performance}
\label{sec:clean}

\begin{figure}[!h]
\centering
\includegraphics[width=\textwidth]{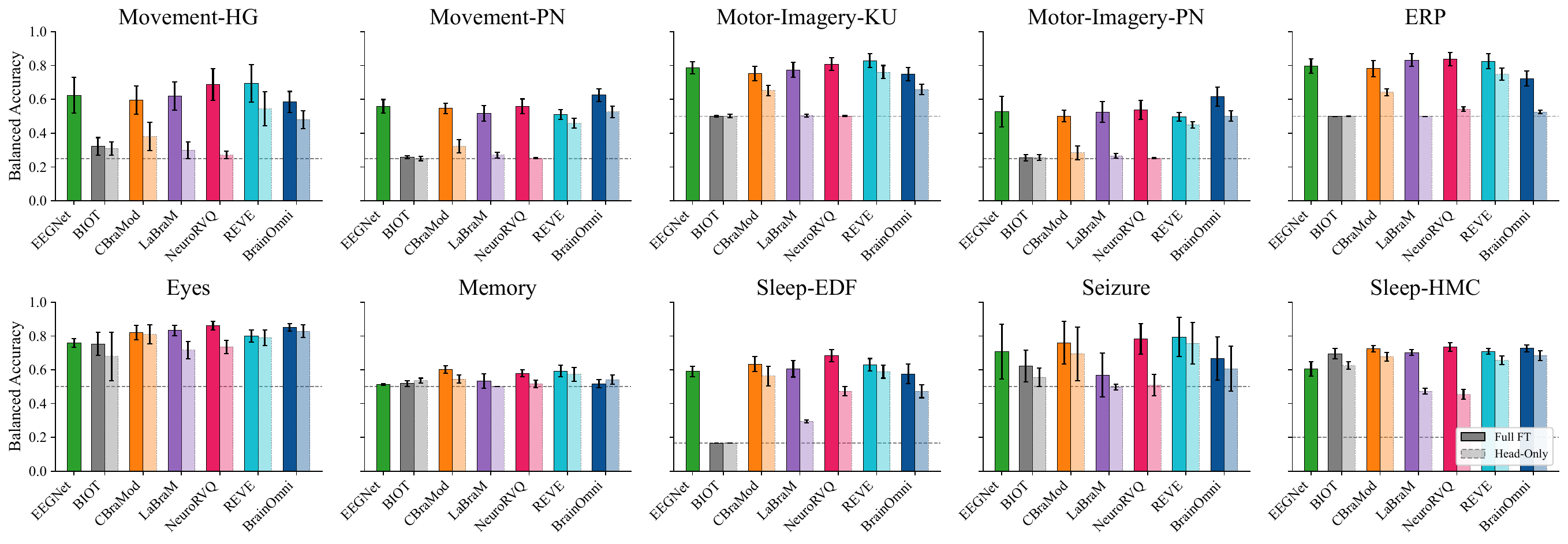}
\caption{Classification balanced accuracy of full fine-tuned and head-only fine-tuned EEG-FMs and a deep learning
baseline across 10 downstream tasks: Movement-HG, Motor-Imagery-KU, Movement-PN, Motor-Imagery-PN, ERP, Eyes, Memory, Sleep-EDF, Seizure and Sleep-HMC. Bars show mean balanced accuracy over ten subject-independent cross-validation folds (error bars show one standard deviation). The horizontal lines mark chance accuracy, $1/n_{\mathrm{classes}}$.}
\label{fig:clean}
\end{figure}

Figure~\ref{fig:clean} reports balanced accuracy per benchmark for each model under full fine-tuning and head-only adaptation (full tables and figures are in Appendix~\ref{app:clean_results_full} while Appendix~\ref{app:scratch} has a training-from-scratch analysis). Under full fine-tuning, the top six models cluster within $6.0$ percentage points ($0.647$--$0.707$) on average balanced accuracy (bacc). Per-benchmark winners differ across tasks. These results confirm the small-gains of EEG-FMs over smaller models finding of other studies. NeuroRVQ leads on average ($0.707$) despite having an order of magnitude fewer parameters than the largest models tested. EEGNet ($0.647$) matches or exceeds several foundation models on some tasks. Memory only peaks at $0.603$ (CBraMod), compared with chance accuracy of $0.500$. The large standard deviations across folds on Seizure ($0.085$--$0.155$) highlight the difficulty of generalizing across patients in clinical seizure data. BIOT fails to learn on almost every task and is excluded from subsequent analyses. Head-only adaptation underperforms full fine-tuning for every foundation model, with average performance spanning $0.432$--$0.633$ (REVE performing best), all below our deep learning baseline EEGNet ($0.647$). The model heads differ in size and pooling (Appendix~\ref{app:models}), and we examine the effect of pooling strategy in Section~\ref{sec:pooling}.

\subsection{Robustness Analysis}
\label{sec:robustness_sec}

\begin{figure}[!h]
\centering
\includegraphics[width=\textwidth]{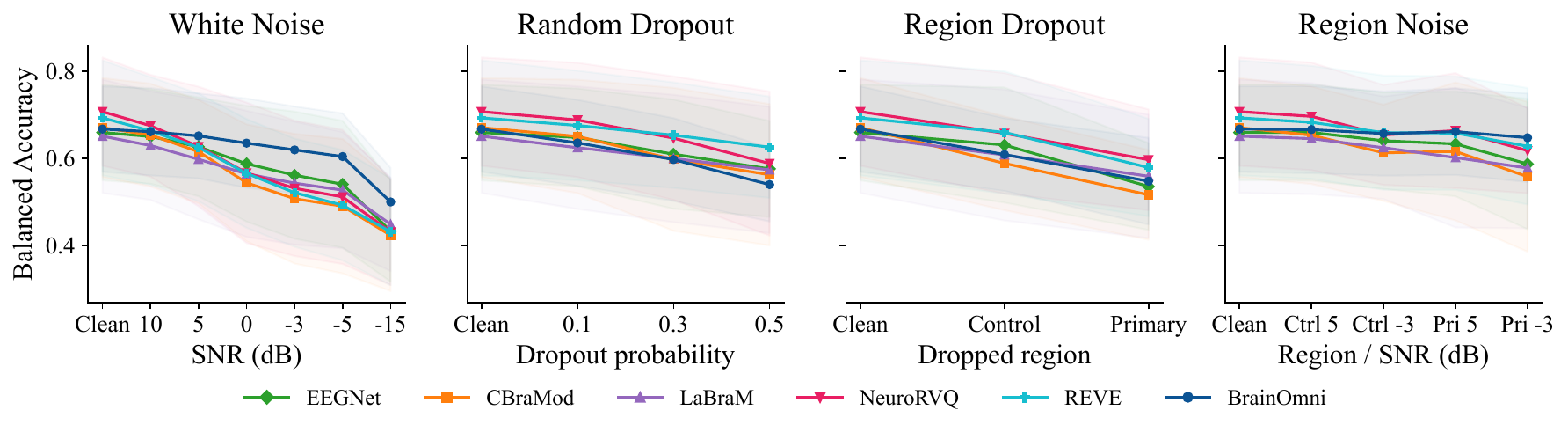}
\caption{Balanced accuracy $\pm$ 1 standard deviation under full fine-tuning, averaged across eight datasets (excluding Sleep-EDF and Sleep-HMC): (a) additive white noise, (b) random channel dropout, (c) region dropout and (d) region noise injection. Per-task figures are in Appendix~\ref{app:robustness}.}
\label{fig:rob_combined}
\end{figure}

Figure~\ref{fig:rob_combined} shows mean balanced accuracy over eight datasets. To use the same dataset set across all four perturbation conditions, we exclude the two small-montage sleep datasets (Sleep-EDF: two bipolar derivations; Sleep-HMC: four derivations) from all averages because their limited channel counts make random channel dropout unsuitable. Dataset-specific perturbation results are in Appendix~\ref{app:robustness}.

\textbf{White Noise.} Degradation patterns appear to be task-dependent. At moderate noise ($5$ to $10$\,dB), models maintain separation, indicating a \textbf{model-limited} regime where architectural differences matter. At extreme noise ($\leq -3$\,dB), most models converge toward a task-specific floor, defining a \textbf{signal-limited} regime in which the corruption dominates regardless of architecture: The Eyes and Motor-Imagery tasks remain above chance even at $-5$\,dB, while the rest degrade more sharply. BrainOmni is the most noise-robust model: at $-5$\,dB it loses only $6.3$ percentage points compared to $11.9$--$20.1$ for all other models. 
Pink noise produces similar patterns (see full results Appendix~\ref{app:robustness}).

\textbf{Random Channel Dropout.} At $p{=}0.5$, average losses range from $-6.8$ percentage points (REVE) to $-12.8$ (BrainOmni). Among the eight included datasets, the largest drop is $-35.1$ percentage points on Movement-HG and smaller on others. BrainOmni, the most noise-robust model, has the largest average loss under zero-padded dropout, showing that robustness to one perturbation type does not transfer to another. When dropped channels are removed entirely rather than zero-padded (Appendix~\ref{app:true_dropout}), the most affected models recover substantially (from $-12.8$ to $-2.4$ percentage points for BrainOmni and from $-11.9$ to $-4.8$ for NeuroRVQ), indicating that their fragility stems from how they handle zero-valued inputs rather than from the loss of task-related information itself.

\textbf{Region Dropout.} Zero-padding the primary (task-related) region hurts every model on average. LaBraM, NeuroRVQ and REVE have the smallest losses, from $-9.2$ to $-11.4$ percentage  points. Removing the dropped region instead of zero-padding it reduces the average loss for all four models that accept variable channel sets, with BrainOmni showing the largest gain from $-12.0$ to $-3.8$ percentage points (Appendix~\ref{app:robustness_region}). Dropping control regions preserves most performance on average, with the notable exception of Movement-PN and Motor-Imagery-PN where most models degrade more under control dropout. We attribute this to eye-movement artefact contamination in these datasets~\citep{lee2025capabilities} and therefore exclude them from all interpretability analyses.

\textbf{Region Noise.} BrainOmni has the smallest average loss under primary-region noise at $-3$\,dB ($-2.0$ percentage points), while CBraMod has the largest loss ($-11.3$ percentage points). Every model loses more on average under primary than control noise at this level. At $+5$\,dB, losses remain visible under primary-region noise but are small under control-region noise ($-1.7$ to $0.0$ percentage points).

\subsection{Attribution Maps}
\label{sec:interpretability}

\begin{figure}[!h]
\centering
\includegraphics[width=0.65\textwidth]{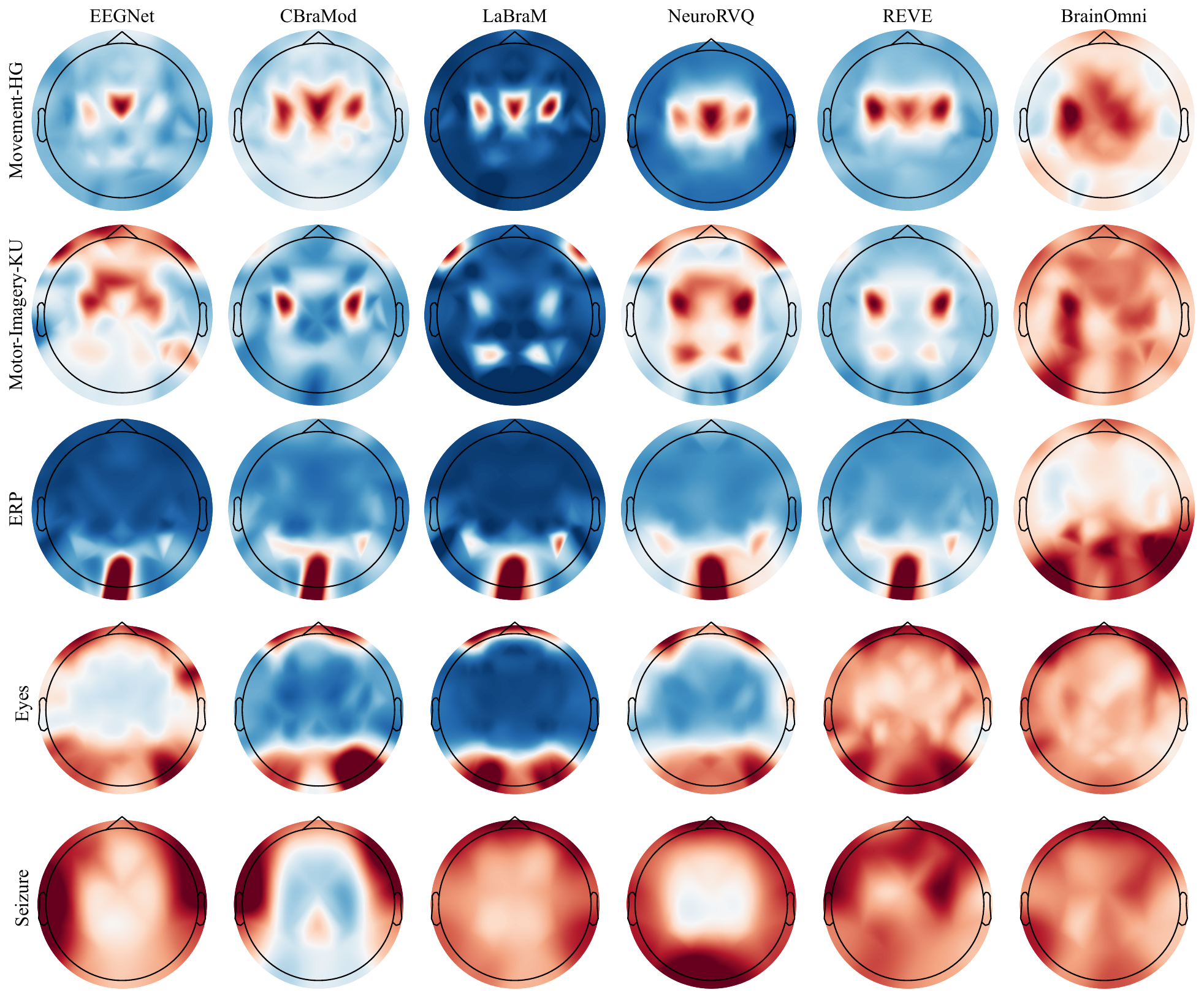}
\caption{Class-averaged attribution topographic maps. Columns (Models): EEGNet, CBraMod, LaBraM, NeuroRVQ, REVE and BrainOmni. Rows (Benchmarks): Movement-HG, Motor-Imagery-KU, ERP, Eyes and Seizure.}
\label{fig:lrp_grid}
\end{figure}

Following the methods and techniques described in Section \ref{sec:methods} (implementation details in Appendix~\ref{app:interpretability}), class-averaged attribution topographic maps were extracted for each EEG-FM across five benchmarks. 
The maps in Figure~\ref{fig:lrp_grid} often emphasise task-appropriate regions: central and centroparietal electrodes for motor tasks (Movement-HG, Motor-Imagery-KU), occipital sites for ERP and occipital and anterior electrodes for Eyes. This alignment varies by task and model. This provides evidence of a connection to expected neurophysiology, one of the concerns raised by \citet{kuruppu2026fmreview}.

\paragraph{Localisation.}  Models differ in how sharply they localise relevance. LaBraM, REVE, CBraMod, and NeuroRVQ produce tightly focused peaks over the expected cortical areas. BrainOmni is the most diffuse, spreading relevance across the scalp rather than concentrating on brain regions. 

\paragraph{Model-Specific Patterns.} On Eyes task, diverging from the occipital focus of other models, REVE and BrainOmni present diffused behaviour. On Motor-Imagery-KU, NeuroRVQ, LaBraM and EEGNet show parietal relevance alongside the expected central peak. These three models also exhibit frontal activation consistent with potential eye movement artefacts, whereas REVE and CBraMod do not. On Seizure, CBraMod concentrates relevance over temporal electrodes, while the other models produce diffused maps.

\vspace{-1pt}
\subsection{Block-wise Probing}
\label{sec:probing}

To understand where task information emerges, we train linear probes on intermediate representations extracted after each fine-tuned and pre-trained transformer block (details and additional analyses in Appendix~\ref{app:probing}). For computational reasons, we focus on REVE and NeuroRVQ as the two highest-performing models.

\begin{figure}[!h]
\centering
\includegraphics[width=\textwidth]{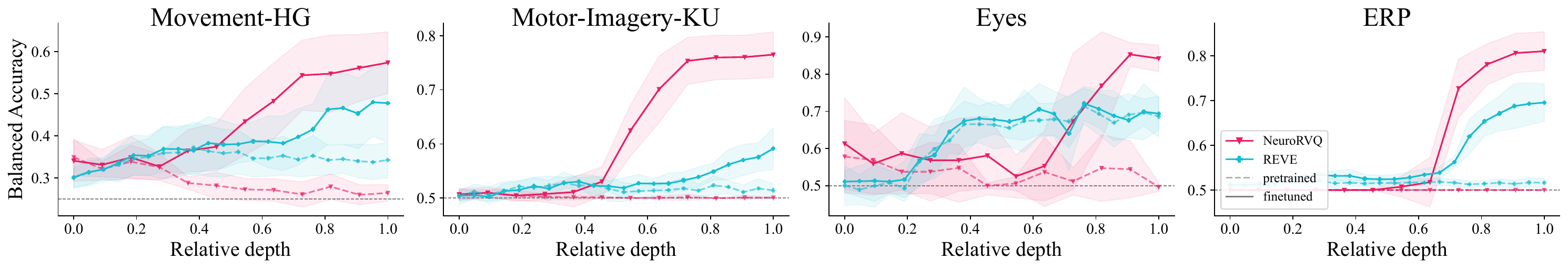}

\caption{Mean-pooled linear probing accuracy ($\pm$ 1 standard deviation over ten folds). Tasks from left to right: Movement-HG, Motor-Imagery-KU, Eyes and ERP. Relative depth runs from the output of the first block ($0$) to the output of the last ($1$). Dashed lines: pre-trained, solid lines: fine-tuned.}

\label{fig:probing}
\end{figure}

Figure~\ref{fig:probing} shows mean-pooled probing accuracy by relative depth ($0$ = first block output, $1$ = final block output) for pre-trained and fine-tuned checkpoints across four benchmarks. For both models, pre-trained and fine-tuned curves remain close in early blocks and tend to diverge in middle to late blocks. This suggests that task-relevant, linearly decodable signal is similar in early blocks before and after fine-tuning, while later blocks gain task-discriminative structure. The exact depth and size of this separation vary by task, with little separation for REVE on Eyes. Complementary analyses (where block-wise flatten probing is performed) show that task information is already present earlier in the pre-trained model (Appendices~\ref{app:probing_concat} and~\ref{app:exit_block}).

\begin{figure}[!h]
    \centering
    \begin{subfigure}{\textwidth}
        \centering
        \includegraphics[width=\textwidth]{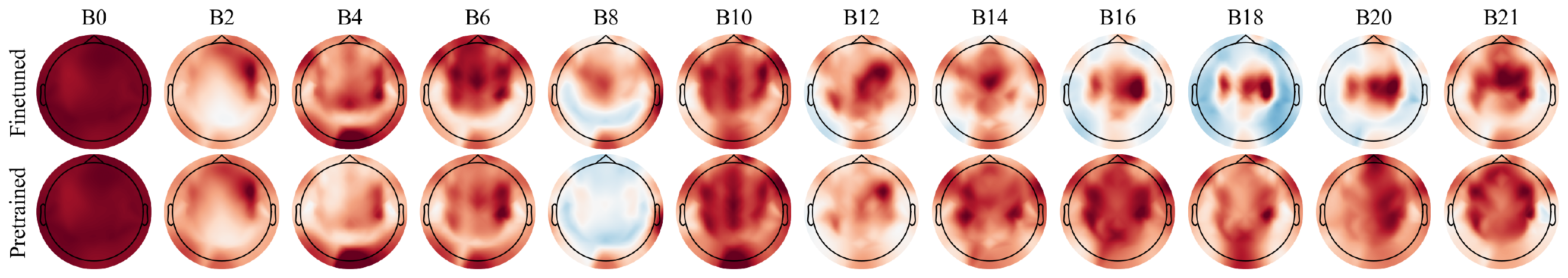}
        \caption{REVE}
        \label{fig:reve_perblock}
    \end{subfigure}
    \begin{subfigure}{\textwidth}
        \centering
        \includegraphics[width=\textwidth]{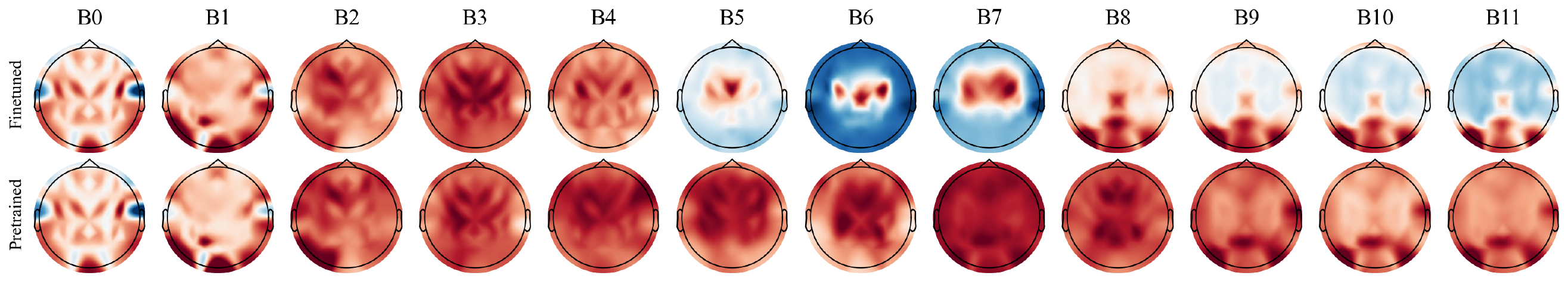}
        \caption{NeuroRVQ}
        \label{fig:neurorvq_perblock}
    \end{subfigure}
    \caption{Per-block attention topographic maps in the Movement-HG task. Top row of each panel: fine-tuned model. Bottom row: pre-trained model. Blocks (B) ordered from input to output.}
    \label{fig:perblock_attn}
\end{figure}

Attention maps show a similar transition point for both models: pre-trained and fine-tuned patterns overlap in early blocks and diverge at the same depth where probing accuracy separates (Figure~\ref{fig:perblock_attn}). Under fine-tuning on the Movement-HG task, REVE’s later blocks sharpen onto task-relevant regions, whereas NeuroRVQ’s only do so in middle blocks while later blocks shift attention occipitally. The full set of tasks is given in Appendix~\ref{app:attn_ft_perblock}. %

\subsection{Head Pooling Strategy}
\label{sec:pooling}

Poor head-only performance of some models can reflect representation quality or pooling strategy. To further examine this, we compare mean pooling, attention pooling, token flattening and a  mean-tile control for LaBraM, NeuroRVQ, REVE, BrainOmni and CBraMod. Mean-tile repeats the pooled vector to match the flattened input size and head parameter count, but does not preserve individual tokens. We use linear classifiers for this ablation and evaluate both head-only adaptation and full fine-tuning on four tasks (Appendix~\ref{app:pooling_strategy}).

\begin{figure}[!h]
\centering
\includegraphics[width=\textwidth]{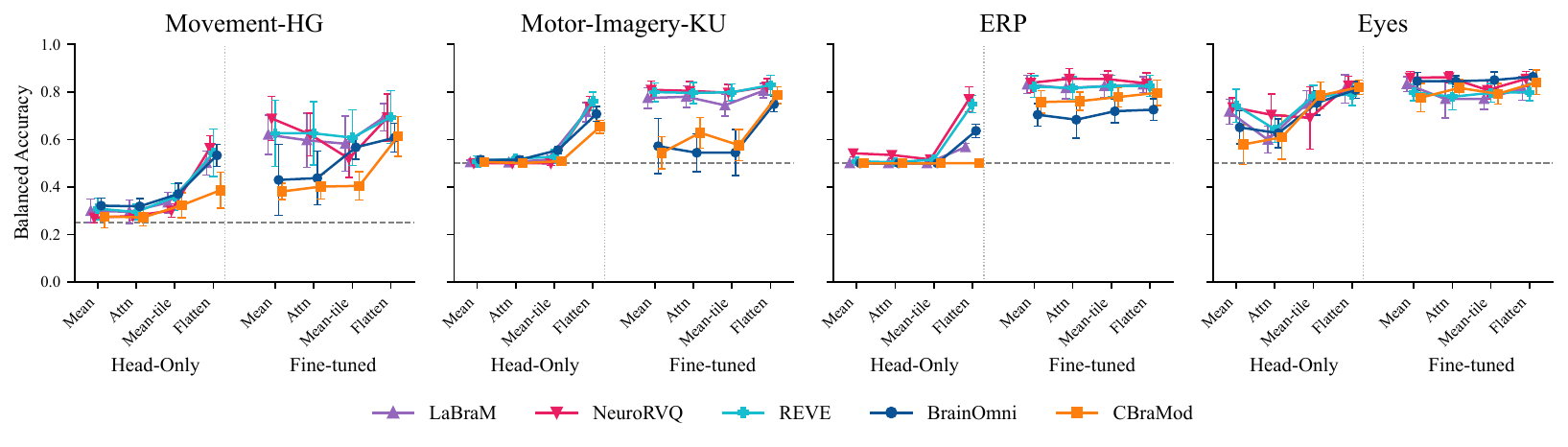}
\caption{Balanced Accuracy $\pm$ 1 standard deviation of mean, mean-tile, attention and flatten pooling for LaBraM, NeuroRVQ, REVE, BrainOmni and CBraMod under head-only adaptation and full fine-tuning. Mean-tile matches the flatten head's parameter count without preserving tokens.}
\label{fig:large_head}
\end{figure}

Under head-only adaptation, flatten pooling gives the highest average accuracy for every model (Figure~\ref{fig:large_head} and Table~\ref{tab:pooling_ablation}). Mean-tile improves several models but does not recover the flattening gain, showing that matching the head's parameter count alone is insufficient. Attention pooling does not lead to substantial improvements over mean pooling. Flatten pooling largely closes the head-only gap with EEGNet for the strongest models: NeuroRVQ reaches $0.725$ and REVE $0.711$, compared with EEGNet's $0.741$ on the same four tasks. These results show that poor head-only performance reflects selected pooling strategy rather than weak pre-trained representations. Under full fine-tuning, the effect of pooling is smaller for most models. The mean-to-flatten gain is negligible for NeuroRVQ and smaller for REVE and LaBraM, but remains present for BrainOmni and CBraMod. During full fine-tuning, the model backbone can be adapted along with the classification head,  thereby reducing the impact of suboptimal pooling strategies.

\section{Discussion}
\label{sec:discussion}

Prior benchmarking works \citep{lee2025capabilities, lee2025finetuninginsights, liu2025eegfmbenchmark} have examined the performance of EEG-FMs on clean in-distribution data, reporting modest gains over supervised baselines and weak frozen representations. In this work, we examined \emph{(i)} whether these conclusions change under synthetic perturbations, \emph{(ii)} whether attributions are spatially plausible and \emph{(iii)} whether weak head-only performance reflects weak representational quality of EEG-FMs. Across six EEG-FMs and ten datasets, we find that several conclusions depend on evaluation choices. 

\paragraph{Robustness.} Robustness is a property of perturbation type and model, and EEG-FMs do not consistently outperform EEGNet. At extreme noise ($\leq -3$\,dB), models start to converge towards a signal-limited floor that varies by task, while at moderate noise and under channel dropout model differences remain decisive. Under additive noise, BrainOmni leads by a wide margin compared to other EEG-FMs and EEGNet. However, no single model dominates all perturbation types: BrainOmni is the most fragile under zero-padded random dropout, where REVE leads (potentially due to its masked-channel pre-training strategy that other models do not follow). Furthermore, our analysis shows that this channel dropout fragility depends largely on the evaluation protocol. When channels are removed rather than zero-padded, the four variable-channel models (NeuroRVQ, LaBraM, BrainOmni and REVE) have smaller average losses (Appendix~\ref{app:true_dropout}). Dropout robustness therefore cannot be compared across studies unless the handling of missing channels is reported. Inference-time quantisation is not necessary for BrainOmni’s noise-robustness advantage, as shown in Appendix~\ref{app:vq_skip}. Two plausible sources remain: \emph{(a)} compression of input channels into latent variables before the transformer and \emph{(b)} multimodal EEG and MEG pre-training. Cross-modal transfer reported by \citet{gokce2026multimodal_brain_scaling} motivates testing the latter, but does not establish its role in robustness.

\paragraph{Interpretability.} Attribution maps generally place relevance on task-appropriate brain regions (central electrodes for motor tasks, occipital for visual ERP and occipital/frontal for eyes open/closed) with several exceptions (e.g. REVE and BrainOmni produce diffuse maps on Eyes). Region dropout supports this by removing electrodes rather than measuring sensitivity: removing primary regions caused larger average losses than removing control regions. Together, these results partially address the concern of \citet{kuruppu2026fmreview} that EEG-FMs must connect to known neurophysiology to be considered trustworthy. However, spatial plausibility is necessary but not sufficient for trustworthy decoding~\citep{jacovi2020faithfulness}. On Motor-Imagery-KU, for example, NeuroRVQ, LaBraM and EEGNet show frontal relevance consistent with eye-movement artefacts alongside the expected central peak, meaning a broadly plausible map can co-exist with artefacts. Differences in attribution patterns between models remain, meaning that architectural or pre-training decisions influence what features models rely on. For example, BrainOmni's attribution maps are generally diffuse, which may reflect that its transformer operates over abstract tokens rather than physical electrodes and/or its inference-time VQ module introduces a non-differentiable step that disrupts relevance propagation. Establishing which features actually drive model predictions will require feature-level methods, such as sparse autoencoders that separate features within model  activations~\citep{lehnschiøler2026mechanisticinterpretabilityeegfoundation}.

\paragraph{Expressiveness.} The poor performance of pre-trained EEG-FMs, previously interpreted as evidence of low-quality representations~\citep{lee2025finetuninginsights, kuruppu2026fmreview, xiong2026eegfmbench}, is substantially a pooling strategy artefact. Replacing mean pooling with flatten pooling improves head-only accuracy for all five models tested and narrows the gap with EEGNet to 1.6--3.0 percentage points for the strongest models, suggesting that mean pooling discards task-discriminative information present in the token-level embeddings. The effect is smaller under full fine-tuning, as the backbone can adapt to the pooling strategy. Frozen EEG-FM representations therefore retain more task information than previous mean-pooled probing results suggested. Linear probing across transformer depth shows that fine-tuning reshapes the later blocks, where linear decodability of the downstream task grows. Complementary analyses on pre-trained representations of the best models motivate the hypothesis that fine-tuning repurposes these blocks from the pre-training objective, a pattern previously observed in pre-trained NLP transformers~\citep{voita2019bottomup, merchant2020happens}. We recommend that future evaluations of pre-trained EEG representations test multiple pooling strategies, to avoid conflating pooling design with representation quality.

\section{Limitations}
Our evaluation has several limitations. All models are trained on clean data only, leaving open  whether noise augmentation or adversarial training would close the observed robustness gaps. No dropout protocol ranks all six models fairly: zero-padding penalises variable-channel models, while true removal excludes fixed-channel baselines. Small-montage datasets were also excluded from robustness experiments. More broadly, our synthetic perturbations do not capture the full scope of signal corruption seen in deployment, such as eye-movement and muscle artefacts or electrode shift and rotation. The roles of multimodal pre-training and latent compression in BrainOmni's noise robustness cannot be disentangled without matched pre-training ablations. Our attribution analyses do not examine incorrect or low confidence trials to show what goes wrong when decoding fails and we leave this for future work. Lastly, attribution maps reveal sensitivity, while region perturbations test the effect of changing inputs, and neither establishes which model features drive a prediction.

\newpage
\section{Conclusion}
In this work, we attempted to go beyond clean accuracy to evaluate the robustness, interpretability and expressiveness of six recent EEG-FMs across ten datasets. We found that several conclusions currently drawn about EEG-FMs depend on evaluation choices. Robustness depends on both perturbation type and model. No single model is most robust across all perturbations. Attribution maps generally place relevance on task-appropriate regions, which is a necessary but not sufficient condition for trustworthy decoding. Finally, the weak head-only performance previously taken as evidence of poor representations is substantially a pooling artefact. We therefore recommend that future evaluations of EEG-FMs report robustness per perturbation type (stating the exact testing configurations), produce attribution maps for spatial plausibility evaluation and examine multiple pooling strategies before drawing conclusions about representation quality.

\bibliographystyle{plainnat}
\bibliography{bibliography}

\begin{thebibliography}{64}
\providecommand{\natexlab}[1]{#1}
\providecommand{\url}[1]{\texttt{#1}}
\expandafter\ifx\csname urlstyle\endcsname\relax
  \providecommand{\doi}[1]{doi: #1}\else
  \providecommand{\doi}{doi: \begingroup \urlstyle{rm}\Url}\fi

\bibitem[Abnar and Zuidema(2020)]{abnar2020rollout}
Samira Abnar and Willem Zuidema.
\newblock Quantifying attention flow in transformers.
\newblock In \emph{Proceedings of the 58th Annual Meeting of the Association
  for Computational Linguistics}, pages 4190--4197. Association for
  Computational Linguistics, 2020.
\newblock \doi{10.18653/v1/2020.acl-main.385}.

\bibitem[Achtibat et~al.(2024)Achtibat, Hatefi, Dreyer, Jain, Wiegand,
  Lapuschkin, and Samek]{achtibat2024attnlrp}
Reduan Achtibat, Sayed Mohammad~Vakilzadeh Hatefi, Maximilian Dreyer, Aakriti
  Jain, Thomas Wiegand, Sebastian Lapuschkin, and Wojciech Samek.
\newblock {A}ttn{LRP}: Attention-aware layer-wise relevance propagation for
  transformers.
\newblock In Ruslan Salakhutdinov, Zico Kolter, Katherine Heller, Adrian
  Weller, Nuria Oliver, Jonathan Scarlett, and Felix Berkenkamp, editors,
  \emph{Proceedings of the 41st International Conference on Machine Learning},
  volume 235 of \emph{Proceedings of Machine Learning Research}, pages
  135--168. PMLR, 2024.
\newblock \doi{10.48550/arXiv.2402.05602}.

\bibitem[Alvarez-Estevez and Rijsman(2022)]{sleep2022hmc}
Diego Alvarez-Estevez and Roselyne Rijsman.
\newblock {Haaglanden Medisch Centrum sleep staging database}.
\newblock \emph{{PhysioNet}}, 2022.
\newblock \doi{10.13026/t79q-fr32}.

\bibitem[Ancona et~al.(2018)Ancona, Ceolini, Öztireli, and
  Gross]{ancona2018inputxgradient}
Marco Ancona, Enea Ceolini, Cengiz Öztireli, and Markus Gross.
\newblock Towards better understanding of gradient-based attribution methods
  for deep neural networks.
\newblock In \emph{International Conference on Learning Representations}, 2018.
\newblock \doi{10.48550/arXiv.1711.06104}.

\bibitem[Bach et~al.(2015)Bach, Binder, Montavon, Klauschen, Müller, and
  Samek]{bach2015lrp}
Sebastian Bach, Alexander Binder, Grégoire Montavon, Frederick Klauschen,
  Klaus-Robert Müller, and Wojciech Samek.
\newblock On pixel-wise explanations for non-linear classifier decisions by
  layer-wise relevance propagation.
\newblock \emph{PLOS ONE}, 10\penalty0 (7):\penalty0 e0130140, 2015.
\newblock \doi{10.1371/journal.pone.0130140}.

\bibitem[Barmpas et~al.(2023{\natexlab{a}})Barmpas, Panagakis, Adamos,
  Laskaris, and Zafeiriou]{Barmpas_Scattering}
Konstantinos Barmpas, Yannis Panagakis, Dimitrios~A Adamos, Nikolaos Laskaris,
  and Stefanos Zafeiriou.
\newblock {BrainWave-Scattering Net}: a lightweight network for {EEG}-based
  motor imagery recognition.
\newblock \emph{Journal of Neural Engineering}, 20\penalty0 (5):\penalty0
  056014, 2023{\natexlab{a}}.
\newblock ISSN 1741-2552.
\newblock \doi{10.1088/1741-2552/acf78a}.

\bibitem[Barmpas et~al.(2023{\natexlab{b}})Barmpas, Panagakis, Bakas, Adamos,
  Laskaris, and Zafeiriou]{dynaminc_conv}
Konstantinos Barmpas, Yannis Panagakis, Stylianos Bakas, Dimitrios~A. Adamos,
  Nikolaos Laskaris, and Stefanos Zafeiriou.
\newblock Improving generalization of {CNN}-based motor-imagery {EEG} decoders
  via dynamic convolutions.
\newblock \emph{IEEE Transactions on Neural Systems and Rehabilitation
  Engineering}, 31:\penalty0 1997--2005, 2023{\natexlab{b}}.
\newblock \doi{10.1109/TNSRE.2023.3265304}.

\bibitem[Barmpas et~al.(2024)Barmpas, Panagakis, Zoumpourlis, Adamos, Laskaris,
  and Zafeiriou]{barmpas2024causalbci}
Konstantinos Barmpas, Yannis Panagakis, Georgios Zoumpourlis, Dimitrios~A.
  Adamos, Nikolaos Laskaris, and Stefanos Zafeiriou.
\newblock A causal perspective on brainwave modeling for brain--computer
  interfaces.
\newblock \emph{Journal of Neural Engineering}, 21\penalty0 (3):\penalty0
  036001, 2024.
\newblock \doi{10.1088/1741-2552/ad3eb5}.

\bibitem[Barmpas et~al.(2026)Barmpas, Lee, Chalatsis, Raftery, Panagakis,
  Adamos, Laskaris, Koliousis, Farina, and Zafeiriou]{barmpas2025neurorvq}
Konstantinos Barmpas, Na~Lee, Dimitrios Chalatsis, William Raftery, Yannis
  Panagakis, Dimitrios~A. Adamos, Nikolaos Laskaris, Alexandros Koliousis,
  Dario Farina, and Stefanos Zafeiriou.
\newblock {NeuroRVQ}: Multi-scale biosignal tokenization for generative
  foundation models.
\newblock \emph{arXiv preprint}, 2026.
\newblock \doi{10.48550/arXiv.2510.13068}.

\bibitem[Barry et~al.(2007)Barry, Clarke, Johnstone, Magee, and
  Rushby]{barry2007eyes}
Robert~J. Barry, Adam~R. Clarke, Stuart~J. Johnstone, Christopher~A. Magee, and
  Jacqueline~A. Rushby.
\newblock {EEG} differences between eyes-closed and eyes-open resting
  conditions.
\newblock \emph{Clinical Neurophysiology}, 118\penalty0 (12):\penalty0
  2765--2773, 2007.
\newblock ISSN 1388-2457.
\newblock \doi{10.1016/j.clinph.2007.07.028}.

\bibitem[Bashashati et~al.(2007)Bashashati, Fatourechi, Ward, and
  Birch]{SignalProcessingMethods1}
Ali Bashashati, Mehrdad Fatourechi, Rabab~K Ward, and Gary~E Birch.
\newblock A survey of signal processing algorithms in
  brain{\textendash}computer interfaces based on electrical brain signals.
\newblock \emph{Journal of Neural Engineering}, 4\penalty0 (2):\penalty0
  R32--R57, 2007.
\newblock \doi{10.1088/1741-2560/4/2/r03}.

\bibitem[Brown et~al.(2020)Brown, Mann, Ryder, Subbiah, Kaplan, Dhariwal,
  Neelakantan, Shyam, Sastry, Askell, Agarwal, Herbert-Voss, Krueger, Henighan,
  Child, Ramesh, Ziegler, Wu, Winter, Hesse, Chen, Sigler, Litwin, Gray, Chess,
  Clark, Berner, McCandlish, Radford, Sutskever, and Amodei]{brown2020fewshot}
Tom Brown, Benjamin Mann, Nick Ryder, Melanie Subbiah, Jared~D Kaplan, Prafulla
  Dhariwal, Arvind Neelakantan, Pranav Shyam, Girish Sastry, Amanda Askell,
  Sandhini Agarwal, Ariel Herbert-Voss, Gretchen Krueger, Tom Henighan, Rewon
  Child, Aditya Ramesh, Daniel Ziegler, Jeffrey Wu, Clemens Winter, Chris
  Hesse, Mark Chen, Eric Sigler, Mateusz Litwin, Scott Gray, Benjamin Chess,
  Jack Clark, Christopher Berner, Sam McCandlish, Alec Radford, Ilya Sutskever,
  and Dario Amodei.
\newblock Language models are few-shot learners.
\newblock In H.~Larochelle, M.~Ranzato, R.~Hadsell, M.F. Balcan, and H.~Lin,
  editors, \emph{Advances in Neural Information Processing Systems}, volume~33,
  pages 1877--1901. Curran Associates, Inc., 2020.
\newblock \doi{10.48550/arXiv.2005.14165}.

\bibitem[Chefer et~al.(2021)Chefer, Gur, and Wolf]{chefer2021attnexplain}
Hila Chefer, Shir Gur, and Lior Wolf.
\newblock Generic attention-model explainability for interpreting bi-modal and
  encoder-decoder transformers.
\newblock In \emph{2021 IEEE/CVF International Conference on Computer Vision
  (ICCV)}, pages 387--396, 2021.
\newblock \doi{10.1109/ICCV48922.2021.00045}.

\bibitem[Cheng et~al.(2020)Cheng, Goh, Dogrusoz, Tuzel, and
  Azemi]{cheng2020subjectawarecontrastivelearningbiosignals}
Joseph~Y. Cheng, Hanlin Goh, Kaan Dogrusoz, Oncel Tuzel, and Erdrin Azemi.
\newblock Subject-aware contrastive learning for biosignals.
\newblock \emph{CoRR}, abs/2007.04871, 2020.
\newblock \doi{10.48550/arXiv.2007.04871}.

\bibitem[Detti(2020)]{siena2020}
Paolo Detti.
\newblock {Siena Scalp {EEG} Database}.
\newblock \emph{{PhysioNet}}, 2020.
\newblock \doi{10.13026/5d4a-j060}.

\bibitem[{El Ouahidi} et~al.(2025){El Ouahidi}, Lys, Th{\"o}lke, Farrugia,
  Pasdeloup, Gripon, Jerbi, and Lioi]{ouahidi2025reve}
Yassine {El Ouahidi}, Jonathan Lys, Philipp Th{\"o}lke, Nicolas Farrugia,
  Bastien Pasdeloup, Vincent Gripon, Karim Jerbi, and Giulia Lioi.
\newblock {REVE}: A foundation model for {EEG} - adapting to any setup with
  large-scale pretraining on 25,000 subjects.
\newblock In \emph{The Thirty-ninth Annual Conference on Neural Information
  Processing Systems}, 2025.
\newblock \doi{10.48550/arXiv.2510.21585}.

\bibitem[Gildenblat and contributors(2021)]{jacobgilpytorchcam}
Jacob Gildenblat and contributors.
\newblock {PyTorch} library for {CAM} methods.
\newblock \url{https://github.com/jacobgil/pytorch-grad-cam}, 2021.

\bibitem[Gokce et~al.(2026)Gokce, Tang, and
  Schrimpf]{gokce2026multimodal_brain_scaling}
Abdulkadir Gokce, Yingtian Tang, and Martin Schrimpf.
\newblock Multimodal scaling laws for task \& data-optimized models of visual
  cortex.
\newblock In \emph{Forty-third International Conference on Machine Learning},
  2026.
\newblock URL \url{https://openreview.net/forum?id=OQ6jQHJPTT}.

\bibitem[Goldberger et~al.(2000)Goldberger, Amaral, Glass, Hausdorff, Ivanov,
  Mark, Mietus, Moody, Peng, and Stanley]{goldberger2000physionet}
Ary~L. Goldberger, Luis A.~N. Amaral, Leon Glass, Jeffrey~M. Hausdorff,
  Plamen~Ch. Ivanov, Roger~G. Mark, Joseph~E. Mietus, George~B. Moody,
  Chung-Kang Peng, and H.~Eugene Stanley.
\newblock {PhysioBank}, {PhysioToolkit}, and {PhysioNet}: Components of a new
  research resource for complex physiologic signals.
\newblock \emph{Circulation}, 101\penalty0 (23):\penalty0 e215--e220, 2000.
\newblock \doi{10.1161/01.CIR.101.23.e215}.

\bibitem[Handy(2009)]{SignalProcessingMethods2}
Todd~C. Handy.
\newblock \emph{Brain Signal Analysis: Advances in Neuroelectric and
  Neuromagnetic Methods}.
\newblock MIT Press, Cambridge, MA, 2009.

\bibitem[Jacovi and Goldberg(2020)]{jacovi2020faithfulness}
Alon Jacovi and Yoav Goldberg.
\newblock Towards faithfully interpretable {NLP} systems: How should we define
  and evaluate faithfulness?
\newblock In \emph{Proceedings of the 58th Annual Meeting of the Association
  for Computational Linguistics (ACL)}, pages 4198--4205, 2020.
\newblock \doi{10.18653/v1/2020.acl-main.386}.

\bibitem[Jiang et~al.(2024)Jiang, Zhao, and Lu]{jiang2024lbm_iclr}
Wei-Bang Jiang, Li-Ming Zhao, and Bao-Liang Lu.
\newblock Large brain model for learning generic representations with
  tremendous {EEG} data in {BCI}.
\newblock In \emph{The Twelfth International Conference on Learning
  Representations}, 2024.
\newblock \doi{10.48550/arXiv.2405.18765}.

\bibitem[Kemp et~al.(2000)Kemp, Zwinderman, Tuk, Kamphuisen, and
  Oberye]{kemp2000sleepedf}
B.~Kemp, A.~H. Zwinderman, B.~Tuk, H.~A.~C. Kamphuisen, and J.~J.~L. Oberye.
\newblock Analysis of a sleep-dependent neuronal feedback loop: the slow-wave
  microcontinuity of the {EEG}.
\newblock \emph{IEEE Transactions on Biomedical Engineering}, 47\penalty0
  (9):\penalty0 1185--1194, 2000.
\newblock \doi{10.1109/10.867928}.

\bibitem[Kingma and Ba(2015)]{kingma2015adam}
Diederik~P. Kingma and Jimmy Ba.
\newblock Adam: A method for stochastic optimization.
\newblock In \emph{International Conference on Learning Representations}, 2015.
\newblock \doi{10.48550/arXiv.1412.6980}.

\bibitem[Kuruppu et~al.(2026)Kuruppu, Wagh, Kremen, and
  Varatharajah]{kuruppu2026fmreview}
Gayal Kuruppu, Neeraj Wagh, Vaclav Kremen, and Yogatheesan Varatharajah.
\newblock {EEG} foundation models: a critical review of current progress and
  future directions.
\newblock \emph{Journal of Neural Engineering}, 23\penalty0 (2):\penalty0
  021001, 2026.
\newblock \doi{10.1088/1741-2552/ae4455}.

\bibitem[Lawhern et~al.(2018)Lawhern, Solon, Waytowich, Gordon, Hung, and
  Lance]{lawhern2018eegnet}
V.~J. Lawhern, A.~J. Solon, N.~R. Waytowich, S.~M. Gordon, C.~P. Hung, and
  B.~J. Lance.
\newblock {EEGNet}: a compact convolutional neural network for {EEG}-based
  brain--computer interfaces.
\newblock \emph{Journal of Neural Engineering}, 15\penalty0 (5):\penalty0
  056013, 2018.
\newblock \doi{10.1088/1741-2552/aace8c}.

\bibitem[LeCun et~al.(2015)LeCun, Bengio, and Hinton]{lecun2015deeplearning}
Yann LeCun, Yoshua Bengio, and Geoffrey Hinton.
\newblock Deep learning.
\newblock \emph{Nature}, 521\penalty0 (7553):\penalty0 436--444, 2015.
\newblock \doi{10.1038/nature14539}.

\bibitem[Lee et~al.(2019)Lee, Kwon, Kim, Kim, Lee, Williamson, Fazli, and
  Lee]{lee2019openbmi}
Min-Ho Lee, O-Yeon Kwon, Yong-Jeong Kim, Hong-Kyung Kim, Young-Eun Lee, John
  Williamson, Siamac Fazli, and Seong-Whan Lee.
\newblock {EEG} dataset and {OpenBMI} toolbox for three {BCI} paradigms: an
  investigation into {BCI} illiteracy.
\newblock \emph{GigaScience}, 8\penalty0 (5):\penalty0 giz002, 2019.
\newblock \doi{10.1093/gigascience/giz002}.

\bibitem[Lee et~al.(2025{\natexlab{a}})Lee, Bakas, Barmpas, Panagakis, Adamos,
  Laskaris, and Zafeiriou]{lee2025capabilities}
Na~Lee, Stylianos Bakas, Konstantinos Barmpas, Yannis Panagakis, Dimitrios
  Adamos, Nikolaos Laskaris, and Stefanos Zafeiriou.
\newblock Assessing the capabilities of large brainwave foundation models.
\newblock In \emph{2025 IEEE 35th International Workshop on Machine Learning
  for Signal Processing (MLSP)}, pages 1--6, 2025{\natexlab{a}}.
\newblock \doi{10.1109/MLSP62443.2025.11204282}.

\bibitem[Lee et~al.(2025{\natexlab{b}})Lee, Barmpas, Panagakis, Adamos,
  Laskaris, and Zafeiriou]{lee2025finetuninginsights}
Na~Lee, Konstantinos Barmpas, Yannis Panagakis, Dimitrios Adamos, Nikolaos
  Laskaris, and Stefanos Zafeiriou.
\newblock Are large brainwave foundation models capable yet? insights from
  fine-tuning.
\newblock In \emph{Forty-second International Conference on Machine Learning},
  2025{\natexlab{b}}.
\newblock \doi{10.48550/arXiv.2507.01196}.

\bibitem[Lehn-Schiøler et~al.(2026)Lehn-Schiøler, Kjær, Thapa, Pedersen,
  Storgaard, Williams, Gatej, Lehn-Schiøler, Brink-Kjær, Puthusserypady,
  Beniczky, Zou, and
  Hansen]{lehnschiøler2026mechanisticinterpretabilityeegfoundation}
William Lehn-Schiøler, Magnus~Ruud Kjær, Rahul Thapa, Magnus~Guldberg
  Pedersen, Anton~Mosquera Storgaard, Nick Williams, Radu Gatej, Tue
  Lehn-Schiøler, Andreas Brink-Kjær, Sadasivan Puthusserypady, Sándor
  Beniczky, James Zou, and Lars~Kai Hansen.
\newblock Mechanistic interpretability of {EEG} foundation models via sparse
  autoencoders.
\newblock \emph{arXiv preprint}, 2026.
\newblock \doi{10.48550/arXiv.2605.13930}.

\bibitem[Liu et~al.(2026)Liu, Chen, Chen, Cui, Wen, An, Luo, and
  Wu]{liu2025eegfmbenchmark}
Dingkun Liu, Yuheng Chen, Zhu Chen, Zhenyao Cui, Yaozhi Wen, Jiayu An, Jingwei
  Luo, and Dongrui Wu.
\newblock {EEG-FM-Compass}: Progress, benchmarking, and future directions for
  {EEG} foundation models.
\newblock \emph{National Science Review}, page nwag466, 2026.
\newblock \doi{10.1093/nsr/nwag466}.

\bibitem[Loshchilov and Hutter(2019)]{loshchilov2019adamw}
Ilya Loshchilov and Frank Hutter.
\newblock Decoupled weight decay regularization.
\newblock In \emph{International Conference on Learning Representations}, 2019.
\newblock \doi{10.48550/arXiv.1711.05101}.

\bibitem[Lotte et~al.(2018)Lotte, Bougrain, Cichocki, Clerc, Congedo,
  Rakotomamonjy, and Yger]{lotte2018bciupdate}
Fabien Lotte, Laurent Bougrain, Andrzej Cichocki, Maureen Clerc, Marco Congedo,
  Alain Rakotomamonjy, and Fabrice Yger.
\newblock A review of classification algorithms for {EEG}-based brain--computer
  interfaces: a 10 year update.
\newblock \emph{Journal of Neural Engineering}, 15\penalty0 (3):\penalty0
  031005, 2018.
\newblock \doi{10.1088/1741-2552/aab2f2}.

\bibitem[Ma and Wang(2026)]{ma2026eegxplain}
Hansong Ma and Junxiao Wang.
\newblock {EEG-Xplain}: Decoding neural black-boxes of {EEG} foundation models.
\newblock \emph{arXiv preprint}, 2026.
\newblock \doi{10.48550/arXiv.2609.15687}.

\bibitem[McFarland and Wolpaw(2011)]{mcfarland2011cacm_bci}
Dennis~J. McFarland and Jonathan~R. Wolpaw.
\newblock Brain--computer interfaces for communication and control.
\newblock \emph{Communications of the ACM}, 54\penalty0 (5):\penalty0 60--66,
  2011.
\newblock \doi{10.1145/1941487.1941506}.

\bibitem[McFarland et~al.(2006)McFarland, Anderson, Muller, Schlogl, and
  Krusienski]{Classical}
D.J. McFarland, C.W. Anderson, K.-R. Muller, A.~Schlogl, and D.J. Krusienski.
\newblock {BCI} meeting 2005-workshop on {BCI} signal processing: feature
  extraction and translation.
\newblock \emph{IEEE Transactions on Neural Systems and Rehabilitation
  Engineering}, 14\penalty0 (2):\penalty0 135--138, 2006.
\newblock \doi{10.1109/TNSRE.2006.875637}.

\bibitem[Merchant et~al.(2020)Merchant, Rahimtoroghi, Pavlick, and
  Tenney]{merchant2020happens}
Amil Merchant, Elahe Rahimtoroghi, Ellie Pavlick, and Ian Tenney.
\newblock What happens to {BERT} embeddings during fine-tuning?
\newblock In \emph{Proceedings of the Third BlackboxNLP Workshop on Analyzing
  and Interpreting Neural Networks for NLP}, pages 33--44, Online, 2020.
  Association for Computational Linguistics.
\newblock \doi{10.18653/v1/2020.blackboxnlp-1.4}.

\bibitem[{Meyer zu Bexten} et~al.(2026){Meyer zu Bexten}, Scherf, Franczyk, and
  Hofmann]{attnLRPcleverhans}
Justus {Meyer zu Bexten}, Nico Scherf, Bogdan Franczyk, and Simon~M. Hofmann.
\newblock From {Clever Hans} to scientific discovery: Interpreting {EEG}
  foundational transformers with {LRP}.
\newblock \emph{arXiv preprint}, 2026.
\newblock \doi{10.48550/arXiv.2605.11885}.

\bibitem[Nam et~al.(2018)Nam, Nijholt, and Lotte]{SignalProcessingMethods4}
Chang~S. Nam, Anton Nijholt, and Fabien Lotte.
\newblock \emph{Brain--Computer Interfaces Handbook: Technological and
  Theoretical Advances}.
\newblock CRC Press, 2018.

\bibitem[Neuper et~al.(2006)Neuper, Wörtz, and
  Pfurtscheller]{neuper2006sensorimotor}
Christa Neuper, Michael Wörtz, and Gert Pfurtscheller.
\newblock {ERD/ERS} patterns reflecting sensorimotor activation and
  deactivation.
\newblock In Christa Neuper and Wolfgang Klimesch, editors, \emph{Event-Related
  Dynamics of Brain Oscillations}, volume 159 of \emph{Progress in Brain
  Research}, pages 211--222. Elsevier, 2006.
\newblock \doi{10.1016/S0079-6123(06)59014-4}.

\bibitem[Nicolas-Alonso and Gomez-Gil(2012)]{nicolasalonso2012bci_review}
L.~F. Nicolas-Alonso and J.~Gomez-Gil.
\newblock Brain--computer interfaces, a review.
\newblock \emph{Sensors}, 12\penalty0 (2):\penalty0 1211--1279, 2012.
\newblock \doi{10.3390/s120201211}.

\bibitem[Pavlov et~al.(2022)Pavlov, Kasanov, Kosachenko, Kotyusov, and
  Busch]{pavlov2022digitspan_dataset}
Yuri~G. Pavlov, Dauren Kasanov, Alexandra~I. Kosachenko, Alexander~I. Kotyusov,
  and Niko~A. Busch.
\newblock Pupillometry and electroencephalography in the digit span task.
\newblock \emph{Scientific Data}, 9\penalty0 (1):\penalty0 325, 2022.
\newblock \doi{10.1038/s41597-022-01414-2}.

\bibitem[Pfurtscheller and Neuper(2001)]{pfurtscheller2001motorimagery}
G.~Pfurtscheller and C.~Neuper.
\newblock Motor imagery and direct brain--computer communication.
\newblock \emph{Proceedings of the IEEE}, 89\penalty0 (7):\penalty0 1123--1134,
  2001.
\newblock \doi{10.1109/5.939829}.

\bibitem[Rao(2013)]{SignalProcessingMethods3}
Rajesh P.~N. Rao.
\newblock \emph{Brain-Computer Interfacing: An Introduction}.
\newblock Cambridge University Press, 2013.

\bibitem[Rauch et~al.(2026)Rauch, Heinrich, Ghaffari, Miklautz, Moummad, Sick,
  and Scholz]{rauch2025unmutepatchtokensrethinking}
Lukas Rauch, René Heinrich, Houtan Ghaffari, Lukas Miklautz, Ilyass Moummad,
  Bernhard Sick, and Christoph Scholz.
\newblock Unmute the patch tokens: Rethinking probing in multi-label audio
  classification.
\newblock In \emph{The Fourteenth International Conference on Learning
  Representations}, 2026.
\newblock \doi{10.48550/arXiv.2509.24901}.

\bibitem[Ravindran and Contreras-Vidal(2023)]{ravindran2023eegxai_groundtruth}
A.~Sujatha Ravindran and Jose Contreras-Vidal.
\newblock An empirical comparison of deep learning explainability approaches
  for {EEG} using simulated ground truth.
\newblock \emph{Scientific Reports}, 13:\penalty0 17709, 2023.
\newblock \doi{10.1038/s41598-023-43871-8}.

\bibitem[Saha and Baumert(2020)]{saha2020variabilityreview}
S.~Saha and M.~Baumert.
\newblock Intra- and inter-subject variability in {EEG}-based sensorimotor
  brain--computer interface: A review.
\newblock \emph{Frontiers in Computational Neuroscience}, 13, 2020.
\newblock \doi{10.3389/fncom.2019.00087}.

\bibitem[Santamaría-Vázquez et~al.(2020)Santamaría-Vázquez,
  Martínez-Cagigal, Vaquerizo-Villar, and Hornero]{EEGInception}
Eduardo Santamaría-Vázquez, Víctor Martínez-Cagigal, Fernando
  Vaquerizo-Villar, and Roberto Hornero.
\newblock {EEG-Inception}: A novel deep convolutional neural network for
  assistive {ERP}-based brain-computer interfaces.
\newblock \emph{IEEE Transactions on Neural Systems and Rehabilitation
  Engineering}, 28\penalty0 (12):\penalty0 2773--2782, 2020.
\newblock \doi{10.1109/TNSRE.2020.3048106}.

\bibitem[Schalk et~al.(2004)Schalk, McFarland, Hinterberger, Birbaumer, and
  Wolpaw]{schalk2004bci2000}
Gerwin Schalk, Dennis~J. McFarland, Thilo Hinterberger, Niels Birbaumer, and
  Jonathan~R. Wolpaw.
\newblock {BCI2000}: a general-purpose brain--computer interface ({BCI})
  system.
\newblock \emph{IEEE Transactions on Biomedical Engineering}, 51\penalty0
  (6):\penalty0 1034--1043, 2004.
\newblock \doi{10.1109/TBME.2004.827072}.

\bibitem[Schirrmeister et~al.(2017)Schirrmeister, Springenberg, Fiederer,
  Glasstetter, Eggensperger, Tangermann, Hutter, Burgard, and
  Ball]{schirrmeister2017deepconvnet}
Robin~Tibor Schirrmeister, Jost~Tobias Springenberg, Lukas Dominique~Josef
  Fiederer, Martin Glasstetter, Katharina Eggensperger, Michael Tangermann,
  Frank Hutter, Wolfram Burgard, and Tonio Ball.
\newblock Deep learning with convolutional neural networks for {EEG} decoding
  and visualization.
\newblock \emph{Human Brain Mapping}, 38\penalty0 (11):\penalty0 5391--5420,
  2017.
\newblock ISSN 1065-9471.
\newblock \doi{10.1002/hbm.23730}.

\bibitem[Selvaraju et~al.(2017)Selvaraju, Cogswell, Das, Vedantam, Parikh, and
  Batra]{selvaraju2017gradcam}
Ramprasaath~R. Selvaraju, Michael Cogswell, Abhishek Das, Ramakrishna Vedantam,
  Devi Parikh, and Dhruv Batra.
\newblock {G}rad-{CAM}: Visual explanations from deep networks via
  gradient-based localization.
\newblock In \emph{2017 IEEE International Conference on Computer Vision
  (ICCV)}, pages 618--626, 2017.
\newblock \doi{10.1109/ICCV.2017.74}.

\bibitem[Shrikumar et~al.(2017)Shrikumar, Greenside, and Kundaje]{deeplift}
Avanti Shrikumar, Peyton Greenside, and Anshul Kundaje.
\newblock Learning important features through propagating activation
  differences.
\newblock In \emph{ICML}, pages 3145--3153, 2017.
\newblock \doi{10.48550/arXiv.1704.02685}.

\bibitem[{\v{S}}irca et~al.(2026){\v{S}}irca, Brulec, and
  Alimardani]{sirca2026peft}
Urban {\v{S}}irca, Lovro Brulec, and Maryam Alimardani.
\newblock Parameter-efficient fine-tuning of {EEG} foundation models for
  plug-and-play motor imagery {BCI}s.
\newblock In \emph{2026 14th International Conference on Brain-Computer
  Interface (BCI)}, pages 1--7, 2026.
\newblock \doi{10.1109/BCI69045.2026.11435102}.

\bibitem[Smilkov et~al.(2017)Smilkov, Thorat, Kim, Vi{\'e}gas, and
  Wattenberg]{smilkov2017smoothgrad}
Daniel Smilkov, Nikhil Thorat, Been Kim, Fernanda Vi{\'e}gas, and Martin
  Wattenberg.
\newblock {S}mooth{G}rad: removing noise by adding noise, 2017.
\newblock Presented at the ICML 2017 Workshop on Visualization for Deep
  Learning.

\bibitem[Song et~al.(2023)Song, Zheng, Liu, and Gao]{EEGConformer}
Yonghao Song, Qingqing Zheng, Bingchuan Liu, and Xiaorong Gao.
\newblock {EEG} {Conformer}: Convolutional transformer for {EEG} decoding and
  visualization.
\newblock \emph{IEEE Transactions on Neural Systems and Rehabilitation
  Engineering}, 31:\penalty0 710--719, 2023.
\newblock \doi{10.1109/TNSRE.2022.3230250}.

\bibitem[Touvron et~al.(2023)Touvron, Lavril, Izacard, Martinet, Lachaux,
  Lacroix, Rozière, Goyal, Hambro, Azhar, Rodriguez, Joulin, Grave, and
  Lample]{touvron2023llama}
Hugo Touvron, Thibaut Lavril, Gautier Izacard, Xavier Martinet, Marie-Anne
  Lachaux, Timothée Lacroix, Baptiste Rozière, Naman Goyal, Eric Hambro,
  Faisal Azhar, Aurelien Rodriguez, Armand Joulin, Edouard Grave, and Guillaume
  Lample.
\newblock {LLaMA}: Open and efficient foundation language models.
\newblock \emph{arXiv preprint}, 2023.
\newblock \doi{10.48550/arXiv.2302.13971}.

\bibitem[Voita et~al.(2019)Voita, Sennrich, and Titov]{voita2019bottomup}
Elena Voita, Rico Sennrich, and Ivan Titov.
\newblock The bottom-up evolution of representations in the transformer: A
  study with machine translation and language modeling objectives.
\newblock In \emph{Proceedings of the 2019 Conference on Empirical Methods in
  Natural Language Processing and the 9th International Joint Conference on
  Natural Language Processing (EMNLP-IJCNLP)}, pages 4395--4405, Hong Kong,
  China, 2019. Association for Computational Linguistics.
\newblock \doi{10.18653/v1/D19-1448}.

\bibitem[Wang et~al.(2025)Wang, Zhao, Luo, Zhou, Jiang, Li, Li, and
  Pan]{wang2024cbramod}
Jiquan Wang, Sha Zhao, Zhiling Luo, Yangxuan Zhou, Haiteng Jiang, Shijian Li,
  Tao Li, and Gang Pan.
\newblock {CBraMod}: A criss-cross brain foundation model for {EEG} decoding.
\newblock In \emph{The Thirteenth International Conference on Learning
  Representations}, 2025.
\newblock \doi{10.48550/arXiv.2412.07236}.

\bibitem[Werth et~al.(1996)Werth, Achermann, and
  Borb{\'e}ly]{werth1996sleeptopography}
Esther Werth, Peter Achermann, and Alexander~A. Borb{\'e}ly.
\newblock Brain topography of the human sleep {EEG}: antero-posterior shifts of
  spectral power.
\newblock \emph{NeuroReport}, 8\penalty0 (1):\penalty0 123--127, 1996.
\newblock \doi{10.1097/00001756-199612200-00025}.

\bibitem[Xiao et~al.(2025)Xiao, Cui, Zhang, Chen, Wu, Thwaites, Woolgar, Zhou,
  and Zhang]{xiao2025brainomni}
Qinfan Xiao, Ziyun Cui, Chi Zhang, Siqi Chen, Wen Wu, Andrew Thwaites,
  Alexandra Woolgar, Bowen Zhou, and Chao Zhang.
\newblock {BrainOmni}: A brain foundation model for unified {EEG} and {MEG}
  signals.
\newblock In \emph{The Thirty-ninth Annual Conference on Neural Information
  Processing Systems}, 2025.
\newblock \doi{10.48550/arXiv.2505.18185}.

\bibitem[Xiong et~al.(2026)Xiong, Li, Li, Zhu, and Jiang]{xiong2026eegfmbench}
Wei Xiong, Jiangtong Li, Jie Li, Kun Zhu, and Changjun Jiang.
\newblock {EEG-FM-Bench}: A comprehensive benchmark for the systematic
  evaluation and diagnostic analyses of {EEG} foundation models.
\newblock In \emph{Forty-third International Conference on Machine Learning},
  2026.
\newblock \doi{10.48550/arXiv.2508.17742}.

\bibitem[Yang et~al.(2023)Yang, Westover, and Sun]{yang2023biot}
Chaoqi Yang, M~Brandon Westover, and Jimeng Sun.
\newblock {BIOT}: Biosignal transformer for cross-data learning in the wild.
\newblock In \emph{Thirty-seventh Conference on Neural Information Processing
  Systems}, 2023.
\newblock \doi{10.48550/arXiv.2305.10351}.

\bibitem[Zhou et~al.(2023)Zhou, Liu, Zhou, Wang, Zhai, Jia, Guan, and
  Liu]{zhou2023robustxaisurvey}
Xinliang Zhou, Chenyu Liu, Jinan Zhou, Zhongruo Wang, Liming Zhai, Ziyu Jia,
  Cuntai Guan, and Yang Liu.
\newblock Interpretable and robust {AI} in {EEG} systems: A survey.
\newblock \emph{CoRR}, abs/2304.10755, 2023.
\newblock \doi{10.48550/arXiv.2304.10755}.

\end{thebibliography}

\newpage
\appendix
\section{Model Architectures}
\label{app:models}

Our model set spans the more recent EEG-FM releases (NeuroRVQ, REVE and BrainOmni) and the established baselines from prior EEG-FM benchmarks (BIOT, LaBraM and CBraMod). The following EEG-FMs differ in many aspects, including positional encoding, pre-training objective and tokenisation, and form the model set used in our analyses. BIOT \citep{yang2023biot} was one of the first foundation models for bio-signals. It tokenises each channel into fixed-length segments and uses contrastive pre-training on clinical EEG and ECG recordings. LaBraM \citep{jiang2024lbm_iclr}, inspired by the vision transformer, segments EEG into channel patches and trains a vector-quantised (VQ) neural tokeniser to build a codebook of neural codes. A transformer-based model is then pre-trained to predict masked tokens from this codebook. CBraMod \citep{wang2024cbramod} separates spatial and temporal attention into two parallel streams using a criss-cross transformer and utilises asymmetric convolutional positional encoding to adapt to arbitrary electrode layouts. NeuroRVQ \citep{barmpas2025neurorvq} extends the VQ approach of LaBraM to multiple frequency scales through four parallel temporal branches, each with its own residual VQ (RVQ) codebook. Combined with a phase-and-amplitude-aware training loss, the NeuroRVQ tokeniser achieves high-fidelity signal reconstruction across all frequency bands. REVE \citep{ouahidi2025reve} represents the largest EEG pre-training effort to date, training on over 60,000 hours of data and 25,000 subjects. It introduces a 4D positional encoding scheme that generates embeddings from the 3D electrode coordinates and the temporal patch index, removing the need for fixed lookup tables. BrainOmni \citep{xiao2025brainomni} is the first model to pre-train on both EEG and MEG data. It encodes sensor properties such as 3D position, orientation, and type, and compresses channels into a fixed set of latent source variables on which the transformer operates, retaining its VQ tokeniser at inference time.

Table~\ref{tab:model_architectures} summarises the architecture of each model, including spatial encoding technique, pooling strategy after the last backbone block and the number of trainable parameters in head-only training mode. Parameter counts are reported for the Eyes task (2 classes) ~\citep{goldberger2000physionet} but head sizes vary slightly across tasks due to input-dependent dimensions: \emph{(i)} BIOT applies an ELU activation followed by a single linear layer, \emph{(ii)} LaBraM applies dropout followed by a single linear layer, \emph{(iii)} NeuroRVQ inserts a LayerNorm before a single linear layer, \emph{(iv)} REVE flattens its tokens then applies RMSNorm, dropout and a single linear layer, \emph{(v)} BrainOmni uses a two-layer MLP (consisting of dropout, linear layer, SELU, linear layer) and \emph{(vi)} CBraMod stacks three linear layers separated by ELU activations and dropout. For EEGNet, we treat the entire model as backbone.

\begin{table}[!h]
\caption{Model architectural details.}
\label{tab:model_architectures}
\centering
\small
\begin{tabular}{l|c|crr}
\toprule
Model & Pooling & Backbone Parameters & Head Parameters & \% \\
\midrule
EEGNet & --- & 3.4K & --- & --- \\
BIOT & Mean & 3.19M & 257 & $<$0.1 \\
CBraMod & Flatten  & 4.88M & 41.12M & 89.4 \\
LaBraM & Mean  & 5.82M & 201 & $<$0.1 \\
NeuroRVQ & Mean  & 5.87M & 2.4K & $<$0.1 \\
BrainOmni$^{*\dagger}$ & Mean + Flatten  & 37.76M & 6.29M & 14.3 \\
REVE & Flatten  & 69.19M & 393.2K & 0.6 \\
\bottomrule
\multicolumn{5}{l}{\footnotesize $^*$Mean pools over temporal tokens per channel; flatten concatenates across latent source variables.} \\
\multicolumn{5}{l}{\footnotesize $^\dagger$BrainOmni keeps its 5.05M VQ tokeniser frozen during full fine-tuning.}
\end{tabular}
\end{table}

\section{Training Protocol}
\label{app:training}

All models are trained on clean data only (datasets described in Appendix~\ref{app:datasets}), using ten-fold leave-many-subjects-out cross-validation (LOMSO-CV). We train for a maximum of 20 epochs with early stopping with patience of 5 epochs. Each model uses the hyperparameters recommended by its original authors. All models are optimised with AdamW \cite{loshchilov2019adamw} except BIOT, which uses Adam \cite{kingma2015adam}. LaBraM and CBraMod use cross-entropy with label smoothing of $0.1$ for multiclass tasks and binary cross-entropy for binary tasks. BrainOmni uses cross-entropy with label smoothing of $0.1$. Additional model-specific settings include: LaBraM uses drop-path $0.1$, NeuroRVQ uses layer-wise learning rate (LR) decay $0.975$, BIOT uses attention dropout $0.2$, REVE and BrainOmni cosine schedulers use $\eta_{\min}=\text{LR}\times 0.1$ and LaBraM and CBraMod use $\eta_{\min}=10^{-6}$. Per-model learning rates and schedules are listed in Table~\ref{tab:hyperparameters}.

All experiments run on a single NVIDIA H100 GPU. Batch size is set per model and per benchmark to the largest value that fits in memory. We report balanced accuracy as mean $\pm$ standard deviation across the ten LOMSO folds. The Average column in result tables averages the per-task means across tasks rather than across folds.

\begin{table}[h]
\caption{Per-model learning rate, weight decay, and schedule.}
\label{tab:hyperparameters}
\centering
\small
\begin{tabular}{lllll}
\toprule
Model & Learning Rate & Weight decay & Scheduler & Warm-up \\
\midrule
EEGNet    & $1 \times 10^{-3}$ & $0$                 & Cosine                 & ---        \\
LaBraM    & $5 \times 10^{-4}$ & $5 \times 10^{-2}$  & Cosine                 & 4 epochs   \\
NeuroRVQ  & $5 \times 10^{-4}$ & $1 \times 10^{-2}$  & Linear Warm-up + Decay  & 4 epochs   \\
CBraMod   & $5 \times 10^{-4}$ & $5 \times 10^{-2}$  & Cosine                 & ---        \\
BrainOmni & $1 \times 10^{-4}$ & $1 \times 10^{-5}$  & Linear Warm-up + Cosine & 10\% steps \\
REVE      & $1 \times 10^{-4}$ & $1 \times 10^{-5}$  & Linear Warm-up + Cosine & 10\% steps \\
BIOT      & $1 \times 10^{-3}$ & $1 \times 10^{-5}$  & None                   & ---        \\
 \bottomrule
\end{tabular}
\end{table}

\subsection{Computational Cost}
\label{app:compute}

Table~\ref{tab:compute_full_finetuned} reports training/inference GPU hours and peak memory use on a single NVIDIA H100. The numbers cover every training and evaluation run in this paper, across both adaptation regimes and all perturbation conditions, on all ten datasets. The total is 875.9 GPU-hours: 838.5 for training and 37.4 for inference. Compared to the EEGNet baseline, the foundation models are far more expensive to train and evaluate.

\begin{table}[h]
\centering
\caption{Computational cost of the complete 10-dataset workload on a single NVIDIA H100. GPU-hours are summed; peak memory is a maximum. Rows are ordered by total cost.}
\label{tab:compute_full_finetuned}
\scriptsize
\setlength{\tabcolsep}{3pt}
\begin{tabular*}{\textwidth}{@{\extracolsep{\fill}}lrrrrr}
\toprule
Model & Train (GPU-h) & Inference (GPU-h) & Total (GPU-h) & Peak train (GB) & Peak inf.\ (GB) \\
\midrule
EEGNet & 4.87 & 0.50 & 5.37 & 1.8 & 1.3 \\
BIOT & 9.44 & 2.16 & 11.60 & 89.1 & 20.0 \\
LaBraM & 20.55 & 2.11 & 22.66 & 98.9 & 8.7 \\
CBraMod & 37.71 & 1.30 & 39.01 & 77.6 & 5.3 \\
REVE & 77.03 & 4.08 & 81.12 & 78.2 & 3.5 \\
NeuroRVQ & 100.77 & 5.61 & 106.38 & 99.1 & 4.1 \\
BrainOmni & 588.09 & 21.66 & 609.75 & 97.3 & 30.5 \\
\bottomrule
\end{tabular*}
\end{table}

\section{Datasets and Preprocessing}
\label{app:datasets}

Table~\ref{tab:datasets} summarises the ten benchmark tasks. All datasets\footnote{In PhysioNet ME and MI, six participants (subjects 88, 89, 92, 100, 104, and 106) were excluded due to differences in either the sampling frequency or duration of the performed tasks.} are as distributed and were not included in the pre-training corpus of any foundation model we evaluate \footnote{With the exception of NeuroRVQ whose pre-training corpus includes PhysioNet Movement and Motor-Imagery tasks}.

\begin{table}[h]
\caption{Benchmark datasets details. Trial length refers to the extracted window duration.}
\label{tab:datasets}
\centering
\small
\begin{tabular}{l|l|l|cccc}
\toprule
Dataset & Name & Paradigm & Classes & Channels & Trial length & Subjects \\
\midrule
High-Gamma & Movement-HG & Motor execution & 4 & 78 & 4\,s & 14 \\
OpenBMI-MI & Motor-Imagery-KU & Motor imagery & 2 & 62 & 4\,s & 54 \\
OpenBMI-ERP & ERP & Visual P300 & 2 & 62 & 1\,s & 54 \\
PhysioNet-MI & Motor-Imagery-PN & Motor imagery & 4 & 64 & 4\,s & 103 \\
PhysioNet-ME & Movement-PN & Motor execution & 4 & 64 & 4\,s & 103\\
PhysioNet Eyes & Eyes & Eyes open/closed & 2 & 64 & 4\,s & 103\\
Pavlov Memory & Memory & Working memory & 2 & 63& 4\,s & 65\\
Sleep EDF & Sleep-EDF & Sleep staging & 6& 2 & 30\,s & 78 \\
Siena & Seizure & Seizure detection & 2 & 19 & 4\,s & 14 \\
HMC & Sleep-HMC & Sleep staging & 5 & 4 & 30\,s & 151 \\
\bottomrule
\multicolumn{7}{l}{\footnotesize $^\dagger$Sleep-EDF channels are bipolar derivations (Fpz-Cz and Pz-Oz).} \\
\multicolumn{7}{l}{\footnotesize $^\ddagger$Sleep-HMC channels are mastoid-referenced derivations (F4-M1, C4-M1, O2-M1 and C3-M2).} \\
\end{tabular}
\end{table}

Trial extraction (length, baseline correction and rejection criteria) follows the original publications and \citep{lee2025capabilities}; we refer readers to those works for full details. We apply a Common Average Reference (CAR) to all recordings before any perturbation or model-specific filtering. Two preprocessing pipelines are used: \emph{(i)} The default pipeline (EEGNet~\citep{lawhern2018eegnet}, LaBraM~\citep{jiang2024lbm_iclr}, CBraMod~\citep{wang2024cbramod}, BIOT~\citep{yang2023biot}, NeuroRVQ~\citep{barmpas2025neurorvq}, REVE~\citep{ouahidi2025reve}) resamples to 200\,Hz, applies a fourth-order Butterworth bandpass at 0.5--45\,Hz, and notch filters at 50 and 60\,Hz. \emph{(ii)} BrainOmni~\citep{xiao2025brainomni} uses its own pipeline: resampling to 256\,Hz, bandpass at 0.1--96\,Hz, and the same notch filters as in \emph{(i)}.

\section{Robustness Perturbations}
\label{app:perturbations}

We apply all perturbations at test time only. Each model is trained once on clean data and the same fold checkpoint is evaluated under every perturbation condition. This mimics the deployment case where data quality drops at inference but retraining is not possible. We test four families of perturbation: additive white and pink noise, random channel dropout, region-based channel dropout, and region-specific noise injection. Noise samples and dropout masks are shared across all models, so every model sees the same perturbed inputs at every level.

\subsection{Additive Noise}

\begin{figure}[!h]
\centering
\includegraphics[width=\textwidth]{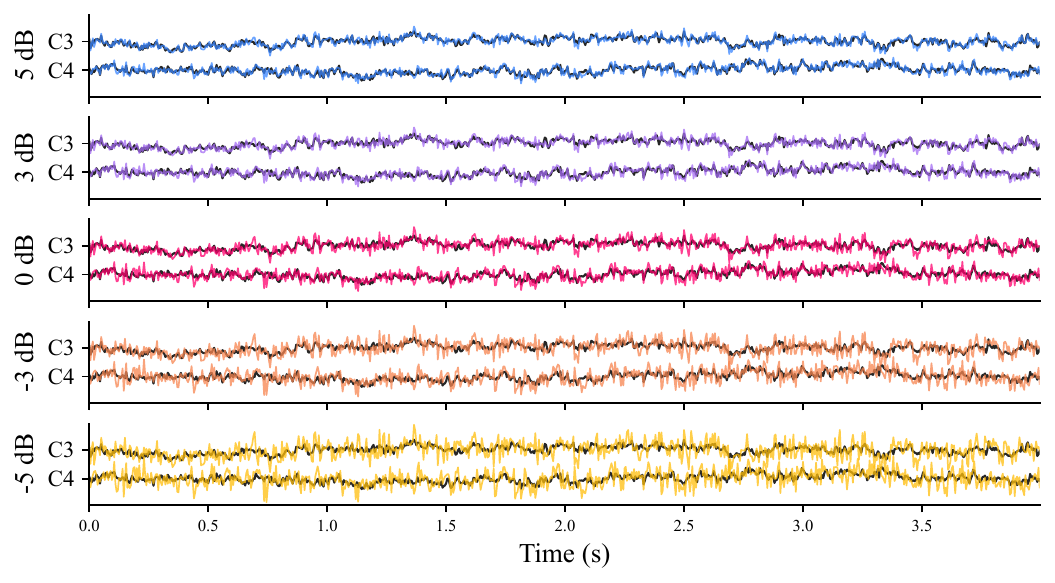}
\caption{White noise examples of two channels at five example SNR levels applied at test-time to a single EEG trial.}
\label{fig:noise_examples}
\end{figure}

\textbf{White noise} is sampled i.i.d.\ from $\mathcal{N}(0,1)$ independently per channel and time point (illustrated in Figure \ref{fig:noise_examples}). \textbf{Pink noise} ($1/f$) is generated by applying the fast Fourier transform (FFT) to white noise along the time axis, scaling the magnitude spectrum by $1/\sqrt{f}$, and applying the inverse FFT. Both noise types are filtered through the same bandpass and notch pipeline used for the clean data of each model, ensuring that the perturbation falls within the recording bandwidth. The baseline recordings already contain physiological and instrumental noise, so the reported SNR values are relative to the provided data rather than absolute. However, since all models are evaluated on the same data, the resulting comparisons reflect relative differences in robustness.

Pink noise is intended to approximate the spectral profile of biological EEG noise, but our implementation generates noise independently across channels. Real physiological noise exhibits spatial correlations due to volume conduction through the skull and scalp. This simplification means our pink noise perturbation might underestimate the disruptive effect of spatially correlated biological noise. For each trial, we compute the signal power as the mean squared amplitude after centring:
\begin{equation}
P_s = \frac{1}{C \cdot T} \sum_{c,t} (X_{c,t} - \bar{X}_c)^2
\end{equation}
where $C$ is the number of channels and $T$ the number of time points. The target noise power is $P_n = P_s / 10^{(\text{SNR}_{\text{dB}} / 10)}$. The raw noise is normalised to zero mean and unit variance per channel, then scaled by $\sqrt{P_n}$:
\begin{equation}
X_{\text{noisy}} = X_{\text{clean}} + \sqrt{P_n} \cdot \hat{n}
\end{equation}

\textbf{SNR levels tested:} $10$, $5$, $0$, $-3$, $-5$, $-15$ (white noise only) dB. At $10$\,dB the signal is clearly dominant; at $0$\,dB signal and noise have equal power; at $-5$\,dB the noise is approximately three times stronger than the signal.

\subsection{Random Channel Dropout}

\begin{table}[h]
\centering
\caption{Channels zero-padded at each per-channel dropout probability, as recorded by the corruption runs. Each benchmark draws its own mask; benchmarks sharing one are listed together. One mask is drawn per dataset and probability and shared across trials and folds. Sleep-EDF, Sleep-HMC are omitted.}
\label{tab:dropped_channels}
\scriptsize
\setlength{\tabcolsep}{3pt}
\resizebox{\textwidth}{!}{%
\begin{tabular}{llcp{10.5cm}}
\toprule
Dataset & $p$ & $n$ & Dropped channels \\
\midrule
\multirow{3}{*}{Movement-HG (78\,ch)}
  & 0.10 & 9 & FPZ, F7, FC6, T7, CP1, CPZ, P6, FT8, PO7 \\
  & 0.30 & 22 & FP2, FPZ, F7, FC6, T7, CZ, M2, CP1, CP2, AF7, C6, CPZ, P1, P6, PO3, FT7, FT8, TP8, PO7, FT9, TPP9H, PO10 \\
  & 0.50 & 38 & FP2, FPZ, F7, FC6, T7, CZ, M2, CP5, CP1, CP2, PZ, O2, AF7, AF8, F5, F6, FCZ, C5, C1, C6, CPZ, P1, P6, PO5, PO3, PO4, FT7, FT8, TP7, TP8, PO7, FT9, TPP9H, PO9, PO10, AFZ, FTT9H, FTT10H \\
\midrule
\multirow{3}{*}{Motor-Imagery-KU (62\,ch)}
  & 0.10 & 8 & AF7, AF3, F8, FT9, FT10, PO9, O1, FTT10H \\
  & 0.30 & 18 & FP2, AF7, AF3, F8, FT9, FC3, FC4, FT10, T7, CP5, P4, PO9, PO4, O1, O2, TPP9H, FTT10H, TPP10H \\
  & 0.50 & 30 & FP2, AF7, AF3, F8, FT9, FC3, FC4, FC6, FT10, T7, CZ, TP7, CP5, CPZ, CP2, TP8, P7, P1, PZ, P4, PO9, PO4, O1, OZ, O2, FTT9H, TPP9H, FTT10H, TPP8H, TPP10H \\
\midrule
\multirow{3}{*}{ERP (62\,ch)}
  & 0.10 & 8 & F7, F3, T7, CZ, CP6, TPP9H, TPP10H, AF8 \\
  & 0.30 & 18 & FP2, F7, F3, T7, CZ, T8, CP1, CP6, TP10, FC3, TTP7H, TPP9H, TPP8H, TPP10H, F10, AF4, AF8, PO4 \\
  & 0.50 & 30 & FP2, F7, F3, T7, CZ, T8, CP1, CP2, CP6, TP10, P4, PO10, FC3, C1, C2, CPZ, P1, POZ, FT9, TTP7H, TPP9H, TPP8H, TPP10H, F9, F10, AF7, AF4, AF8, PO3, PO4 \\
\midrule
\multirow{3}{*}{\shortstack[l]{Movement-PN\\Motor-Imagery-PN\\Eyes\\(64\,ch)}}
  & 0.10 & 9 & FC1, FCZ, C2, C6, CP6, P3, P6, PO8, O2 \\
  & 0.30 & 19 & FC3, FC1, FCZ, C2, C6, CP3, CP2, CP6, FP1, F1, P7, P3, P2, P6, PO7, PO4, PO8, OZ, O2 \\
  & 0.50 & 31 & FC3, FC1, FCZ, C2, C6, CP3, CP2, CP4, CP6, FP1, AF3, F3, F1, F4, F6, FT8, T8, T10, TP7, P7, P3, P2, P6, P8, PO7, PO3, PO4, PO8, O1, OZ, O2 \\
\midrule
\multirow{3}{*}{Memory (63\,ch)}
  & 0.10 & 9 & F3, F7, CP1, P3, TP10, PO4, CP4, F6, F2 \\
  & 0.30 & 19 & FZ, F3, F7, CP1, P3, O1, P4, TP10, CP6, AF7, PO3, PO4, P2, CP4, C6, FT8, F6, AF4, F2 \\
  & 0.50 & 31 & FZ, F3, F7, CP1, P3, O1, P4, P8, TP10, CP6, T8, FP2, AF7, F1, F5, C1, TP7, P1, P5, PO3, PO4, P2, CP4, TP8, C6, C2, FT8, F6, AF8, AF4, F2 \\
\midrule
\multirow{3}{*}{Seizure (19\,ch)}
  & 0.10 & 4 & C3, P3, FP2, C4 \\
  & 0.30 & 7 & F3, C3, P3, FP2, C4, O2, T6 \\
  & 0.50 & 7 & F3, C3, P3, FP2, C4, O2, T6 \\ 
\bottomrule
\end{tabular}}
\end{table}

Each channel is independently selected for zero-padding with probability $p \in \{0.10, 0.30, 0.50\}$ (Bernoulli dropout). Here, $p$ is a per-channel dropout probability, not a fixed fraction of channels. The resulting dataset-level mask is applied to all trials and folds. Zero-padding preserves the original channel dimension. We omit Sleep-EDF and Sleep-HMC from all random dropout figures and tables because of their limited channel counts.

In addition, we also test \textit{true channel dropout}, where dropped channels are removed from the input rather than zero-padded, for the variable-channel models: REVE, NeuroRVQ, LaBraM, and BrainOmni. We define a model as variable-channel if its forward pass conditions on channel identity (via learned channel embeddings or 3D electrode coordinates) and accepts a subset of the training channels at inference without retraining or padding. These results are reported in Appendix~\ref{app:true_dropout}.

Table~\ref{tab:dropped_channels} lists the exact channels dropped at each dropout probability for the eight included datasets. The three PhysioNet benchmarks (Eyes, Motor-Imagery-PN, Movement-PN) share the same 64-channel montage and therefore the same dropout mask.

\subsection{Region-based Channel Dropout}

\begin{figure}[!h]
\centering
\includegraphics[width=\textwidth]{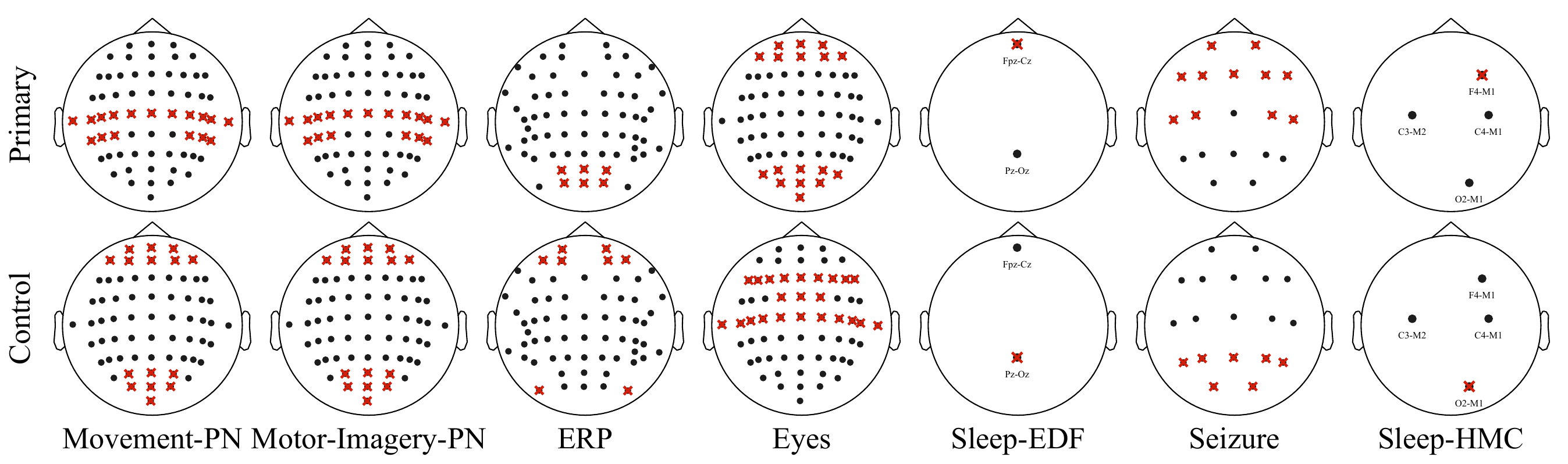}
\caption{Primary and control channels for region-based dropout, shown in red for seven tasks. Movement-HG and Movement-PN share region assignments, as do Motor-Imagery-KU and Motor-Imagery-PN.}
\label{fig:region_dropout_examples}
\end{figure}

We group channels into anatomical regions defined by the international 10--20 system (extended to 10--5 positions where applicable). For each task we assign two region sets: a primary set covering brain regions known to generate the task-relevant signal, and a control set used for comparison outside the primary region (Figure~\ref{fig:region_dropout_examples}). Dropping a region zeroes out all channels in that set for all models, preserving the original channel dimension. For the models that accept variable-length channel sets (REVE, NeuroRVQ, LaBraM, and BrainOmni), we also test \textit{true removal of the dropped region} and report these results in Appendix~\ref{app:robustness_region}. Table~\ref{tab:regions} lists the region assignments per task.

Each primary assignment maps the scalp regions that carry the dominant neural source for the task. For motor execution and motor imagery, mu and beta event-related de-synchronisation localises over the contralateral sensorimotor cortex, so central and centroparietal sites form the primary set in both tasks~\citep{pfurtscheller2001motorimagery, neuper2006sensorimotor}. For ERP, \citet{lee2019openbmi} report posterior-occipital sites as carrying the target P300 response, so we mark the posterior middle group as primary, capturing the occipital component of the P300, and treat anterior and lateral posterior sites as control. Posterior alpha rhythms dominate the eyes-closed condition, and the transition to eyes-open produces band-specific topographic changes over both anterior sites (delta and beta) and posterior sites (theta and beta)~\citep{barry2007eyes}. The eyes-open condition also adds oculomotor activity (blinks and saccades) that loads onto the anterior electrodes, so we mark both region groups as primary. For both sleep datasets, the frontal predominance of slow waves motivates using the frontal derivation as primary and the posterior derivation as control~\citep{werth1996sleeptopography}. However, Sleep-EDF provides only two bipolar derivations~\citep{kemp2000sleepedf}, and Sleep-HMC only four, so a single region is defined by a single derivation. We therefore exclude both sleep datasets from the region-based averages, but report their per-dataset results for completeness. For Seizure, Siena's temporal and frontal seizure localisations motivate our primary region~\citep{siena2020}, although this shared assignment is coarse, and control channels may also carry seizure information.

We define 18 region groups using the international 10--20 system extended to 10--5 positions:
\begin{enumerate}
    \item \textbf{Anterior}: Fp1, Fpz, Fp2, AF3, AF4, AF7, AF8, AFz.
\item \textbf{Frontal} (Left/Middle/Right): F7, F5, F3, F9 / F1, Fz, F2, FC1, FCz, FC2 / F4, F6, F8, F10.
\item \textbf{Frontotemporal} (Left/Right): FT7, FT9, FC5, FC3, FTT9h / FC4, FC6, FT8, FT10, FTT10h.
\item \textbf{Central} (Left/Middle/Right): T7, T9, C5, C3 / C1, Cz, C2 / C4, C6, T8, T10.
\item \textbf{Centroparietal} (Left/Middle/Right): TP7, TP9, CP5, CP3, TTP7h, TPP9h / CP1, CPz, CP2 / CP4, CP6, TP8, TP10, TPP8h, TPP10h.
\item \textbf{Parietal} (Left/Middle/Right): P7, P5, P3, P9 / P1, Pz, P2 / P4, P6, P8, P10.
\item \textbf{Posterior} (Left/Middle/Right): PO7, PO5, PO9, CB1 / PO3, POz, PO4, O1, Oz, O2, Iz / PO6, PO8, PO10, CB2.
\end{enumerate}

\begin{table}[h]
\caption{Region assignments for region-based perturbations. Primary regions are expected to contain task-relevant features; control regions provide a comparison.}
\label{tab:regions}
\centering
\small
\begin{tabular}{p{3cm}p{4.8cm}p{4.8cm}}
\toprule
Task & Primary & Control \\
\midrule
Movement & Central L/M/R, Centroparietal L/R & Posterior Middle, Anterior \\
Motor imagery & Central L/M/R, Centroparietal L/R & Posterior Middle, Anterior  \\
Visual P300  & Posterior Middle & Anterior,  Posterior L/R \\
Working memory & Frontal L/M/R & Posterior Middle, Central L/R \\
Eyes open/closed & Anterior, Posterior L/M/R& Frontal L/M/R, Central L/M/R \\
Sleep staging & EDF$^\dagger$: Fpz-Cz (frontal)\newline HMC: F4 (frontal) & EDF$^\dagger$: Pz-Oz (posterior)\newline HMC: O2 (posterior) \\
Seizure detection & Anterior, Frontal L/M/R, Central L/R & Parietal L/M/R, Posterior Middle \\
\bottomrule
\multicolumn{3}{l}{\footnotesize $^\dagger$Sleep-EDF channels are bipolar derivations.} \\
\end{tabular}
\end{table}

\subsection{Region Signal Substitution}
\label{app:substitution}

This condition removes the signal in a region without setting it to zero. For each trial we replace every channel in the primary region with white noise. We filter the noise with the same settings as the dataset, normalise it to zero mean and unit variance, then scale it so its root mean square matches the median of the intact channels. The substituted channels keep a plausible amplitude and carry none of the original signal. We use this condition for attribution only, not for accuracy, because Zero padding is a poor control for Input $\times$ Gradient and AttnLRP.

\subsection{Region Noise Injection}

Instead of removing channels, white noise at a given SNR is injected into channels belonging to a single region while all other channels remain non-perturbed. This is a softer perturbation than region dropout, so models can still partially extract features from noisy channels. It simulates degraded electrodes, the most common clinical scenario. SNR levels tested: $5$\,dB and $-3$\,dB, for each of primary and control regions. The same region assignments as Table~\ref{tab:regions} are used.

\section{Statistical Significance Testing}
\label{app:stats}

We compare models on 10-fold leave-many-subjects-out cross-validation (LOMSO-CV) with identical subject splits across models. Although the training sets overlap across folds, the comparison within each fold is paired because all models are evaluated on exactly the same held-out subjects. A two-sided Wilcoxon signed-rank test was therefore applied to compare models to our baseline. When we average over datasets, we pair over every dataset and fold pair. We adjust the $p$-values for multiple comparisons with the Holm-Bonferroni method, at significance $p = 0.05$.

\newpage
\section{Clean Performance Tables}
\label{app:clean_results_full}

Tables~\ref{tab:clean_full_ft_all} and~\ref{tab:clean_head_all} report test balanced accuracy (mean $\pm$ standard deviation across ten folds) for all models under full fine-tuning and head-only adaptation respectively. 
Models are ranked by average balanced accuracy. EEGNet is included in both tables as the supervised baseline.

\begin{table}[h]
\centering
\caption{Clean balanced accuracy (full fine-tuned), mean $\pm$ std over 10 folds, ranked by Average. Best per column in \textbf{bold}, second best \underline{underlined}. Average equally weights the 10 datasets (Movement-HG, Movement-PN, Motor-Imagery-KU, Motor-Imagery-PN, ERP, Eyes, Memory, Sleep-EDF, Seizure, Sleep-HMC). $^{\ast}$ on the Average marks a model significantly better than EEGNet, $^{\nabla}$ significantly worse (two-sided Wilcoxon signed-rank over the shared folds, Holm-corrected).}
\label{tab:clean_full_ft_all}
\scriptsize
\setlength{\tabcolsep}{3pt}
\resizebox{\textwidth}{!}{%
\begin{tabular}{rlccccccccccc}
\toprule
\# & Model & Movement-HG & Movement-PN & Motor-Imagery-KU & Motor-Imagery-PN & ERP & Eyes & Memory & Sleep-EDF & Seizure & Sleep-HMC & Average \\
\midrule
1 & NeuroRVQ & \underline{$0.689{\pm}0.089$} & \underline{$0.559{\pm}0.041$} & \underline{$0.809{\pm}0.035$} & \underline{$0.538{\pm}0.053$} & {\boldmath $0.838{\pm}0.038$} & {\boldmath $0.861{\pm}0.025$} & $0.579{\pm}0.020$ & {\boldmath $0.684{\pm}0.033$} & \underline{$0.783{\pm}0.085$} & {\boldmath $0.735{\pm}0.024$} & {\boldmath $0.707$$^{\ast}$} \\
2 & REVE & {\boldmath $0.694{\pm}0.105$} & $0.510{\pm}0.028$ & {\boldmath $0.829{\pm}0.038$} & $0.496{\pm}0.025$ & $0.826{\pm}0.043$ & $0.799{\pm}0.034$ & \underline{$0.593{\pm}0.032$} & $0.629{\pm}0.035$ & {\boldmath $0.794{\pm}0.111$} & $0.709{\pm}0.017$ & \underline{$0.688$$^{\ast}$} \\
3 & CBraMod & $0.594{\pm}0.079$ & $0.547{\pm}0.029$ & $0.752{\pm}0.040$ & $0.501{\pm}0.032$ & $0.783{\pm}0.045$ & $0.821{\pm}0.040$ & {\boldmath $0.603{\pm}0.021$} & \underline{$0.634{\pm}0.042$} & $0.760{\pm}0.121$ & $0.725{\pm}0.018$ & $0.672$$^{\ast}$ \\
4 & BrainOmni & $0.585{\pm}0.058$ & {\boldmath $0.625{\pm}0.036$} & $0.750{\pm}0.038$ & {\boldmath $0.615{\pm}0.053$} & $0.723{\pm}0.041$ & \underline{$0.852{\pm}0.022$} & $0.518{\pm}0.023$ & $0.576{\pm}0.055$ & $0.667{\pm}0.121$ & \underline{$0.727{\pm}0.019$} & $0.664$ \\
5 & LaBraM & $0.620{\pm}0.079$ & $0.517{\pm}0.044$ & $0.775{\pm}0.041$ & $0.526{\pm}0.059$ & \underline{$0.832{\pm}0.036$} & $0.833{\pm}0.029$ & $0.534{\pm}0.041$ & $0.606{\pm}0.046$ & $0.568{\pm}0.123$ & $0.702{\pm}0.018$ & $0.651$ \\
6 & EEGNet & $0.624{\pm}0.100$ & \underline{$0.559{\pm}0.038$} & $0.786{\pm}0.034$ & $0.528{\pm}0.086$ & $0.796{\pm}0.040$ & $0.758{\pm}0.024$ & $0.512{\pm}0.005$ & $0.590{\pm}0.029$ & $0.708{\pm}0.155$ & $0.605{\pm}0.040$ & $0.647$ \\
7 & BIOT & $0.322{\pm}0.050$ & $0.258{\pm}0.008$ & $0.501{\pm}0.005$ & $0.253{\pm}0.018$ & $0.500{\pm}0.000$ & $0.753{\pm}0.065$ & $0.519{\pm}0.016$ & $0.167{\pm}0.000$ & $0.623{\pm}0.089$ & $0.695{\pm}0.028$ & $0.459$$^{\nabla}$ \\
\bottomrule
\end{tabular}}
\end{table}

\begin{table}[h]
\centering
\caption{Clean balanced accuracy (head-only), mean $\pm$ std over 10 folds, ranked by Average. Best per column in \textbf{bold}, second best \underline{underlined}. EEGNet has no frozen-backbone variant - it trains from scratch, so its row repeats its supervised score as the reference the foundation models are read against. Average equally weights the 10 datasets (Movement-HG, Movement-PN, Motor-Imagery-KU, Motor-Imagery-PN, ERP, Eyes, Memory, Sleep-EDF, Seizure, Sleep-HMC). $^{\ast}$ on the Average marks a model significantly better than EEGNet, $^{\nabla}$ significantly worse (two-sided Wilcoxon signed-rank over the shared folds, Holm-corrected).}
\label{tab:clean_head_all}
\scriptsize
\setlength{\tabcolsep}{3pt}
\resizebox{\textwidth}{!}{%
\begin{tabular}{rlccccccccccc}
\toprule
\# & Model & Movement-HG & Movement-PN & Motor-Imagery-KU & Motor-Imagery-PN & ERP & Eyes & Memory & Sleep-EDF & Seizure & Sleep-HMC & Average \\
\midrule
1 & EEGNet & {\boldmath $0.624{\pm}0.100$} & {\boldmath $0.559{\pm}0.038$} & {\boldmath $0.786{\pm}0.034$} & {\boldmath $0.528{\pm}0.086$} & {\boldmath $0.796{\pm}0.040$} & $0.758{\pm}0.024$ & $0.512{\pm}0.005$ & {\boldmath $0.590{\pm}0.029$} & \underline{$0.708{\pm}0.155$} & $0.605{\pm}0.040$ & {\boldmath $0.647$} \\
2 & REVE & \underline{$0.544{\pm}0.095$} & $0.458{\pm}0.027$ & \underline{$0.761{\pm}0.037$} & $0.449{\pm}0.018$ & \underline{$0.749{\pm}0.035$} & $0.790{\pm}0.043$ & {\boldmath $0.574{\pm}0.039$} & \underline{$0.589{\pm}0.036$} & {\boldmath $0.756{\pm}0.116$} & $0.656{\pm}0.024$ & \underline{$0.633$} \\
3 & BrainOmni & $0.481{\pm}0.050$ & \underline{$0.527{\pm}0.032$} & $0.658{\pm}0.029$ & \underline{$0.502{\pm}0.030$} & $0.525{\pm}0.010$ & {\boldmath $0.829{\pm}0.036$} & $0.541{\pm}0.026$ & $0.473{\pm}0.036$ & $0.606{\pm}0.126$ & {\boldmath $0.684{\pm}0.029$} & $0.583$$^{\nabla}$ \\
4 & CBraMod & $0.381{\pm}0.078$ & $0.322{\pm}0.037$ & $0.653{\pm}0.030$ & $0.284{\pm}0.040$ & $0.642{\pm}0.019$ & \underline{$0.809{\pm}0.053$} & \underline{$0.545{\pm}0.022$} & $0.563{\pm}0.055$ & $0.694{\pm}0.150$ & \underline{$0.677{\pm}0.023$} & $0.557$$^{\nabla}$ \\
5 & NeuroRVQ & $0.271{\pm}0.022$ & $0.252{\pm}0.003$ & $0.501{\pm}0.003$ & $0.252{\pm}0.003$ & $0.542{\pm}0.013$ & $0.736{\pm}0.038$ & $0.517{\pm}0.021$ & $0.474{\pm}0.026$ & $0.509{\pm}0.060$ & $0.455{\pm}0.027$ & $0.451$$^{\nabla}$ \\
6 & BIOT & $0.309{\pm}0.038$ & $0.250{\pm}0.012$ & $0.501{\pm}0.010$ & $0.256{\pm}0.016$ & $0.500{\pm}0.000$ & $0.679{\pm}0.135$ & $0.536{\pm}0.014$ & $0.167{\pm}0.000$ & $0.554{\pm}0.052$ & $0.624{\pm}0.021$ & $0.438$$^{\nabla}$ \\
7 & LaBraM & $0.300{\pm}0.047$ & $0.270{\pm}0.017$ & $0.504{\pm}0.008$ & $0.266{\pm}0.013$ & $0.500{\pm}0.000$ & $0.717{\pm}0.049$ & $0.500{\pm}0.000$ & $0.295{\pm}0.008$ & $0.497{\pm}0.016$ & $0.473{\pm}0.016$ & $0.432$$^{\nabla}$ \\
\bottomrule
\end{tabular}}
\end{table}

\newpage
\section{Pre-training versus Architecture}
\label{app:scratch}

Pre-training improves average performance for most models compared with training the same architecture from scratch under the same schedule (Table~\ref{tab:scratch_vs_pretrained}). The gain is small for CBraMod, while BIOT shows no benefit from pre-training on these tasks.

\begin{figure}[!h]
\centering
\includegraphics[width=\textwidth]{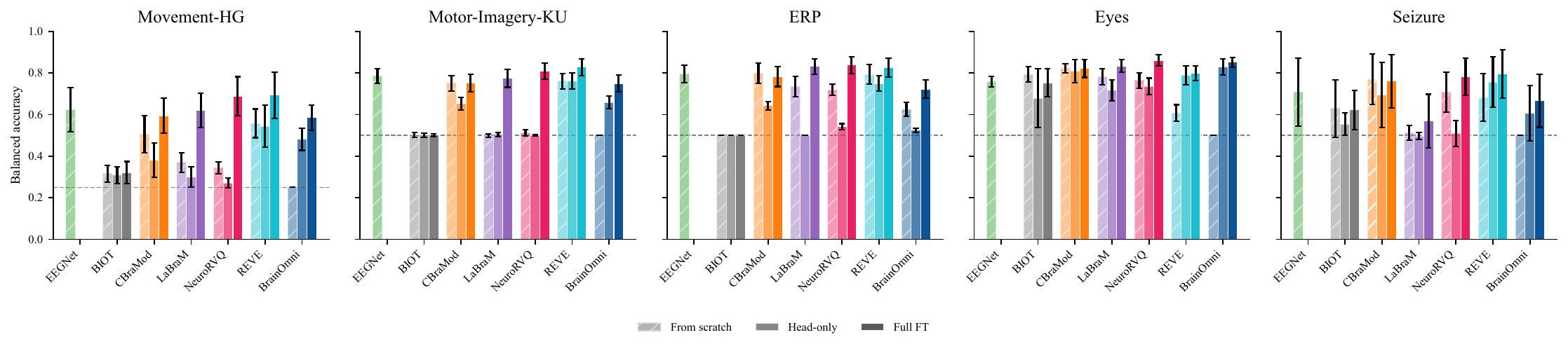}
\caption{From-scratch training (hatched bars), pre-trained frozen backbone (light bars) and full fine-tuning (saturated bars) on Movement-HG, Motor-Imagery-KU, ERP, Eyes and Seizure. Bars show mean balanced accuracy over 10 folds; error bars show $\pm 1$ standard deviation. The dashed horizontal line marks chance-level balanced accuracy.}
\label{fig:scratch}
\end{figure}

\begin{table}[h]
\centering
\caption{Pre-training versus architecture. Balanced accuracy for the same architecture trained from random initialisation, adapted with a frozen pre-trained backbone, and fully fine-tuned from the pre-trained checkpoint. Restricted to the 5 datasets with from-scratch runs (Movement-HG, Motor-Imagery-KU, ERP, Eyes, Seizure); Average weighs them equally. EEGNet is from-scratch. Best per column in \textbf{bold}, second best \underline{underlined}.}
\label{tab:scratch_vs_pretrained}
\scriptsize
\setlength{\tabcolsep}{3pt}
\begin{tabular*}{\textwidth}{@{\extracolsep{\fill}}llcccccc}
\toprule
Model & Regime & Movement-HG & Motor-Imagery-KU & ERP & Eyes & Seizure & Average \\
\midrule
NeuroRVQ & From scratch & $0.344$ & $0.511$ & $0.719$ & $0.765$ & $0.708$ & $0.609$ \\
 & Head-only & $0.271$ & $0.501$ & $0.542$ & $0.736$ & $0.509$ & $0.512$ \\
 & Full FT & \underline{$0.689$} & \underline{$0.809$} & {\boldmath $0.838$} & {\boldmath $0.861$} & \underline{$0.783$} & {\boldmath $0.796$} \\
\addlinespace
REVE & From scratch & $0.558$ & $0.761$ & $0.794$ & $0.607$ & $0.682$ & $0.680$ \\
 & Head-only & $0.544$ & $0.761$ & $0.749$ & $0.790$ & $0.756$ & $0.720$ \\
 & Full FT & {\boldmath $0.694$} & {\boldmath $0.829$} & $0.826$ & $0.799$ & {\boldmath $0.794$} & \underline{$0.788$} \\
\addlinespace
CBraMod & From scratch & $0.506$ & $0.751$ & $0.799$ & $0.824$ & $0.769$ & $0.730$ \\
 & Head-only & $0.381$ & $0.653$ & $0.642$ & $0.809$ & $0.694$ & $0.636$ \\
 & Full FT & $0.594$ & $0.752$ & $0.783$ & $0.821$ & $0.760$ & $0.742$ \\
\addlinespace
LaBraM & From scratch & $0.370$ & $0.498$ & $0.735$ & $0.782$ & $0.512$ & $0.580$ \\
 & Head-only & $0.300$ & $0.504$ & $0.500$ & $0.717$ & $0.497$ & $0.504$ \\
 & Full FT & $0.620$ & $0.775$ & \underline{$0.832$} & $0.833$ & $0.568$ & $0.726$ \\
\addlinespace
BrainOmni & From scratch & $0.250$ & $0.500$ & $0.625$ & $0.500$ & $0.500$ & $0.475$ \\
 & Head-only & $0.481$ & $0.658$ & $0.525$ & $0.829$ & $0.606$ & $0.620$ \\
 & Full FT & $0.585$ & $0.750$ & $0.723$ & \underline{$0.852$} & $0.667$ & $0.715$ \\
\addlinespace
BIOT & From scratch & $0.316$ & $0.502$ & $0.500$ & $0.793$ & $0.630$ & $0.548$ \\
 & Head-only & $0.309$ & $0.501$ & $0.500$ & $0.679$ & $0.554$ & $0.509$ \\
 & Full FT & $0.322$ & $0.501$ & $0.500$ & $0.753$ & $0.623$ & $0.540$ \\
\addlinespace
EEGNet & From scratch & $0.624$ & $0.786$ & $0.796$ & $0.758$ & $0.708$ & $0.735$ \\
\bottomrule
\end{tabular*}
\end{table}

\newpage
\section{Robustness Results}
\label{app:robustness}

\begin{figure}[!h]
\centering
\begin{subfigure}[t]{\textwidth}
\centering
\includegraphics[width=\textwidth]{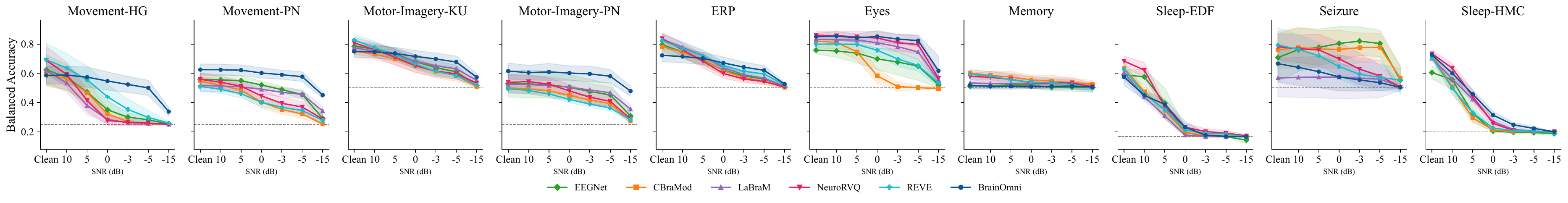}
\caption{Additive white noise.}
\label{fig:full_white_noise}
\end{subfigure}
\vspace{0.3em}

\centering
\begin{subfigure}[t]{\textwidth}
\centering
\includegraphics[width=\textwidth]{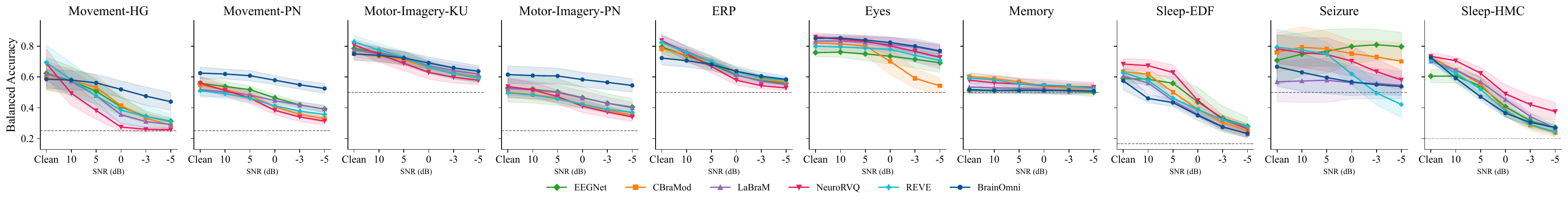}
\caption{Additive pink noise.}
\label{fig:full_pink_noise}
\end{subfigure}
\vspace{0.3em}

\begin{subfigure}[t]{\textwidth}
\centering
\includegraphics[width=\textwidth]{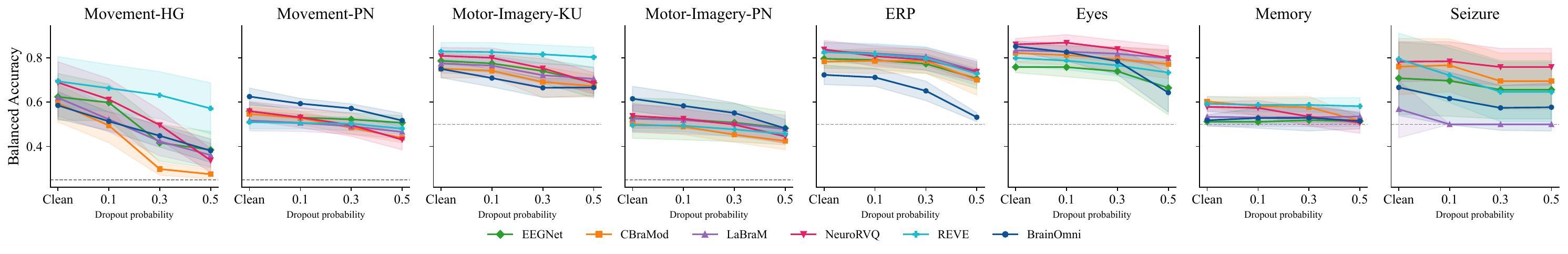}
\caption{Random channel dropout.}
\label{fig:full_random_dropout}
\end{subfigure}
\vspace{0.3em}
\begin{subfigure}[t]{\textwidth}
\centering
\includegraphics[width=\textwidth]{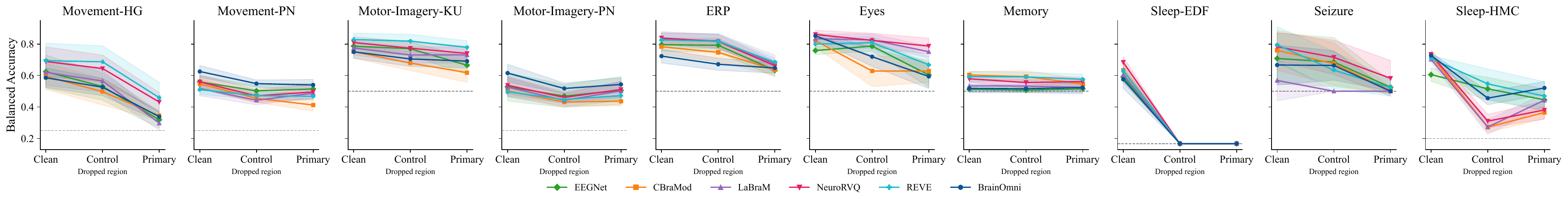}
\caption{Region dropout.}
\label{fig:full_region_dropout}
\end{subfigure}
\vspace{0.3em}
\begin{subfigure}[t]{\textwidth}
\centering
\includegraphics[width=\textwidth]{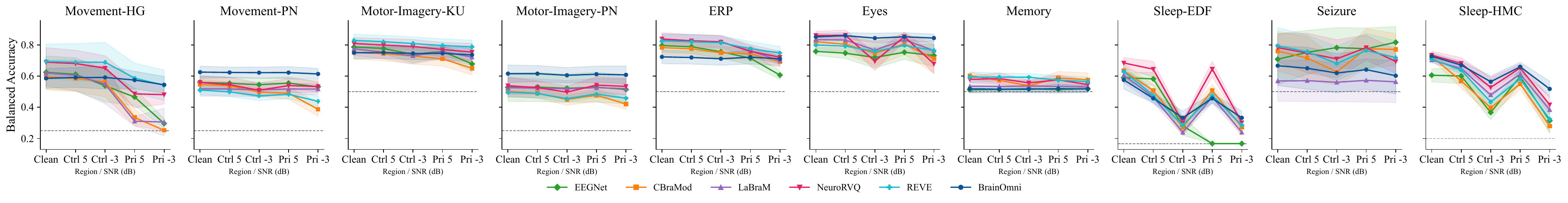}
\caption{Region noise injection.}
\label{fig:full_region_noise}
\end{subfigure}
\caption{Robustness evaluation under five perturbation types (full fine-tuned models): (a) Additive White Noise, (b) Additive Pink Noise, (c) Random Channel Dropout, (d) Region Dropout and (e) Region Noise Injection. Each line shows mean balanced accuracy over ten folds as a function of perturbation level; shaded bands show $\pm 1$ standard deviation across folds and dashed lines mark chance level. Noise and region panels cover ten datasets; random dropout panels cover eight.}
\end{figure}

\begin{figure}[!h]
\centering
\begin{subfigure}[t]{\textwidth}
\centering
\includegraphics[width=\textwidth]{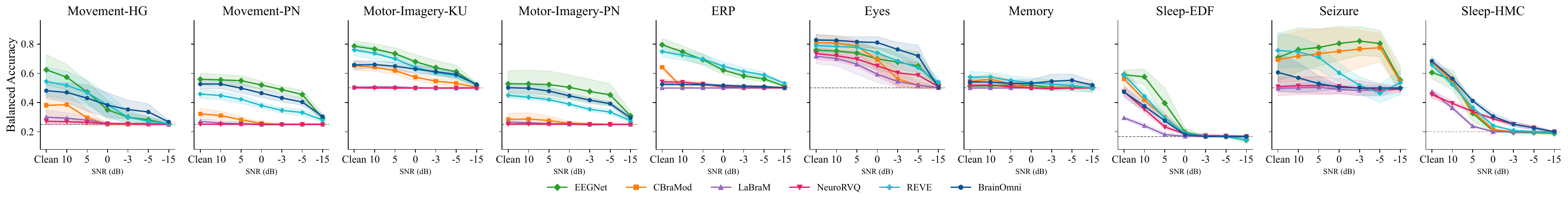}
\caption{Additive white noise degradation under head-only adaptation.}
\label{fig:head_white_noise}
\end{subfigure}
\vspace{0.3em}
\begin{subfigure}[t]{\textwidth}
\centering
\includegraphics[width=\textwidth]{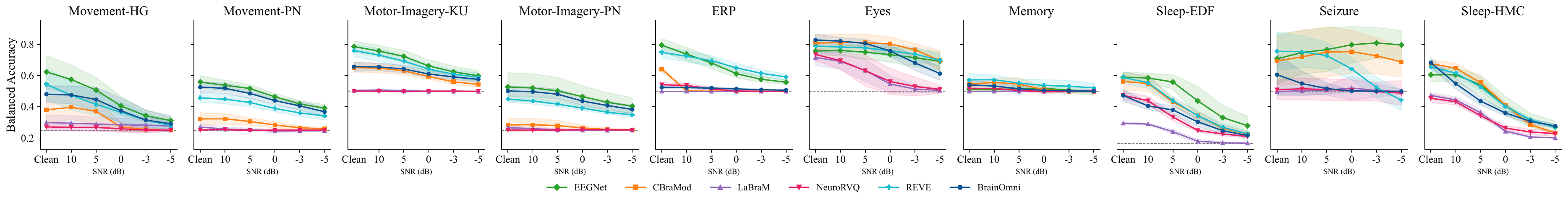}
\caption{Additive pink noise degradation under head-only adaptation.}
\label{fig:head_pink_noise}
\end{subfigure}
\vspace{0.3em}
\begin{subfigure}[t]{\textwidth}
\centering
\includegraphics[width=\textwidth]{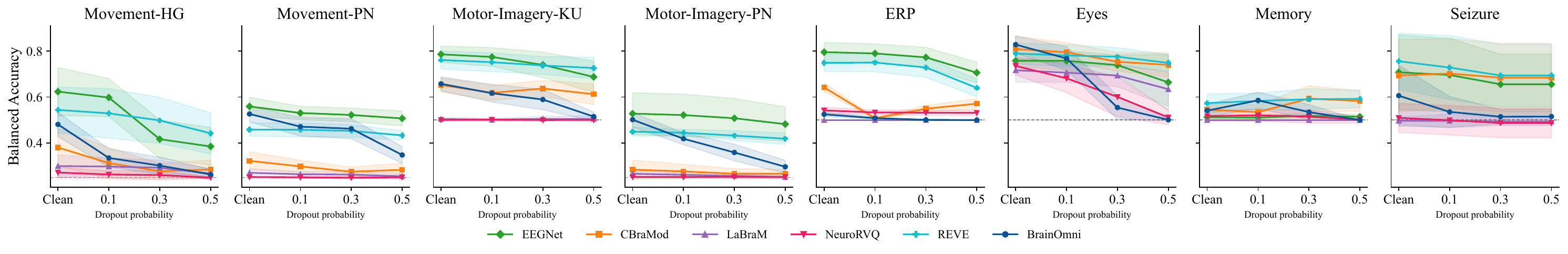}
\caption{Random channel dropout under head-only adaptation.}
\label{fig:head_dropout}
\end{subfigure}
\vspace{0.3em}
\begin{subfigure}[t]{\textwidth}
\centering
\includegraphics[width=\textwidth]{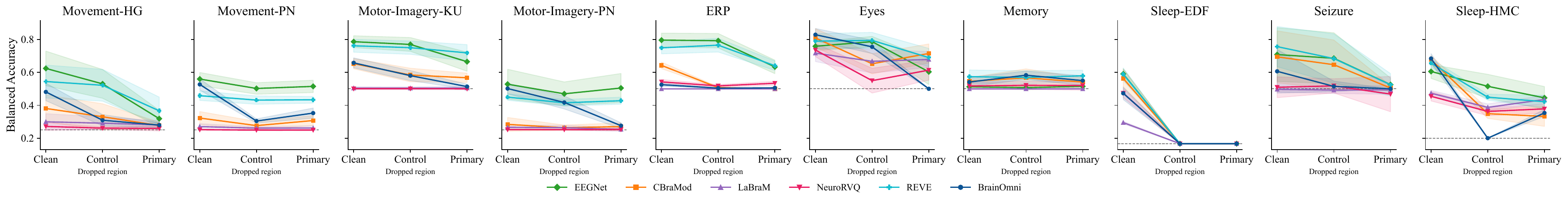}
\caption{Region dropout under head-only adaptation.}
\label{fig:head_region_dropout}
\end{subfigure}
\vspace{0.3em}
\begin{subfigure}[t]{\textwidth}
\centering
\includegraphics[width=\textwidth]{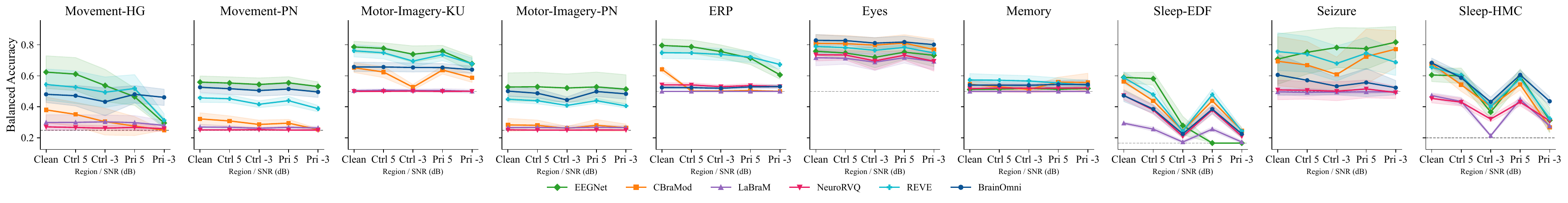}
\caption{Region noise injection under head-only adaptation.}
\label{fig:head_region_noise}
\end{subfigure}
\caption{Robustness evaluation under five perturbation types (head-only adaptation): (a) Additive White Noise, (b) Additive Pink Noise, (c) Random Channel Dropout, (d) Region Dropout and (e) Region Noise Injection. Each line shows mean balanced accuracy over ten folds as a function of perturbation level; shaded bands show $\pm 1$ standard deviation across folds and dashed lines mark chance level. Noise and region panels cover ten datasets; random dropout panels cover eight.}
\label{fig:robustness}
\end{figure}

Figures \ref{fig:full_white_noise}--\ref{fig:full_region_noise} show robustness curves under full fine-tuning. Figures~\ref{fig:head_white_noise}--\ref{fig:head_region_noise} show robustness curves under head-only adaptation.

\newpage
\subsection{Additive Noise}
\label{app:robustness_noise}

Tables~\ref{tab:white_noise_per_bench} and~\ref{tab:pink_noise_per_bench} report per-benchmark changes in percentage points from clean at $-5$\,dB white and pink noise respectively. Models are ranked by average degradation across eight datasets, excluding Sleep-EDF and Sleep-HMC.

White-noise degradation varies across tasks and models. At $-5$\,dB, several models approach chance on Movement-HG and ERP, while most models remain above chance on Eyes and all do on Motor-Imagery-KU. Both sleep datasets approach chance under white noise at this level, but remain above chance under pink noise. BrainOmni appears to be the most noise robust model, remaining above chance even at $-15$\,dB for certain tasks. It also remains the most noise robust model with both sleep datasets included ($-14.2$ vs.\ $-17.9$ to $-25.9$ percentage points for all other models at $-5$\,dB additive white noise). Several architectural choices may contribute. BrainOmni compresses input channels to 16 latent source variables before the transformer; the original paper's ablation supports this channel-compression hypothesis, with Table 8 in \citep{xiao2025brainomni} showing that replacing the cross-attention compression with electrode-level self-attention degrades clean balanced accuracy on every dataset they test. They attribute the gain to elimination of redundant information between adjacent electrodes. The ablation considers clean inputs without perturbation, but the same bottleneck may underlie the noise robustness shown in our results. Pre-training on both EEG and MEG may also contribute. The quantiser ablation (Appendix~\ref{app:vq_skip}) shows that BrainOmni retains its noise-robustness advantage without quantisation. Latent compression and multimodal pre-training thus remain possible explanations, and separating them requires matched pre-training variants.

\vspace{-0.1in}
\begin{table}[h]
\centering
\caption{Per-benchmark degradation under White $-5$\,dB (full fine-tuning). $\Delta$ percentage points from clean, paired per fold. Ranked by Average. Best per column in \textbf{bold}, second best \underline{underlined}. Average equally weights the 8 datasets (Movement-HG, Movement-PN, Motor-Imagery-KU, Motor-Imagery-PN, ERP, Eyes, Memory, Seizure); Sleep-EDF, Sleep-HMC excluded from the average. $^{\ast}$ on the Average marks a model significantly better than EEGNet, $^{\nabla}$ significantly worse (two-sided Wilcoxon signed-rank over the shared folds, Holm-corrected).}
\label{tab:white_noise_per_bench}
\scriptsize
\setlength{\tabcolsep}{3pt}
\resizebox{\textwidth}{!}{%
\begin{tabular}{rlccccccccccc}
\toprule
\# & Model & Movement-HG & Movement-PN & Motor-Imagery-KU & Motor-Imagery-PN & ERP & Eyes & Memory & Sleep-EDF & Seizure & Sleep-HMC & Average \\
\midrule
1 & BrainOmni & {\boldmath $-8.5$} & {\boldmath $-4.8$} & {\boldmath $-7.2$} & {\boldmath $-3.5$} & {\boldmath $-10.2$} & {\boldmath $-2.7$} & {\boldmath $-0.5$} & {\boldmath $-40.7$} & $-13.1$ & $-50.5$ & {\boldmath $-6.3$$^{\ast}$} \\
2 & EEGNet & $-34.2$ & $-10.5$ & $-17.6$ & $-7.6$ & $-23.5$ & $-10.9$ & $-0.7$ & \underline{$-42.4$} & {\boldmath $+9.6$} & {\boldmath $-41.1$} & \underline{$-11.9$} \\
3 & LaBraM & $-36.3$ & \underline{$-6.1$} & \underline{$-14.3$} & \underline{$-5.9$} & $-26.5$ & $-8.6$ & \underline{$-0.6$} & $-43.9$ & $-0.9$ & \underline{$-49.9$} & $-12.4$ \\
4 & CBraMod & \underline{$-33.9$} & $-22.6$ & $-16.6$ & $-10.9$ & $-23.7$ & $-31.9$ & $-6.7$ & $-46.2$ & \underline{$+1.9$} & $-52.5$ & $-18.1$$^{\nabla}$ \\
5 & NeuroRVQ & $-43.0$ & $-19.0$ & $-21.3$ & $-13.0$ & $-29.5$ & \underline{$-6.3$} & $-4.5$ & $-49.3$ & $-20.2$ & $-53.4$ & $-19.6$$^{\nabla}$ \\
6 & REVE & $-39.6$ & $-16.5$ & $-24.4$ & $-13.4$ & \underline{$-23.0$} & $-14.5$ & $-6.9$ & $-44.5$ & $-22.2$ & $-50.5$ & $-20.1$$^{\nabla}$ \\
\bottomrule
\end{tabular}}
\end{table}

\vspace{-0.1in}
\begin{table}[h]
\centering
\caption{Per-benchmark degradation under Pink $-5$\,dB (full fine-tuning). $\Delta$ percentage points from clean, paired per fold. Ranked by Average. Best per column in \textbf{bold}, second best \underline{underlined}. Average equally weights the 8 datasets (Movement-HG, Movement-PN, Motor-Imagery-KU, Motor-Imagery-PN, ERP, Eyes, Memory, Seizure); Sleep-EDF, Sleep-HMC excluded from the average. $^{\ast}$ on the Average marks a model significantly better than EEGNet, $^{\nabla}$ significantly worse (two-sided Wilcoxon signed-rank over the shared folds, Holm-corrected).}
\label{tab:pink_noise_per_bench}
\scriptsize
\setlength{\tabcolsep}{3pt}
\resizebox{\textwidth}{!}{%
\begin{tabular}{rlccccccccccc}
\toprule
\# & Model & Movement-HG & Movement-PN & Motor-Imagery-KU & Motor-Imagery-PN & ERP & Eyes & Memory & Sleep-EDF & Seizure & Sleep-HMC & Average \\
\midrule
1 & BrainOmni & {\boldmath $-14.6$} & {\boldmath $-10.0$} & {\boldmath $-11.2$} & {\boldmath $-7.0$} & {\boldmath $-13.9$} & $-8.2$ & {\boldmath $-1.0$} & \underline{$-34.3$} & $-12.8$ & $-45.6$ & {\boldmath $-9.8$$^{\ast}$} \\
2 & EEGNet & $-31.2$ & $-16.8$ & $-18.7$ & \underline{$-12.4$} & $-23.8$ & {\boldmath $-6.6$} & \underline{$-1.2$} & {\boldmath $-31.1$} & {\boldmath $+8.9$} & {\boldmath $-33.4$} & \underline{$-12.7$} \\
3 & LaBraM & $-32.9$ & \underline{$-12.6$} & \underline{$-15.6$} & $-12.6$ & $-28.1$ & \underline{$-6.7$} & $-2.5$ & $-37.7$ & \underline{$-2.2$} & $-43.6$ & $-14.2$ \\
4 & CBraMod & \underline{$-30.8$} & $-21.9$ & \underline{$-15.6$} & $-15.0$ & \underline{$-21.7$} & $-27.9$ & $-7.5$ & $-38.4$ & $-5.8$ & $-48.7$ & $-18.3$$^{\nabla}$ \\
5 & REVE & $-38.6$ & $-15.4$ & $-23.7$ & $-12.6$ & $-24.8$ & $-9.0$ & $-6.8$ & $-35.3$ & $-37.3$ & $-46.7$ & $-21.0$$^{\nabla}$ \\
6 & NeuroRVQ & $-43.1$ & $-24.6$ & $-23.0$ & $-19.8$ & $-31.0$ & $-13.3$ & $-4.5$ & $-42.0$ & $-20.2$ & \underline{$-35.9$} & $-22.4$$^{\nabla}$ \\
\bottomrule
\end{tabular}}
\end{table}

\subsection{Random Channel Dropout}
\label{app:robustness_dropout}

\subsubsection{Zero Padded Channels}

Table~\ref{tab:dropout_per_bench} reports per-benchmark degradation at $p{=}0.50$ random channel dropout. We omit Sleep-EDF and Sleep-HMC from all random channel dropout figures and tables. Channel dropout is model-limited (with some task-specific variation), suggesting that architectural choices, particularly how models encode or handle missing information, play a role. REVE has the smallest average loss under zero-padded random dropout. A possible contributing factor is REVE's pre-training, which drops 10\% of channels on top of token masking and could plausibly give the model exposure to missing-channel scenarios.

\vspace{-0.1in}
\begin{table}[h]
\centering
\caption{Per-benchmark degradation under Rand.\ drop $p{=}0.50$ (full fine-tuning). $\Delta$ percentage points from clean, paired per fold. Ranked by Average. Best per column in \textbf{bold}, second best \underline{underlined}. Average equally weights the 8 datasets (Movement-HG, Movement-PN, Motor-Imagery-KU, Motor-Imagery-PN, ERP, Eyes, Memory, Seizure); Sleep-EDF, Sleep-HMC excluded from the average. $^{\ast}$ on the Average marks a model significantly better than EEGNet, $^{\nabla}$ significantly worse (two-sided Wilcoxon signed-rank over the shared folds, Holm-corrected).}
\label{tab:dropout_per_bench}
\scriptsize
\setlength{\tabcolsep}{3pt}
\resizebox{\textwidth}{!}{%
\begin{tabular}{rlccccccccc}
\toprule
\# & Model & Movement-HG & Movement-PN & Motor-Imagery-KU & Motor-Imagery-PN & ERP & Eyes & Memory & Seizure & Average \\
\midrule
1 & REVE & {\boldmath $-12.3$} & {\boldmath $-3.0$} & {\boldmath $-2.6$} & {\boldmath $-4.0$} & $-9.7$ & $-6.6$ & $-1.1$ & $-14.8$ & {\boldmath $-6.8$} \\
2 & LaBraM & $-25.9$ & \underline{$-5.1$} & \underline{$-6.9$} & $-4.7$ & $-9.1$ & {\boldmath $-3.5$} & {\boldmath $+0.2$} & $-6.8$ & \underline{$-7.7$} \\
3 & EEGNet & $-23.9$ & $-5.2$ & $-9.8$ & \underline{$-4.6$} & \underline{$-9.0$} & $-9.4$ & {\boldmath $+0.2$} & \underline{$-5.2$} & $-8.4$ \\
4 & CBraMod & $-31.9$ & $-10.8$ & $-7.9$ & $-7.7$ & {\boldmath $-8.2$} & \underline{$-5.0$} & $-8.6$ & $-6.5$ & $-10.8$$^{\nabla}$ \\
5 & NeuroRVQ & $-35.1$ & $-12.8$ & $-12.4$ & $-9.3$ & $-10.1$ & $-6.2$ & $-7.1$ & {\boldmath $-2.4$} & $-11.9$$^{\nabla}$ \\
6 & BrainOmni & \underline{$-20.4$} & $-10.8$ & $-8.4$ & $-13.3$ & $-19.1$ & $-20.8$ & \underline{$-0.3$} & $-8.9$ & $-12.8$$^{\nabla}$ \\
\bottomrule
\end{tabular}}
\end{table}

\newpage
\subsubsection{True Random Channel Dropout}
\label{app:true_dropout}

\begin{figure}[!h]
\centering
\begin{subfigure}[t]{\textwidth}
\centering
\includegraphics[width=\textwidth]{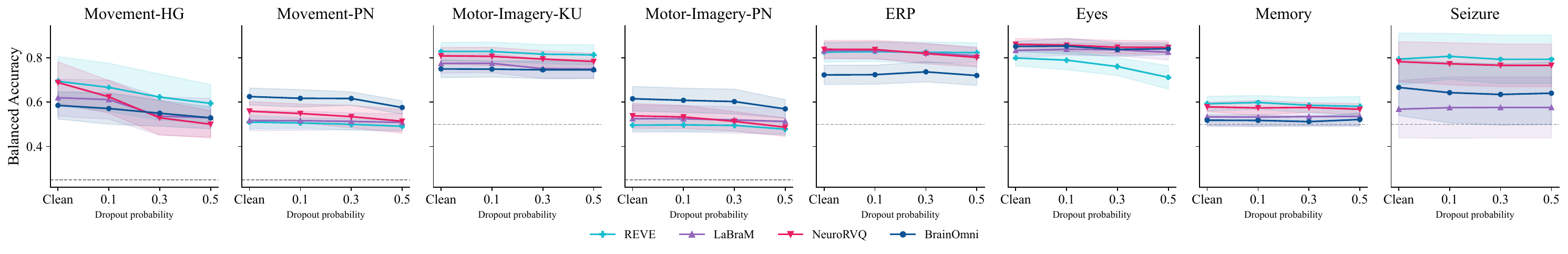}
\caption{Full fine-tuning.}
\label{fig:full_real_random_dropout}
\end{subfigure}
\vspace{0.3em}
\begin{subfigure}[t]{\textwidth}
\centering
\includegraphics[width=\textwidth]{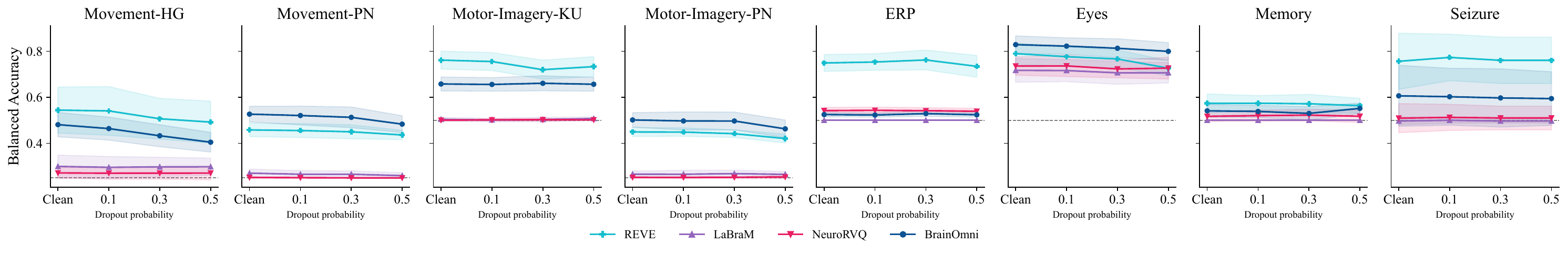}
\caption{Head-only adaptation.}
\label{fig:head_real_random_dropout}
\end{subfigure}
\caption{Per-benchmark degradation under true random channel dropout for the models that accept variable-length channel sets: REVE, NeuroRVQ, LaBraM, and BrainOmni.}
\end{figure}

In the main analysis, dropped channels are zeroed-out for all models to ensure a fair comparison. Here, we report results for true channel dropout, where dropped channels are removed from the input rather than zero-padded. This setting is restricted to models that accept variable-length channel sets: REVE, NeuroRVQ, LaBraM, and BrainOmni. At $p{=}0.50$, average losses range from $2.0$ to $4.8$ percentage points across the four models. Figures~\ref{fig:full_real_random_dropout} and \ref{fig:head_real_random_dropout} show per-benchmark curves under full fine-tuning and head-only adaptation respectively, and Table~\ref{tab:true_random_dropout_per_bench} reports per-benchmark degradation at $p{=}0.50$.

\vspace{-0.1in}
\begin{table}[h]
\centering
\caption{Per-benchmark degradation under Rand.\ removal $p{=}0.50$ (full fine-tuning). $\Delta$ percentage points from clean, paired per fold. Ranked by Average. Best per column in \textbf{bold}, second best \underline{underlined}. Average equally weights the 8 datasets (Movement-HG, Movement-PN, Motor-Imagery-KU, Motor-Imagery-PN, ERP, Eyes, Memory, Seizure); Sleep-EDF, Sleep-HMC excluded from the average.}
\label{tab:true_random_dropout_per_bench}
\scriptsize
\setlength{\tabcolsep}{3pt}
\resizebox{\textwidth}{!}{%
\begin{tabular}{rlccccccccc}
\toprule
\# & Model & Movement-HG & Movement-PN & Motor-Imagery-KU & Motor-Imagery-PN & ERP & Eyes & Memory & Seizure & Average \\
\midrule
1 & LaBraM & \underline{$-9.0$} & {\boldmath $-0.9$} & $-2.5$ & {\boldmath $-1.2$} & $-2.2$ & {\boldmath $-0.8$} & \underline{$+0.2$} & {\boldmath $+0.8$} & {\boldmath $-2.0$} \\
2 & BrainOmni & {\boldmath $-5.6$} & $-4.9$ & {\boldmath $-0.4$} & $-4.6$ & \underline{$-0.3$} & \underline{$-1.1$} & {\boldmath $+0.3$} & $-2.6$ & \underline{$-2.4$} \\
3 & REVE & $-10.0$ & \underline{$-1.9$} & \underline{$-1.5$} & \underline{$-1.8$} & {\boldmath $-0.2$} & $-8.7$ & $-1.2$ & \underline{$-0.1$} & $-3.2$ \\
4 & NeuroRVQ & $-18.8$ & $-4.5$ & $-2.6$ & $-5.1$ & $-3.7$ & $-1.4$ & $-1.1$ & $-1.6$ & $-4.8$ \\
\bottomrule
\end{tabular}}
\end{table}

\subsection{Region-based Perturbations}
\label{app:robustness_region}

Tables~\ref{tab:region_dropout_per_bench} and~\ref{tab:dropout_region_control_full} report per-benchmark degradation under primary and control region dropout respectively. Table~\ref{tab:region_noise_per_bench} reports degradation under primary region noise.

Across the eight datasets in the average, LaBraM, NeuroRVQ and REVE have the smallest losses under primary-region dropout ($9.2$, $11.1$ and $11.4$ percentage points). BrainOmni and EEGNet lose $12.0$ and $12.4$ points, and CBraMod has the largest loss ($15.4$ points). On Movement-HG, all models lose $23.5$--$32.3$ percentage points, consistent with reliance on central electrodes for motor execution. Motor-Imagery-KU follows the same pattern as Movement-HG but with smaller drops, as expected for imagery versus overt movement. ERP losses range from $7.5$ to $17.6$ points, consistent with posterior electrodes carrying the P300 signal. On Eyes, BrainOmni ($25.8$ points) and CBraMod ($19.3$ points) lose more than the other models ($7.4$--$15.5$ points). Sleep-EDF reaches chance when either channel is zero-padded, with losses of $40.9$--$51.7$ percentage points. Sleep-HMC remains above chance under primary-region dropout, with balanced accuracy of $0.364$--$0.521$. All six models lose more under control than primary dropout on Motor-Imagery-PN and three of six do on Movement-PN. This reversal is unexpected: neuroscience predicts that motor features should concentrate in central electrodes. Given that \citet{lee2025capabilities} identified eye-movement artefact contamination in these datasets (PhysioNet), we exclude Movement-PN and Motor-Imagery-PN from interpretability analyses. Under primary region noise, including sleep datasets, BrainOmni remains the most robust model at $-3$\,dB ($-6.1$ vs.\ $-12.5$ to $-17.1$ percentage points for other models).

\vspace{-0.1in}
\begin{table}[h]
\centering
\caption{Per-benchmark degradation under Region drop (pri.) (full fine-tuning). $\Delta$ percentage points from clean, paired per fold. Ranked by Average. Best per column in \textbf{bold}, second best \underline{underlined}. Average equally weights the 8 datasets (Movement-HG, Movement-PN, Motor-Imagery-KU, Motor-Imagery-PN, ERP, Eyes, Memory, Seizure); Sleep-EDF, Sleep-HMC excluded from the average. $^{\ast}$ on the Average marks a model significantly better than EEGNet, $^{\nabla}$ significantly worse (two-sided Wilcoxon signed-rank over the shared folds, Holm-corrected).}
\label{tab:region_dropout_per_bench}
\scriptsize
\setlength{\tabcolsep}{3pt}
\resizebox{\textwidth}{!}{%
\begin{tabular}{rlccccccccccc}
\toprule
\# & Model & Movement-HG & Movement-PN & Motor-Imagery-KU & Motor-Imagery-PN & ERP & Eyes & Memory & Sleep-EDF & Seizure & Sleep-HMC & Average \\
\midrule
1 & LaBraM & $-32.3$ & {\boldmath $-3.0$} & {\boldmath $-4.4$} & \underline{$-2.4$} & $-15.7$ & \underline{$-8.2$} & $-0.9$ & $-43.9$ & {\boldmath $-6.8$} & $-25.9$ & {\boldmath $-9.2$$^{\ast}$} \\
2 & NeuroRVQ & $-25.7$ & $-6.3$ & $-6.9$ & $-2.7$ & $-17.6$ & {\boldmath $-7.4$} & $-1.8$ & $-51.7$ & $-20.1$ & $-35.4$ & \underline{$-11.1$} \\
3 & REVE & {\boldmath $-23.5$} & \underline{$-4.3$} & \underline{$-5.0$} & \underline{$-2.4$} & \underline{$-14.1$} & $-13.1$ & $-1.7$ & $-46.3$ & $-27.2$ & $-24.0$ & $-11.4$ \\
4 & BrainOmni & \underline{$-24.7$} & $-8.5$ & $-5.9$ & $-7.2$ & {\boldmath $-7.5$} & $-25.8$ & {\boldmath $+0.5$} & {\boldmath $-40.9$} & \underline{$-16.7$} & \underline{$-20.6$} & $-12.0$ \\
5 & EEGNet & $-30.5$ & $-4.5$ & $-12.1$ & {\boldmath $-2.3$} & $-16.3$ & $-15.5$ & \underline{$+0.2$} & \underline{$-42.4$} & $-18.3$ & {\boldmath $-16.0$} & $-12.4$ \\
6 & CBraMod & $-25.1$ & $-13.5$ & $-13.4$ & $-6.5$ & $-14.6$ & $-19.3$ & $-5.7$ & $-46.8$ & $-25.6$ & $-36.1$ & $-15.4$$^{\nabla}$ \\
\bottomrule
\end{tabular}}
\end{table}

\begin{table}[h]
\centering
\caption{Per-benchmark degradation under Region drop (ctrl.) (full fine-tuning). $\Delta$ percentage points from clean, paired per fold. Ranked by Average. Best per column in \textbf{bold}, second best \underline{underlined}. Average equally weights the 8 datasets (Movement-HG, Movement-PN, Motor-Imagery-KU, Motor-Imagery-PN, ERP, Eyes, Memory, Seizure); Sleep-EDF, Sleep-HMC excluded from the average. $^{\ast}$ on the Average marks a model significantly better than EEGNet, $^{\nabla}$ significantly worse (two-sided Wilcoxon signed-rank over the shared folds, Holm-corrected).}
\label{tab:dropout_region_control_full}
\scriptsize
\setlength{\tabcolsep}{3pt}
\resizebox{\textwidth}{!}{%
\begin{tabular}{rlccccccccccc}
\toprule
\# & Model & Movement-HG & Movement-PN & Motor-Imagery-KU & Motor-Imagery-PN & ERP & Eyes & Memory & Sleep-EDF & Seizure & Sleep-HMC & Average \\
\midrule
1 & EEGNet & $-9.4$ & \underline{$-5.7$} & \underline{$-1.7$} & \underline{$-5.8$} & {\boldmath $-0.4$} & {\boldmath $+2.8$} & $-0.5$ & \underline{$-42.4$} & \underline{$-2.2$} & {\boldmath $-9.0$} & {\boldmath $-2.9$} \\
2 & REVE & {\boldmath $-0.7$} & {\boldmath $-3.9$} & {\boldmath $-1.0$} & {\boldmath $-4.9$} & \underline{$-0.9$} & \underline{$+0.9$} & \underline{$-0.2$} & $-46.3$ & $-16.1$ & \underline{$-16.2$} & \underline{$-3.3$} \\
3 & LaBraM & $-5.5$ & $-7.5$ & $-4.5$ & $-8.3$ & $-1.1$ & $-0.7$ & {\boldmath $-0.1$} & $-43.9$ & $-6.8$ & $-43.0$ & $-4.3$$^{\nabla}$ \\
4 & NeuroRVQ & \underline{$-4.4$} & $-8.9$ & $-3.6$ & $-7.8$ & $-2.3$ & $-3.8$ & $-2.4$ & $-51.7$ & $-6.7$ & $-42.5$ & $-5.0$$^{\nabla}$ \\
5 & BrainOmni & $-6.0$ & $-7.7$ & $-4.5$ & $-9.8$ & $-5.2$ & $-13.2$ & \underline{$-0.2$} & {\boldmath $-40.9$} & {\boldmath $-0.3$} & $-27.1$ & $-5.9$$^{\nabla}$ \\
6 & CBraMod & $-9.7$ & $-9.2$ & $-7.3$ & $-7.0$ & $-3.6$ & $-19.3$ & $-1.0$ & $-46.8$ & $-8.6$ & $-45.4$ & $-8.2$$^{\nabla}$ \\
\bottomrule
\end{tabular}}
\end{table}

\vspace{-0.1in}
\begin{table}[h]
\centering
\caption{Per-benchmark degradation under Primary Region noise $-3$\,dB (full fine-tuning). $\Delta$ percentage points from clean, paired per fold. Ranked by Average. Best per column in \textbf{bold}, second best \underline{underlined}. Average equally weights the 8 datasets (Movement-HG, Movement-PN, Motor-Imagery-KU, Motor-Imagery-PN, ERP, Eyes, Memory, Seizure); Sleep-EDF, Sleep-HMC excluded from the average. $^{\ast}$ on the Average marks a model significantly better than EEGNet, $^{\nabla}$ significantly worse (two-sided Wilcoxon signed-rank over the shared folds, Holm-corrected).}
\label{tab:region_noise_per_bench}
\scriptsize
\setlength{\tabcolsep}{3pt}
\resizebox{\textwidth}{!}{%
\begin{tabular}{rlccccccccccc}
\toprule
\# & Model & Movement-HG & Movement-PN & Motor-Imagery-KU & Motor-Imagery-PN & ERP & Eyes & Memory & Sleep-EDF & Seizure & Sleep-HMC & Average \\
\midrule
1 & BrainOmni & {\boldmath $-4.1$} & \underline{$-1.2$} & {\boldmath $-1.4$} & $-0.8$ & {\boldmath $-1.2$} & {\boldmath $-0.8$} & \underline{$-0.1$} & {\boldmath $-24.3$} & $-6.4$ & {\boldmath $-21.0$} & {\boldmath $-2.0$$^{\ast}$} \\
2 & REVE & \underline{$-15.3$} & $-7.2$ & \underline{$-4.1$} & $-3.9$ & \underline{$-7.7$} & $-3.5$ & $-2.9$ & \underline{$-34.4$} & $-7.6$ & $-38.5$ & \underline{$-6.5$} \\
3 & EEGNet & $-32.8$ & $-2.8$ & $-10.8$ & $-1.4$ & $-18.9$ & \underline{$-2.6$} & {\boldmath $+0.6$} & $-42.4$ & {\boldmath $+10.9$} & \underline{$-29.2$} & $-7.2$ \\
4 & LaBraM & $-31.5$ & {\boldmath $-0.0$} & $-5.3$ & \underline{$-0.6$} & $-13.2$ & $-6.9$ & $-0.4$ & $-36.6$ & $-0.5$ & $-31.8$ & $-7.3$ \\
5 & NeuroRVQ & $-20.8$ & $-2.7$ & $-5.5$ & {\boldmath $-0.4$} & $-11.8$ & $-18.2$ & $-3.4$ & $-38.3$ & $-8.9$ & $-31.9$ & $-9.0$ \\
6 & CBraMod & $-34.3$ & $-16.0$ & $-10.2$ & $-8.1$ & $-9.2$ & $-10.9$ & $-2.7$ & $-36.3$ & \underline{$+1.0$} & $-44.7$ & $-11.3$$^{\nabla}$ \\
\bottomrule
\end{tabular}}
\end{table}

\begin{figure}[!h]
\centering
\begin{subfigure}[t]{\textwidth}
\centering
\includegraphics[width=\textwidth]{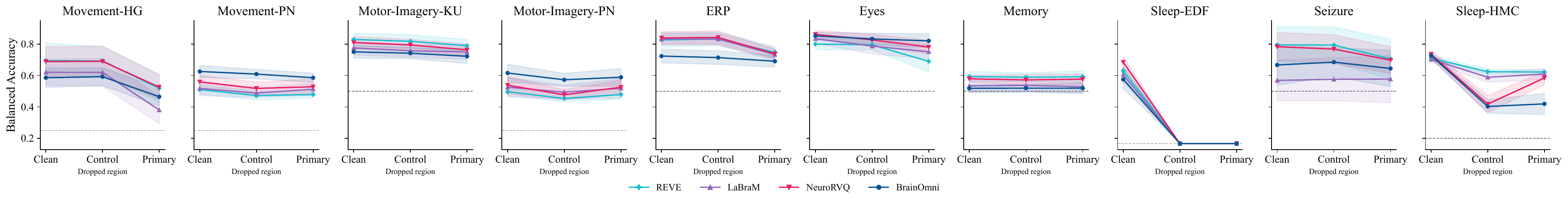}
\caption{Full fine-tuning.}
\label{fig:full_real_region_dropout}
\end{subfigure}
\vspace{0.3em}
\begin{subfigure}[t]{\textwidth}
\centering
\includegraphics[width=\textwidth]{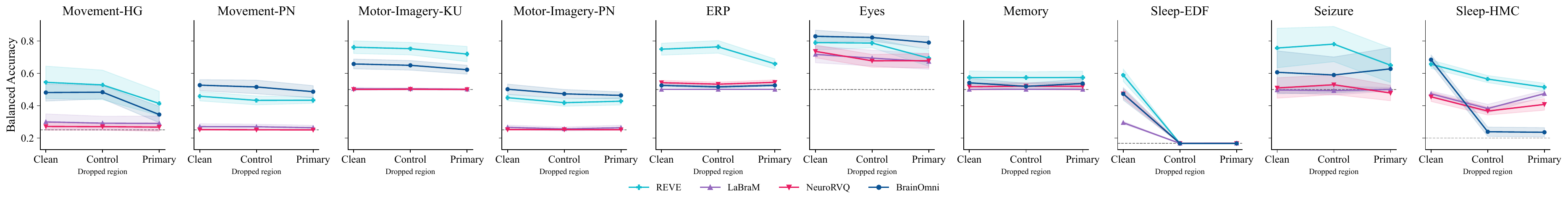}
\caption{Head-only adaptation.}
\label{fig:head_real_region_dropout}
\end{subfigure}
\caption{Per-benchmark degradation under true primary region dropout for the models that accept variable-length channel sets: REVE, NeuroRVQ, LaBraM, and BrainOmni.}
\end{figure}

Similarly to \ref{app:true_dropout}, we report results for true region dropout, where dropped regions are removed from the input rather than zero-padded. This setting is restricted to models that accept variable-length channel sets: REVE, NeuroRVQ, LaBraM, and BrainOmni. Figures~\ref{fig:full_real_region_dropout} and \ref{fig:head_real_region_dropout} show per-benchmark curves under full fine-tuning and head-only adaptation respectively, and Table~\ref{tab:true_region_dropout_per_bench} reports per-benchmark degradation under primary region dropout. As under true random dropout (Appendix~\ref{app:true_dropout}), all four variable-channel models improve on average when dropped regions are removed rather than zero-padded: average primary-region losses fall from $9.2$--$12.0$ to $3.8$--$6.8$ percentage points, with BrainOmni showing the largest gain.

\newpage
\vspace{-0.1in}
\begin{table}[h]
\centering
\caption{Per-benchmark degradation under Region removal (primary) (full fine-tuning). $\Delta$ percentage points from clean, paired per fold. Ranked by Average. Best per column in \textbf{bold}, second best \underline{underlined}. Average equally weights the 8 datasets (Movement-HG, Movement-PN, Motor-Imagery-KU, Motor-Imagery-PN, ERP, Eyes, Memory, Seizure).}
\label{tab:true_region_dropout_per_bench}
\scriptsize
\setlength{\tabcolsep}{3pt}
\resizebox{\textwidth}{!}{%
\begin{tabular}{rlccccccccccc}
\toprule
\# & Model & Movement-HG & Movement-PN & Motor-Imagery-KU & Motor-Imagery-PN & ERP & Eyes & Memory & Sleep-EDF & Seizure & Sleep-HMC & Average \\
\midrule
1 & BrainOmni & {\boldmath $-12.1$} & $-4.0$ & \underline{$-2.8$} & $-2.7$ & {\boldmath $-3.3$} & {\boldmath $-3.2$} & {\boldmath $+0.1$} & {\boldmath $-40.9$} & \underline{$-2.2$} & $-30.8$ & {\boldmath $-3.8$} \\
2 & LaBraM & $-23.9$ & {\boldmath $-0.5$} & {\boldmath $-2.5$} & {\boldmath $-0.8$} & $-9.8$ & $-8.3$ & $-0.5$ & \underline{$-43.9$} & {\boldmath $+0.9$} & \underline{$-9.4$} & \underline{$-5.7$} \\
3 & NeuroRVQ & \underline{$-16.3$} & \underline{$-3.1$} & $-4.6$ & \underline{$-1.3$} & $-10.1$ & \underline{$-8.0$} & $-0.3$ & $-51.7$ & $-8.5$ & $-15.0$ & $-6.5$ \\
4 & REVE & $-17.6$ & \underline{$-3.1$} & $-4.0$ & $-1.7$ & \underline{$-7.9$} & $-11.0$ & \underline{$-0.1$} & $-46.3$ & $-8.9$ & {\boldmath $-8.7$} & $-6.8$ \\
\bottomrule
\end{tabular}}
\end{table}
\FloatBarrier%

\newpage
\subsection{Vector Quantisation Ablation}
\label{app:vq_skip}

BrainOmni is the most robust under additive noise (Section~\ref{sec:robustness_sec}). We tested whether its vector quantiser, which maps the compressed latents to codebook entries before the transformer, accounts for this advantage.

\begin{figure}[!h]
\centering
\includegraphics[width=0.5\textwidth]{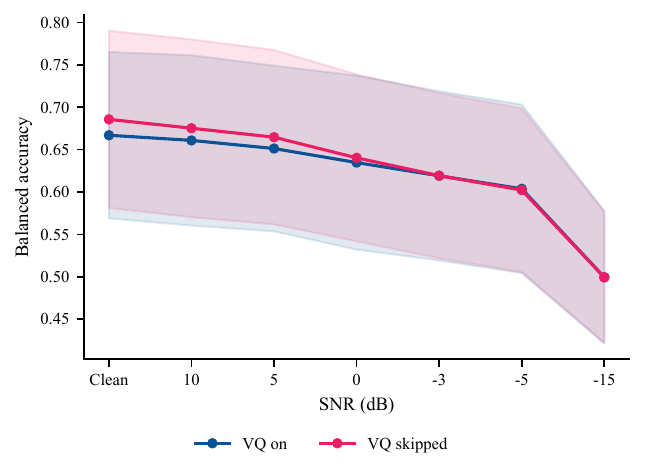}
\caption{BrainOmni with and without its inference-time vector quantiser, averaged over the eight datasets in the perturbation set. Mean balanced accuracy under additive white noise; shading shows one standard deviation across datasets.}
\label{fig:vq_skip}
\end{figure}

We skipped the quantiser at inference and passed the continuous latents straight to the transformer. Figure~\ref{fig:vq_skip} compares both variants across the same eight datasets used in the robustness averages. At $-5$\,dB, the average loss rises from $6.3$ to $8.4$ percentage points when VQ is skipped. The skipped variant still loses less than every other foundation model ($12.4$--$20.1$ percentage points) and EEGNet ($11.9$ percentage points). Across all ten datasets, skipping VQ raises mean clean balanced accuracy from $0.664$ to $0.681$. Inference-time quantisation is therefore not critical for BrainOmni’s noise-robustness. Its multimodal pre-training on MEG and EEG, or the latent compression in its tokeniser remain as two plausible sources of noise robustness. Separating them needs matched pre-training runs, which lie outside our compute budget.

\section{Interpretability}
\label{app:interpretability}

This section provides details of the interpretability methods used in the main paper: attribution techniques (AttnLRP, Gradient $\times$ Input and GradCAM), raw attention extraction, linear probing and block truncation ablation. Each subsection covers method and results together. Table~\ref{tab:interp_methods} summarises which methods were applied to each model.

\begin{table}[h]
\caption{Interpretability methods applied to each model.}
\label{tab:interp_methods}
\centering
\small
\begin{tabular}{lcccccc}
\toprule
Method & LaBraM & NeuroRVQ & REVE & CBraMod & BrainOmni & EEGNet \\
\midrule
Attribution & LRP & Grad$\times$Input & LRP & LRP & Grad$\times$Input & LRP \\
GradCAM & \checkmark & \checkmark & \checkmark & \checkmark & --- & \checkmark \\
Raw attention & --- & \checkmark & \checkmark & --- & --- & --- \\
Linear probing & --- & \checkmark & \checkmark & --- & --- & --- \\
\bottomrule
\end{tabular}
\end{table}

\subsection{Attention-Aware Layer-Wise Relevance Propagation}
\label{apx:lrp_g}

\paragraph{AttnLRP.} We apply AttnLRP \citep{achtibat2024attnlrp}, a recent extension of LRP to transformers. LRP propagates relevance backward through each layer using conservation rules, producing attributions that sum to the model output. LRP passes  label- and model-weight randomisation sensitivity checks on EEG and accurately recovers ground-truth spatial features~\citep{ravindran2023eegxai_groundtruth}. AttnLRP extends LRP by adding dedicated rules for the operations inside self-attention (the bilinear query-key and attention-value products, and the softmax). We use AttnLRP's standard propagation rules as defined in \citep{achtibat2024attnlrp}: an $\varepsilon$-stabilised rule for linear layers, an identity rule for activations and normalisations, and an equal-split bilinear rule for the two attention matrix multiplications, with a Taylor-decomposition rule for the softmax. These are applied via the authors' own LXT library~\citep{achtibat2024attnlrp}, and we implement custom auto-grad functions for operations without LXT equivalents. 

\paragraph{Gradient $\times$ Input.} For NeuroRVQ and BrainOmni, AttnLRP does not produce interpretable maps. NeuroRVQ's multi-scale convolutional encoder produces $O(0.01)$ token embeddings, so the $\varepsilon$-rule's $R / (z_j + \varepsilon)$ division by near-zero activations causes relevance to compound (${\sim}100\times$ per block) and destroy spatial structure across the 12 blocks. Likewise, BrainOmni's frozen tokeniser and VQ discretisation on the forward pass create a similar barrier for relevance propagation. For both models, we use Gradient $\times$ Input ($\partial y / \partial x \cdot x$), which produces clearer spatial maps where LRP fails.

\begin{figure}[!h]
\centering
\includegraphics[width=0.7\textwidth]{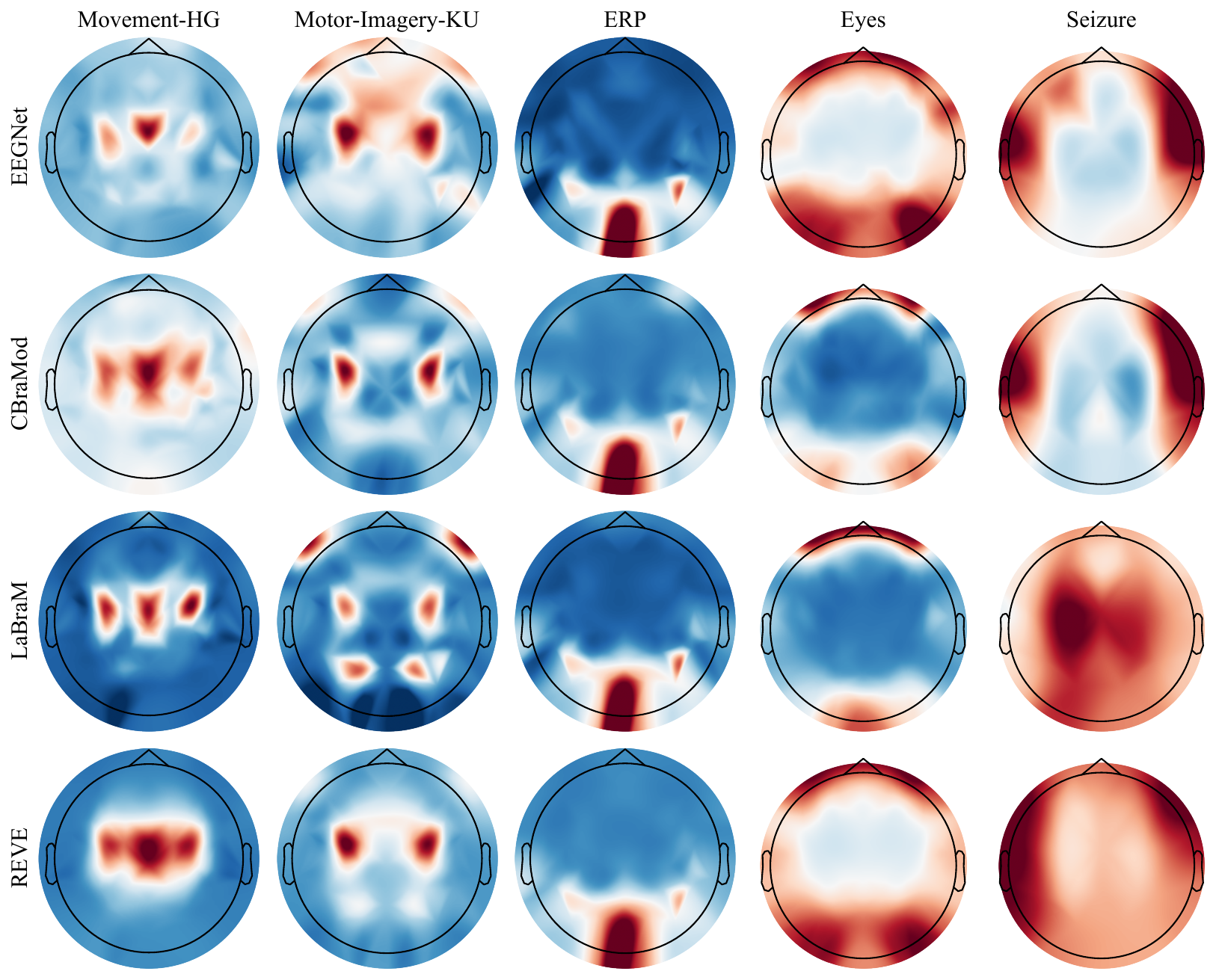}
\caption{
Class-averaged Gradient $\times$ Input  topographic maps. Rows (Models): EEGNet, CBraMod, LaBraM and REVE. Columns (Benchmarks): Movement-HG, Motor-Imagery-KU, ERP, Eyes and Seizure.}
\label{fig:ixg_grid}
\end{figure}

We therefore use Gradient $\times$ Input (G$\times$I) as a complementary attribution method for these models. G$\times$I is among the methods that pass sanity checks and recover ground-truth features in simulated EEG~\citep{ravindran2023eegxai_groundtruth}. To further verify that Gradient $\times$ Input is a reasonable alternative rather than a model-specific workaround, we also compute it for the models where AttnLRP is numerically stable (LaBraM, REVE, CBraMod, EEGNet). Figure~\ref{fig:ixg_grid} shows class-averaged Gradient $\times$ Input topographic maps across five benchmarks and the remaining EEG-FMs. The spatial patterns agree with the AttnLRP maps in Figure~\ref{fig:lrp_grid} for the LRP-stable models. However, G$\times$I has known limitations such as noise sensitivity~\citep{smilkov2017smoothgrad} and does not adhere to conservation rules like LRP, which we accept in exchange for numerical stability. \textbf{This supports using G$\times$I as an alternative for the two models where AttnLRP is unsuitable.}

Both methods (AttnLRP and Gradient $\times$ Input.) produce a relevance map of the same shape as the input ($C \times T$). We restrict the analysis to correctly classified test samples whose prediction confidence exceeds the 75th percentile, excluding ambiguous trials whose attributions would otherwise add noise to the class average. Each map is normalised per sample by the 99th percentile of $|R|$ and clipped to $[-1, 1]$. We average $|R|$ over time, trials, and folds. Taking the absolute value discards sign, so the maps show the magnitude of relevance rather than whether evidence supports or opposes the prediction. The resulting class-averaged topographic maps are shown in Figure~\ref{fig:lrp_grid}; per-class (non-averaged) AttnLRP maps appear alongside GradCAM in Figure~\ref{fig:attr_perclass_all}.

\subsection{Attribution under Perturbation}
\label{sec:lrp_perturbation}

\begin{figure}[!h]
\centering
\begin{subfigure}[t]{0.49\textwidth}
\centering
\includegraphics[width=\textwidth]{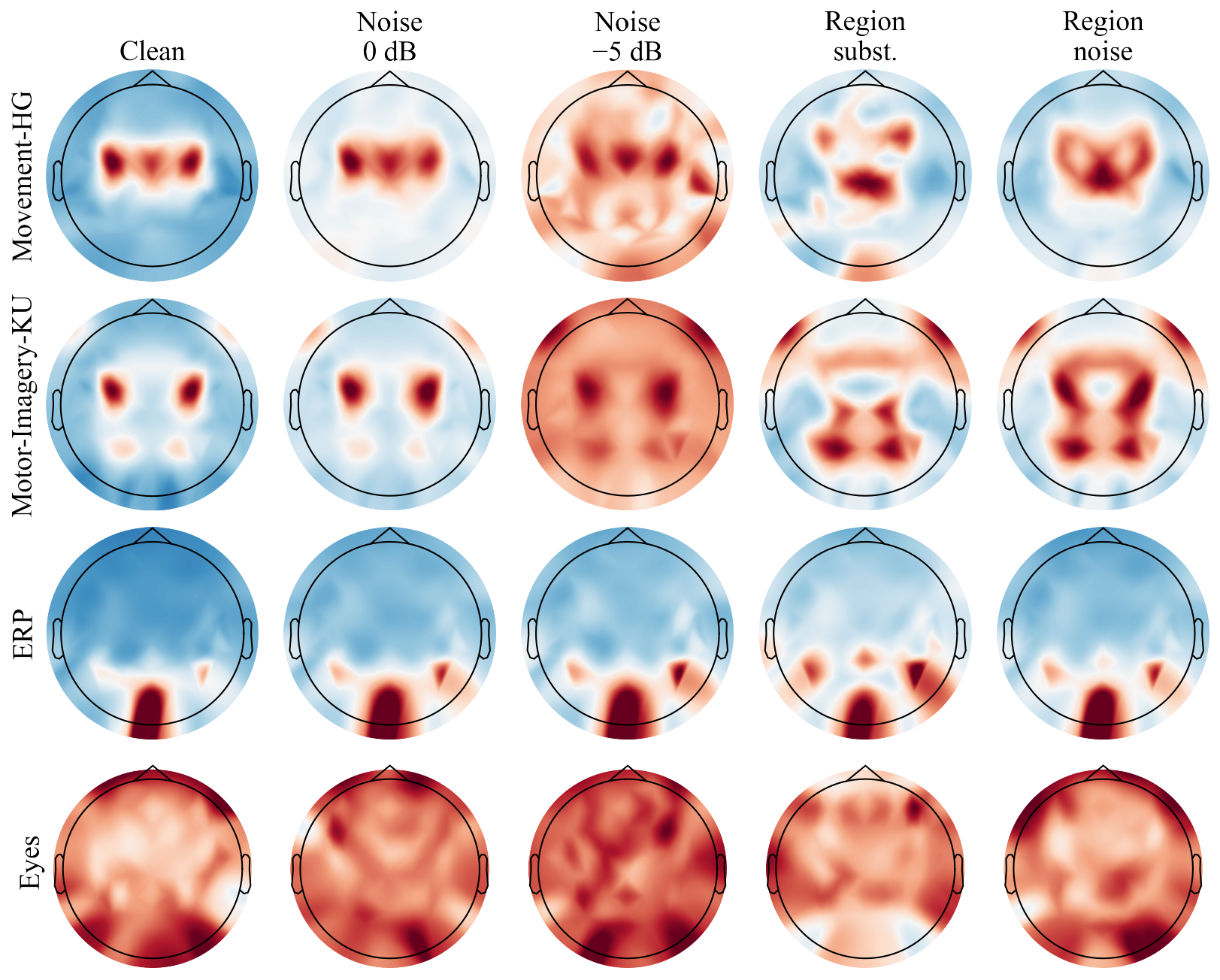}
\caption{REVE}
\label{fig:lrp_perturb_reve}
\end{subfigure}
\hfill
\begin{subfigure}[t]{0.49\textwidth}
\centering
\includegraphics[width=\textwidth]{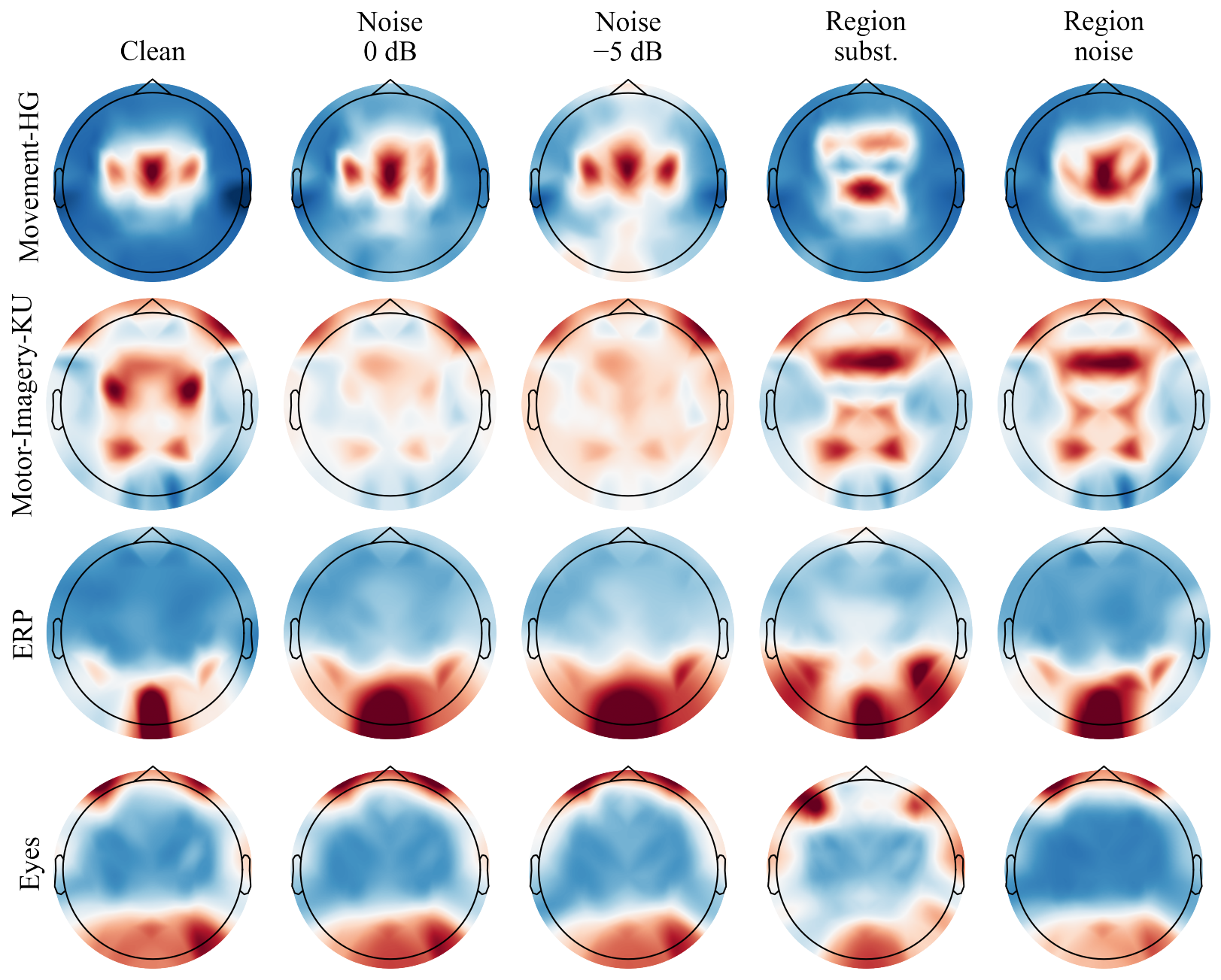}
\caption{NeuroRVQ}
\label{fig:lrp_perturb_neurorqv}
\end{subfigure}
\caption{Class-averaged attribution maps, averaged over folds, for REVE (AttnLRP) and NeuroRVQ (Input $\times$ Gradient). Rows: Movement-HG, Motor-Imagery-KU, ERP and Eyes. Columns: clean input, white noise at $0$\,dB and $-5$\,dB, matched white-noise substitution in the primary region, and primary-region white noise at $5$\,dB. Substitution replaces zero-padded region dropout in this analysis.}
\label{fig:lrp_perturb}
\end{figure}

The robustness results (Section~\ref{sec:robustness_sec}) show how much performance degrades, while the attribution maps (Section~\ref{sec:interpretability}) show where models assign relevance on clean data. This section connects the two by examining attribution under perturbation. We compare REVE and NeuroRVQ, the two highest-performing models in full fine-tuning (Section~\ref{sec:clean}), on Movement-HG, Motor-Imagery-KU, ERP and Eyes. Figure~\ref{fig:lrp_perturb} shows averaged maps from clean inputs and correctly-classified samples under four perturbation conditions: white noise at $0$\,dB and $-5$\,dB, primary-region noise at $5$\,dB, and primary-region substitution. Substitution replaces regional signals with filtered white noise scaled to the intact channels, avoiding potential zero-padding artefacts (Appendix~\ref{app:substitution}). Under stronger noise, Movement-HG and Motor-Imagery-KU maps become more diffuse, though Movement-HG retains a central focus and Motor-Imagery-KU gains frontal relevance. ERP retains a posterior focus, while Eyes changes little: NeuroRVQ preserves its spatial pattern and REVE remains diffuse. Substitution shifts relevance towards neighbouring motor electrodes, preserves the posterior focus on ERP and causes smaller changes on Eyes. Some spatial patterns therefore persist despite overall accuracy losses. However, we only plot correct trials in the highest confidence quartile within each condition, so the selected trials may differ and do not show what happens when predictions fail.  Further analyses of attention under perturbed conditions are presented in Appendix~\ref{app:attn_perturbation}.

\newpage
\subsection{GradCAM}
\label{app:gradcam_appendix}

\begin{figure}
\centering
\begin{subfigure}[b]{1\textwidth}
    \centering
    \includegraphics[width=\textwidth]{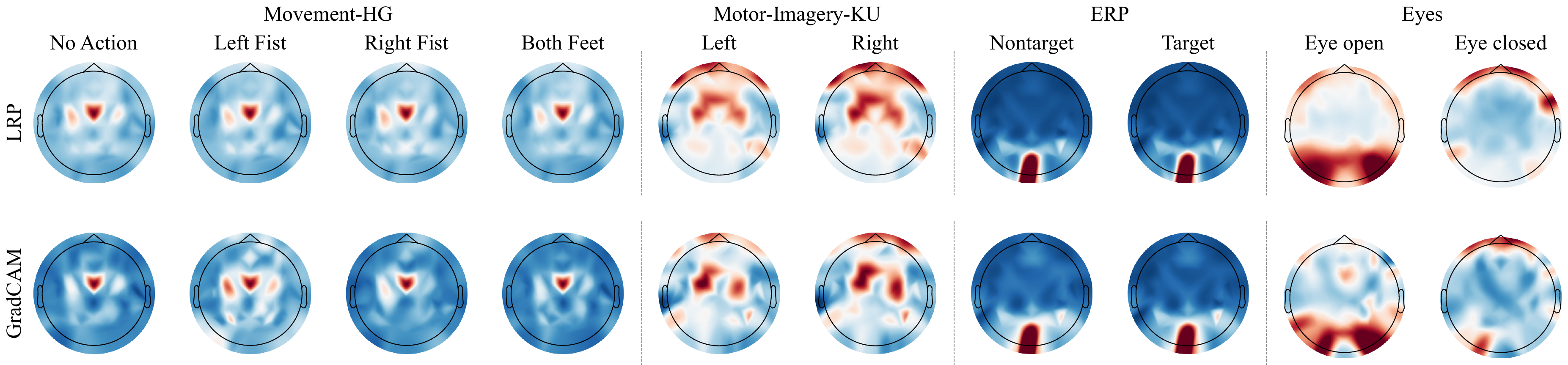}
    \caption{EEGNet}
    \label{fig:attr_perclass_CBraMod}
\end{subfigure}
\begin{subfigure}[b]{1\textwidth}
    \centering
    \includegraphics[width=\textwidth]{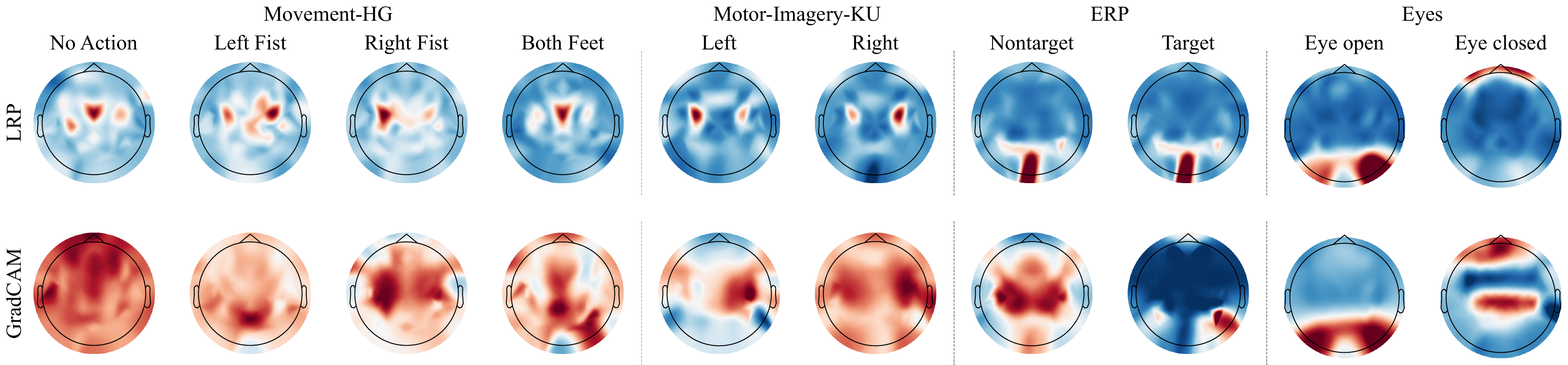}
    \caption{CBraMod}
    \label{fig:attr_perclass_CBraMod}
\end{subfigure}
\hfill
\begin{subfigure}[b]{1\textwidth}
    \centering
    \includegraphics[width=\textwidth]{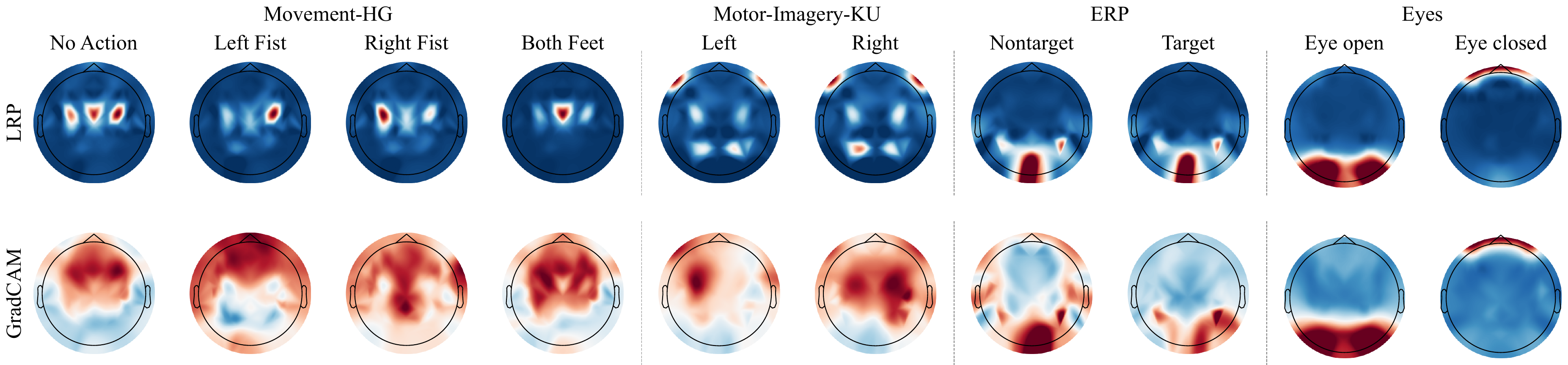}
    \caption{LaBraM}
    \label{fig:attr_perclass_LaBraM}
\end{subfigure}
\begin{subfigure}[b]{1\textwidth}
    \centering
    \includegraphics[width=\textwidth]{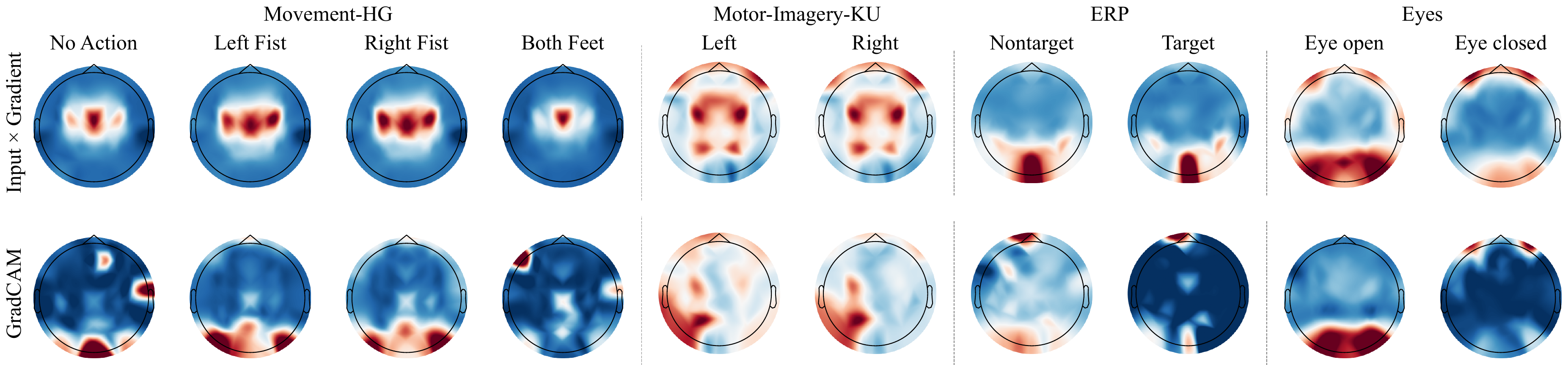}
    \caption{NeuroRVQ}
    \label{fig:attr_perclass_NeuroRVQ}
\end{subfigure}
\hfill
\begin{subfigure}[b]{1\textwidth}
    \centering
    \includegraphics[width=\textwidth]{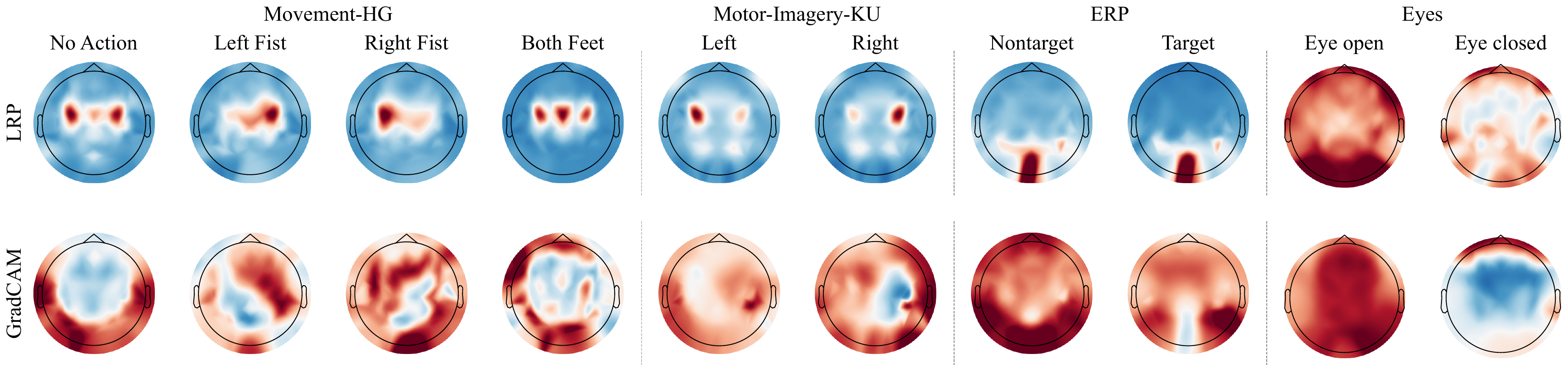}
    \caption{REVE}
    \label{fig:attr_perclass_REVE}
\end{subfigure}
\hfill

\begin{subfigure}[b]{1\textwidth}
    \centering
    \includegraphics[width=\textwidth]{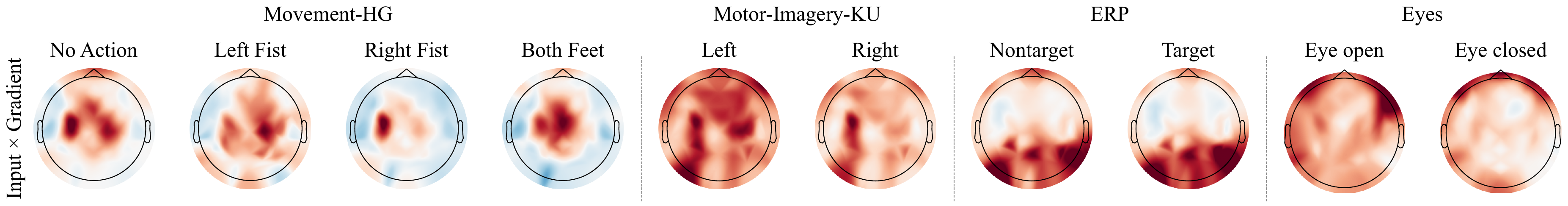}
    \caption{BrainOmni}
    \label{fig:attr_perclass_BrainOmni}
\end{subfigure}
\caption{Per-class attribution maps (AttnLRP / GradCAM) across all benchmarks for each model. Note: BrainOmni has no direct mapping of attention to channels making it incompatible with GradCAM.}
\label{fig:attr_perclass_all}
\end{figure}

We apply GradCAM \citep{selvaraju2017gradcam} as a supplementary attribution method using the pytorch-grad-cam library~\citep{jacobgilpytorchcam}, targeting the last transformer block for transformer-based models and the final convolutional layer for EEGNet. The same sample selection, normalisation, and averaging pipeline as AttnLRP is applied (see \ref{apx:lrp_g}). GradCAM requires a one-to-one mapping between spatial tokens and electrodes. BrainOmni compresses channels into latent source variables, breaking this mapping, so its GradCAM maps are not spatially interpretable at the electrode level. Figure \ref{fig:attr_perclass_all} shows the comparison results between GradCAM and AttLRP, with AttnLRP producing clearer, more interpretable topographic maps.

\newpage
\subsection{Attention Maps}
\label{app:raw_attention}

We extract raw softmax attention weights from each transformer block for REVE and NeuroRVQ, the two highest-performing architectures. Both models produce standard ($batch$, $heads$, $n_{\text{tokens}}$, $n_{\text{tokens}}$) attention matrices. Per-channel attention is computed by summing each attention matrix over the query dimension, then averaging over heads, frequencies (for NeuroRVQ), time patches, samples, and all cross-validation folds. NeuroRVQ carries a $CLS$ token, which is excluded.

\subsubsection{Per-block Attention: Pre-trained vs. Fine-tuned}
\label{app:attn_ft_perblock}

\begin{figure}[h]
\centering
\begin{subfigure}[b]{0.95\textwidth}
    \centering
    \includegraphics[width=\textwidth]{sec8_ft_attn_perblock_REVE_Movement-HG_average.pdf}
    \caption{Movement-HG}
    \label{fig:ft_attn_reve_hg}
\end{subfigure}
\hfill
\begin{subfigure}[b]{0.95\textwidth}
    \centering
    \includegraphics[width=\textwidth]{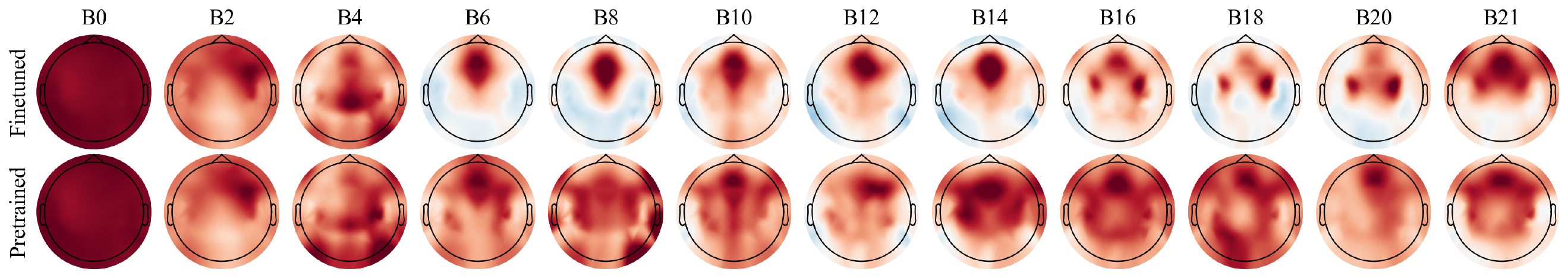}
    \caption{Motor-Imagery-KU}
    \label{fig:ft_attn_reve_kumi}
\end{subfigure}

\begin{subfigure}[b]{0.95\textwidth}
    \centering
    \includegraphics[width=\textwidth]{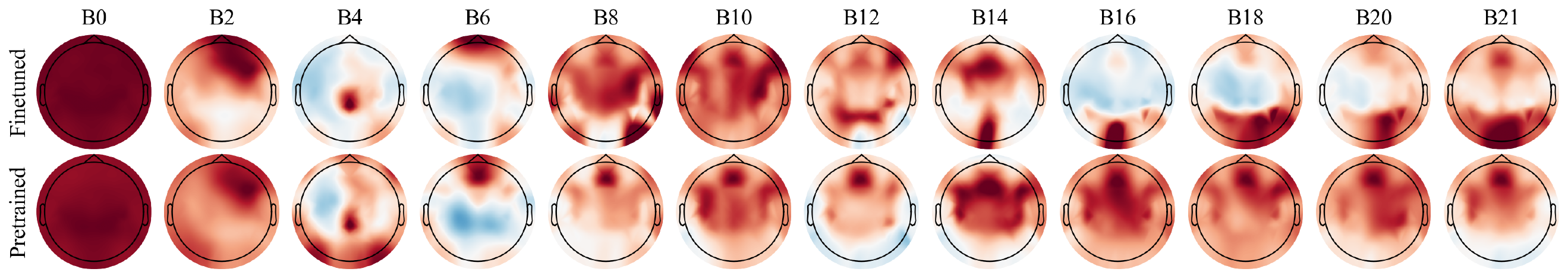}
    \caption{ERP}
    \label{fig:ft_attn_reve_erp}
\end{subfigure}
\hfill
\begin{subfigure}[b]{0.95\textwidth}
    \centering
    \includegraphics[width=\textwidth]{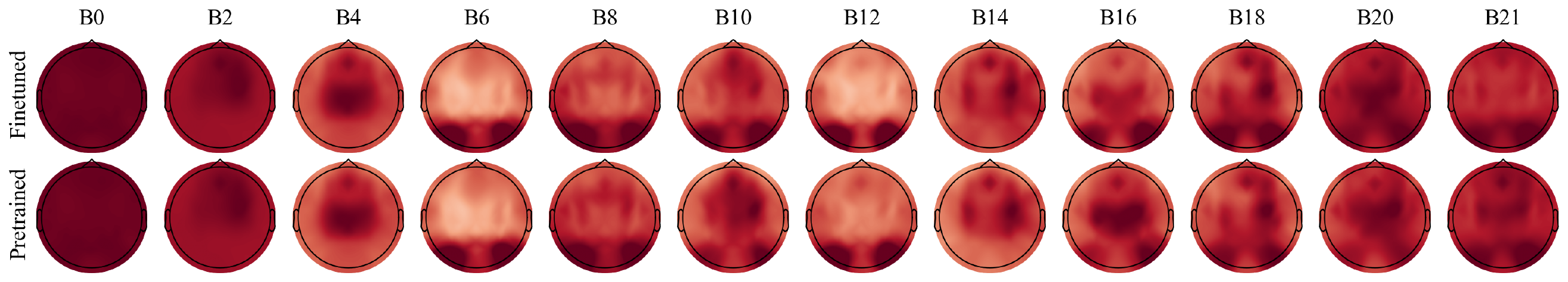}
    \caption{Eyes}
    \label{fig:ft_attn_reve_eyes}
\end{subfigure}
\caption{REVE per-block attention  topographic maps across datasets. Top: fine-tuned. Bottom: pre-trained. Blocks (B) ordered from input (left) to output (right).}
\label{fig:ft_attn_reve}
\end{figure}

\begin{figure}[!h]
\centering
\begin{subfigure}[b]{\textwidth}
    \centering
    \includegraphics[width=\textwidth]{sec8_ft_attn_perblock_NeuroRVQ_Movement-HG_average1.pdf}
    \caption{Movement-HG}
    \label{fig:ft_attn_neurorqv_hg}
\end{subfigure}
\hfill
\begin{subfigure}[b]{\textwidth}
    \centering
    \includegraphics[width=\textwidth]{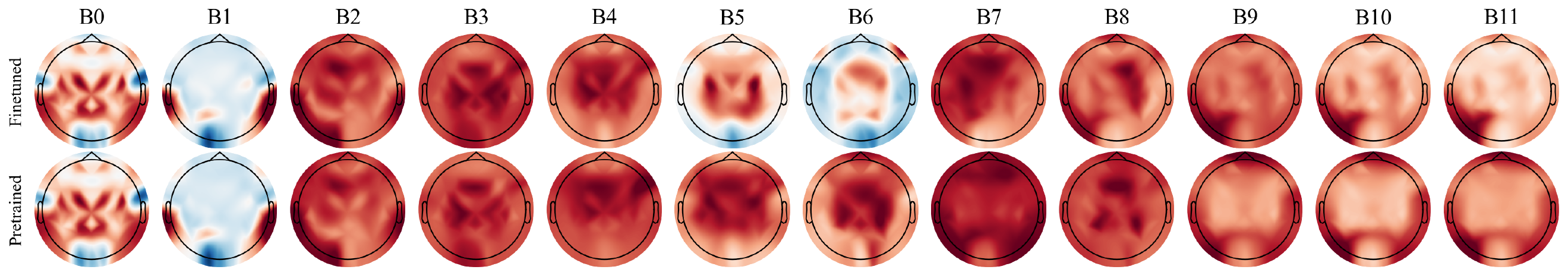}
    \caption{Motor-Imagery-KU}
    \label{fig:ft_attn_neurorqv_kumi}
\end{subfigure}

\begin{subfigure}[b]{\textwidth}
    \centering
    \includegraphics[width=\textwidth]{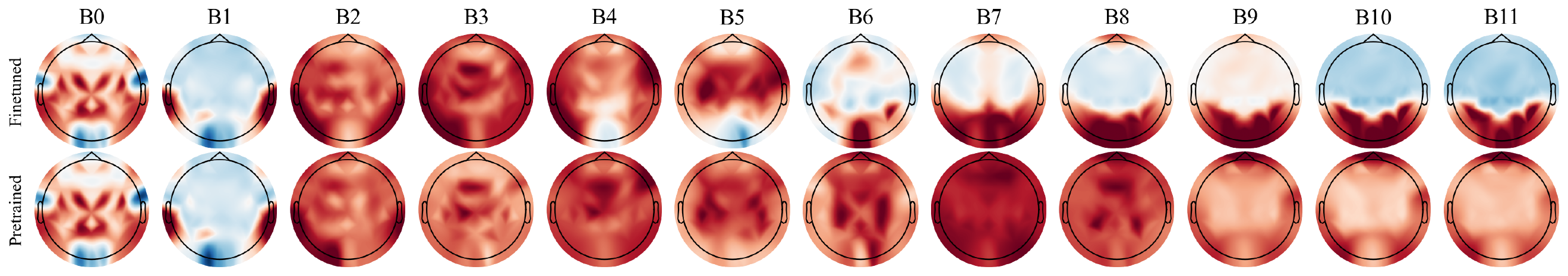}
    \caption{ERP}
    \label{fig:ft_attn_neurorqv_erp}
\end{subfigure}
\hfill
\begin{subfigure}[b]{\textwidth}
    \centering
    \includegraphics[width=\textwidth]{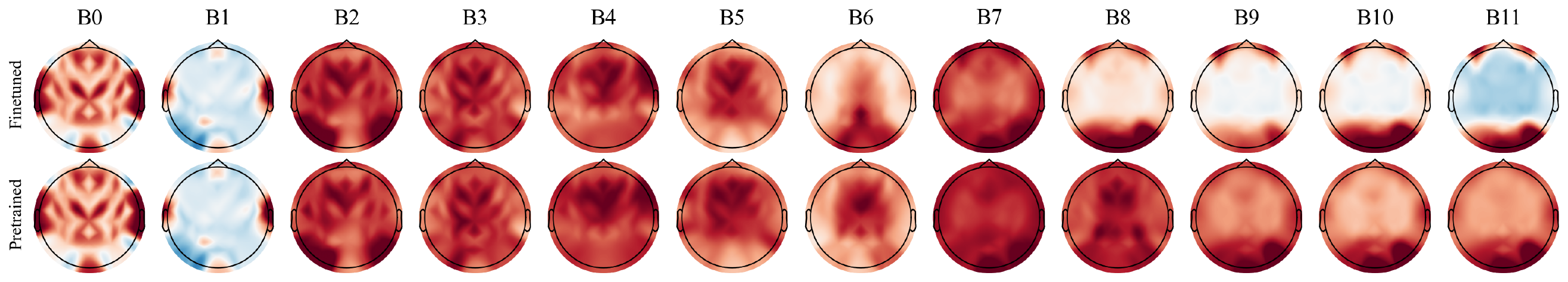}
    \caption{Eyes}
    \label{fig:ft_attn_neurorqv_eyes}
\end{subfigure}
\caption{NeuroRVQ per-block attention topographic maps across datasets. Top: fine-tuned. Bottom: pre-trained. Blocks (B) ordered from input (left) to output (right).}
\label{fig:ft_attn_neurorqv}
\end{figure}

Figures~\ref{fig:ft_attn_reve} and~\ref{fig:ft_attn_neurorqv} compare per-block attention topographic maps between  fine-tuned (top row) and pre-trained (bottom row) models across four benchmarks. Each column corresponds to one transformer block, ordered from input (left) to output (right). Early blocks show near-identical attention under both training modes. For REVE, the divergence varies by task. On Movement-HG, consistent divergence appears from block 16 onward, with some earlier blocks already showing small differences. On Motor-Imagery-KU, fine-tuned attention diverges from block 6 and focuses on more task-appropriate regions than the head-only maps. On ERP, differences emerge as early as block 4, but only blocks 14--21 settle on the occipital focus. On Eyes, fine-tuned and pre-trained attention are both diffuse across all blocks with no clear divergence point, consistent with the pre-trained checkpoint already carrying task signal (Figure \ref{fig:probing}). For NeuroRVQ, the fine-tuned and pre-trained attention patterns diverge around block five on ERP and motor tasks (Movement-HG, Motor-Imagery-KU) and around block eight on Eyes. This is broadly consistent with the probing transitions in Section~\ref{sec:probing}. On Motor-Imagery-KU, ERP and Movement-HG, middle blocks produce attention maps with clearer alignment to expected cortical regions than the final blocks.

\newpage

\subsubsection{Attention under Perturbation}
\label{app:attn_perturbation}
\begin{figure}[!h]
\centering
\begin{subfigure}[b]{\textwidth}
\centering
\includegraphics[width=\textwidth]{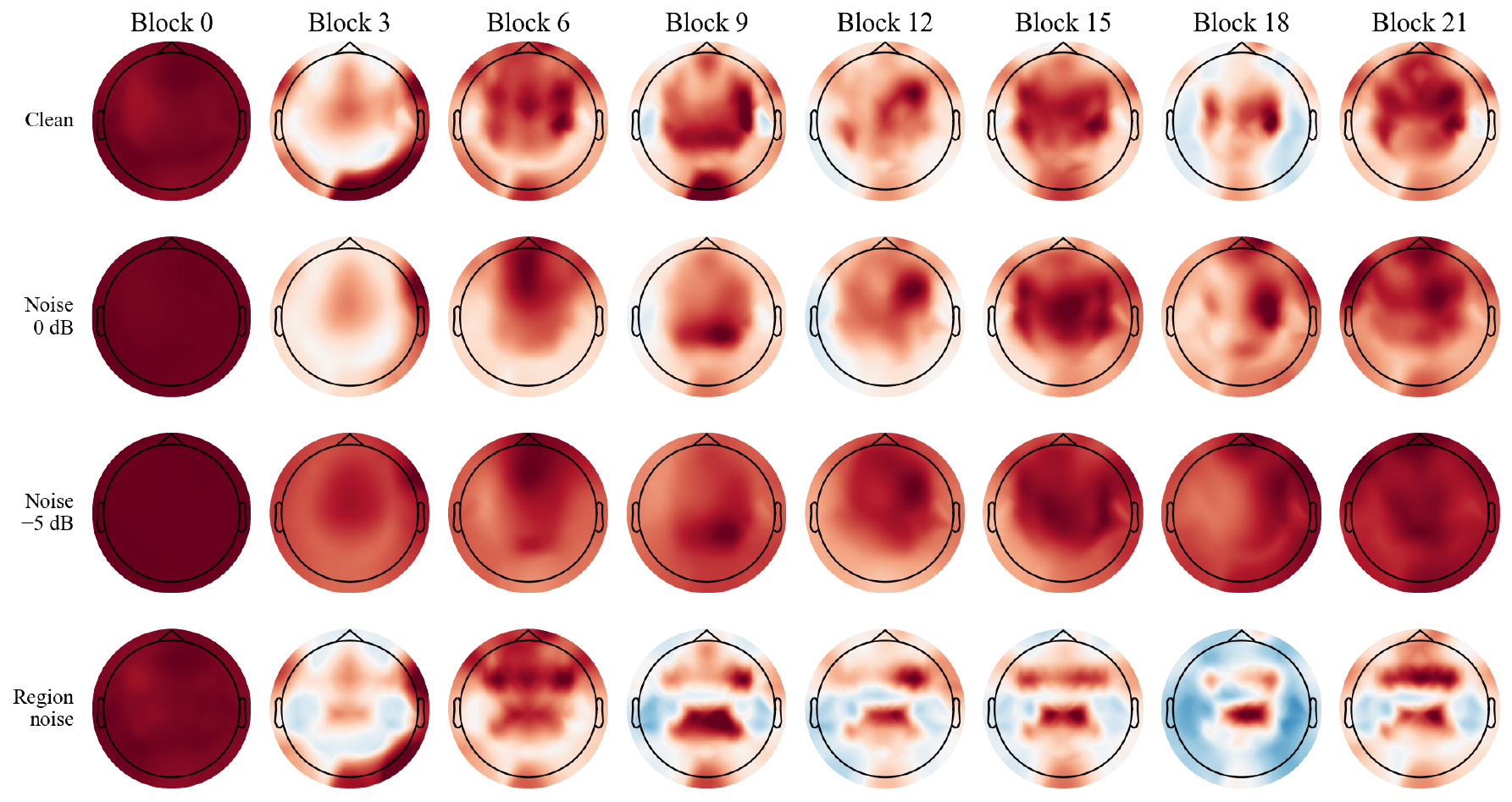}
\caption{Movement-HG}
\label{fig:attn_perturb_reve_movement}
\end{subfigure}
\vspace{0.5em}
\begin{subfigure}[b]{\textwidth}
\centering
\includegraphics[width=\textwidth]{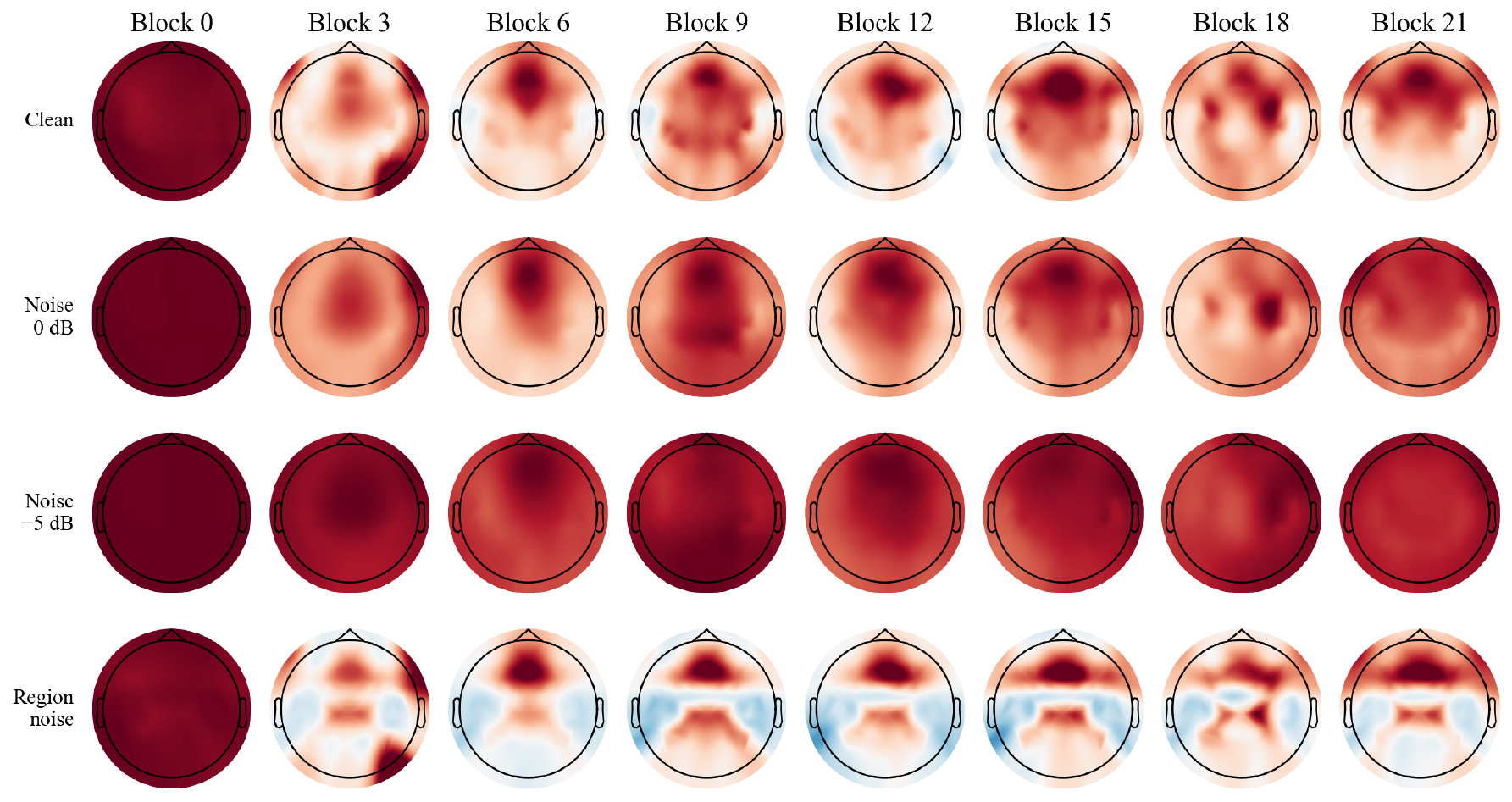}
\caption{Motor-Imagery-KU}
\label{fig:attn_perturb_reve_imagery}
\end{subfigure}
\caption{REVE per-block attention under perturbation across datasets. Blocks ordered from input (left) to output (right) \textit{- Plot 1/2}.}
\end{figure}

\begin{figure}[!h]
\centering
\begin{subfigure}[b]{\textwidth}
\centering
\includegraphics[width=\textwidth]{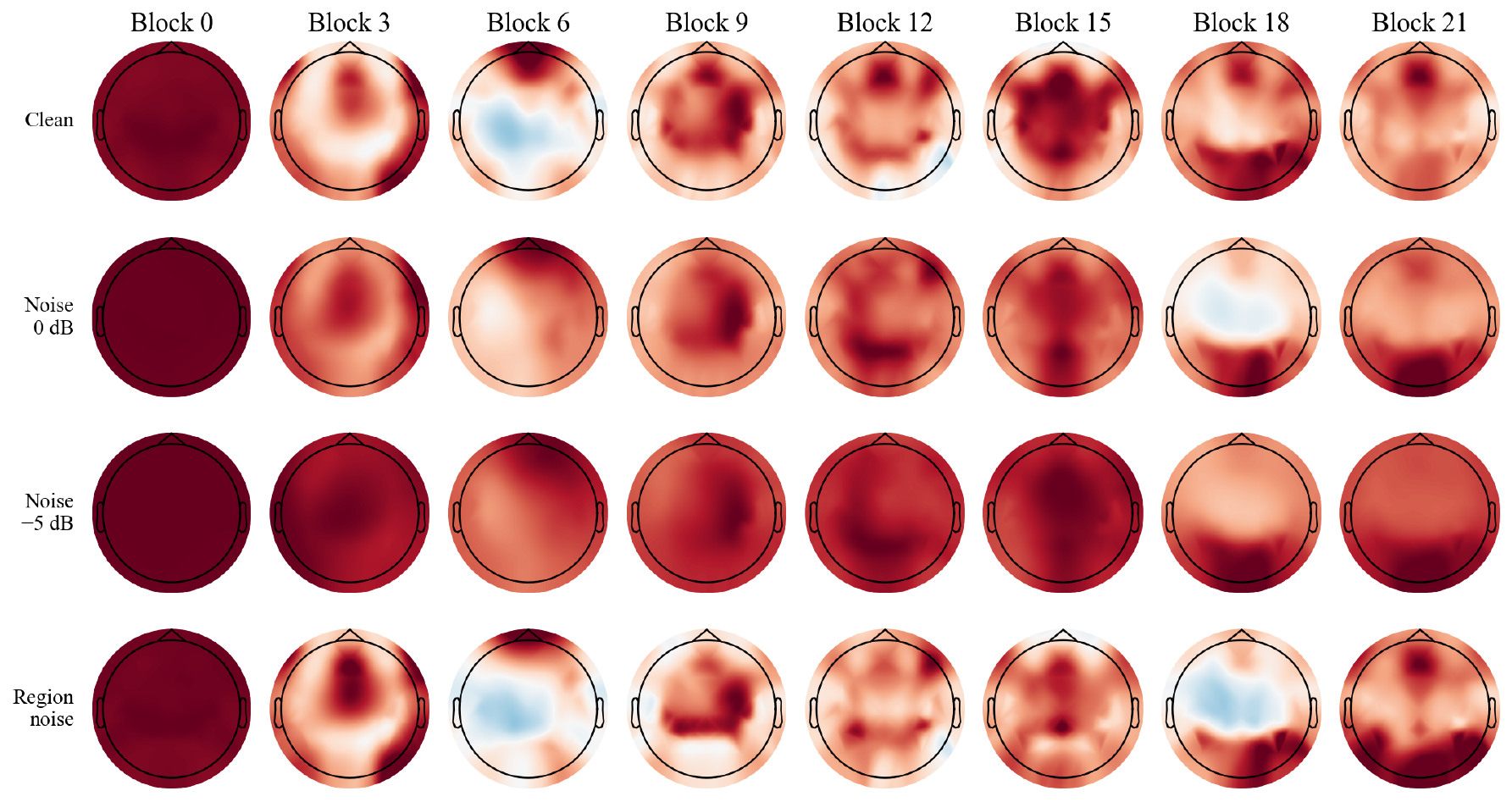}
\caption{ERP}
\label{fig:attn_perturb_reve_erp}
\end{subfigure}
\vspace{0.5em}
\begin{subfigure}[b]{\textwidth}
\centering
\includegraphics[width=\textwidth]{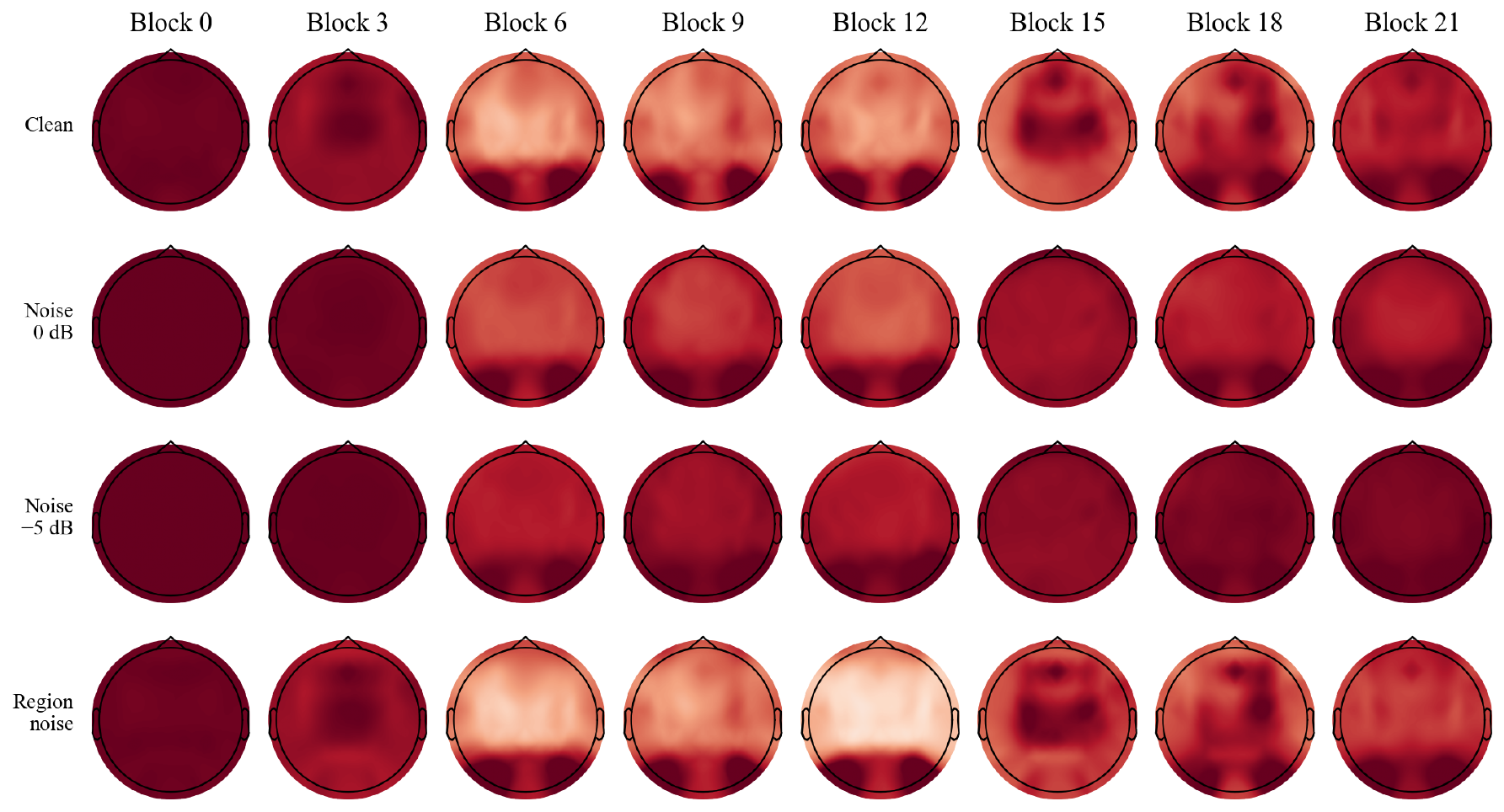}
\caption{Eyes}
\label{fig:attn_perturb_reve_eyes}
\end{subfigure}
\caption{REVE per-block attention under perturbation across datasets. Blocks ordered from input (left) to output (right) \textit{- Plot 2/2}.}
\end{figure}

\begin{figure}[!h]
\centering
\begin{subfigure}[b]{\textwidth}
\centering
\includegraphics[width=\textwidth]{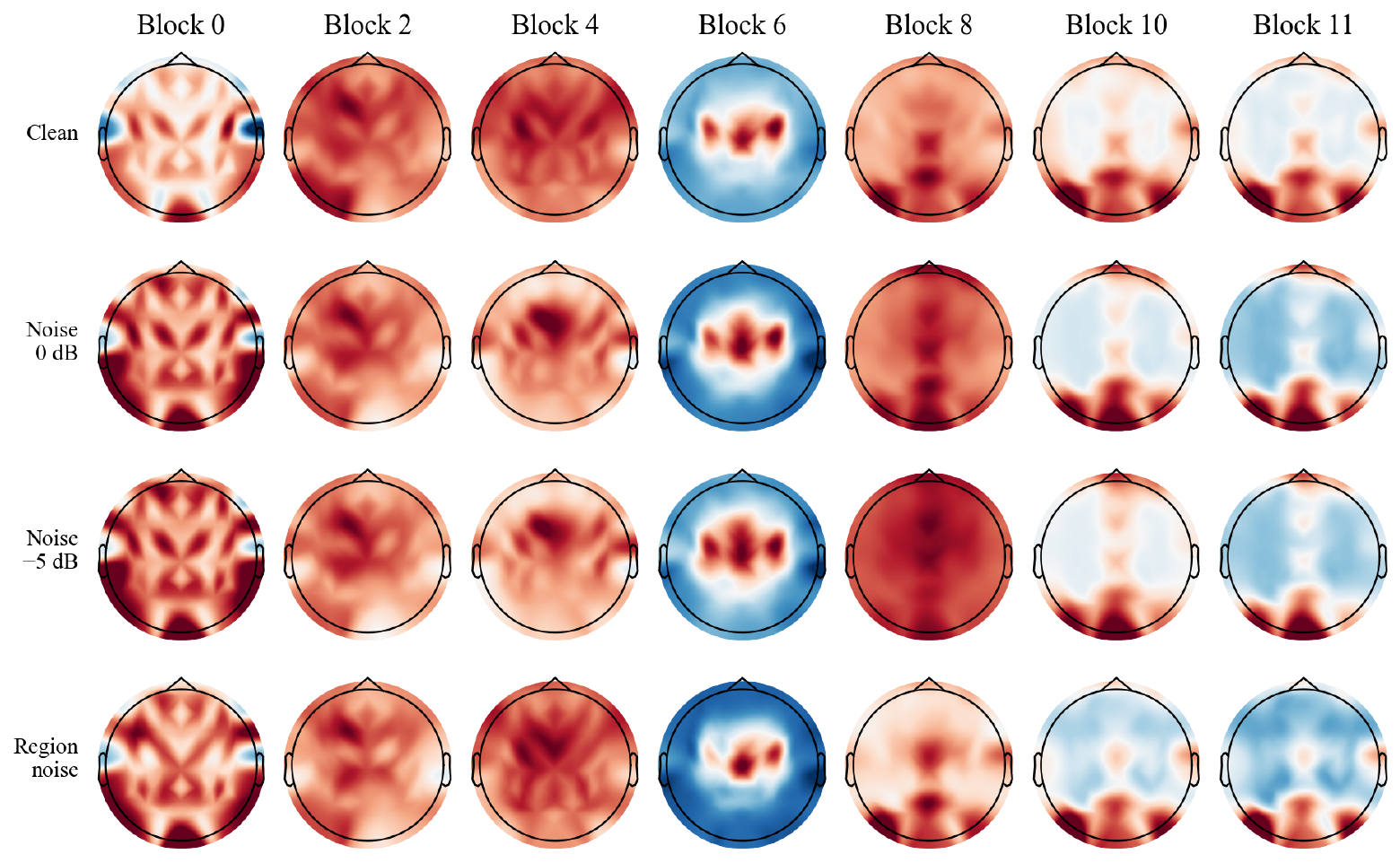}
\caption{Movement-HG}
\label{fig:attn_perturb_neurorqv_movement}
\end{subfigure}
\vspace{0.5em}
\begin{subfigure}[b]{\textwidth}
\centering
\includegraphics[width=\textwidth]{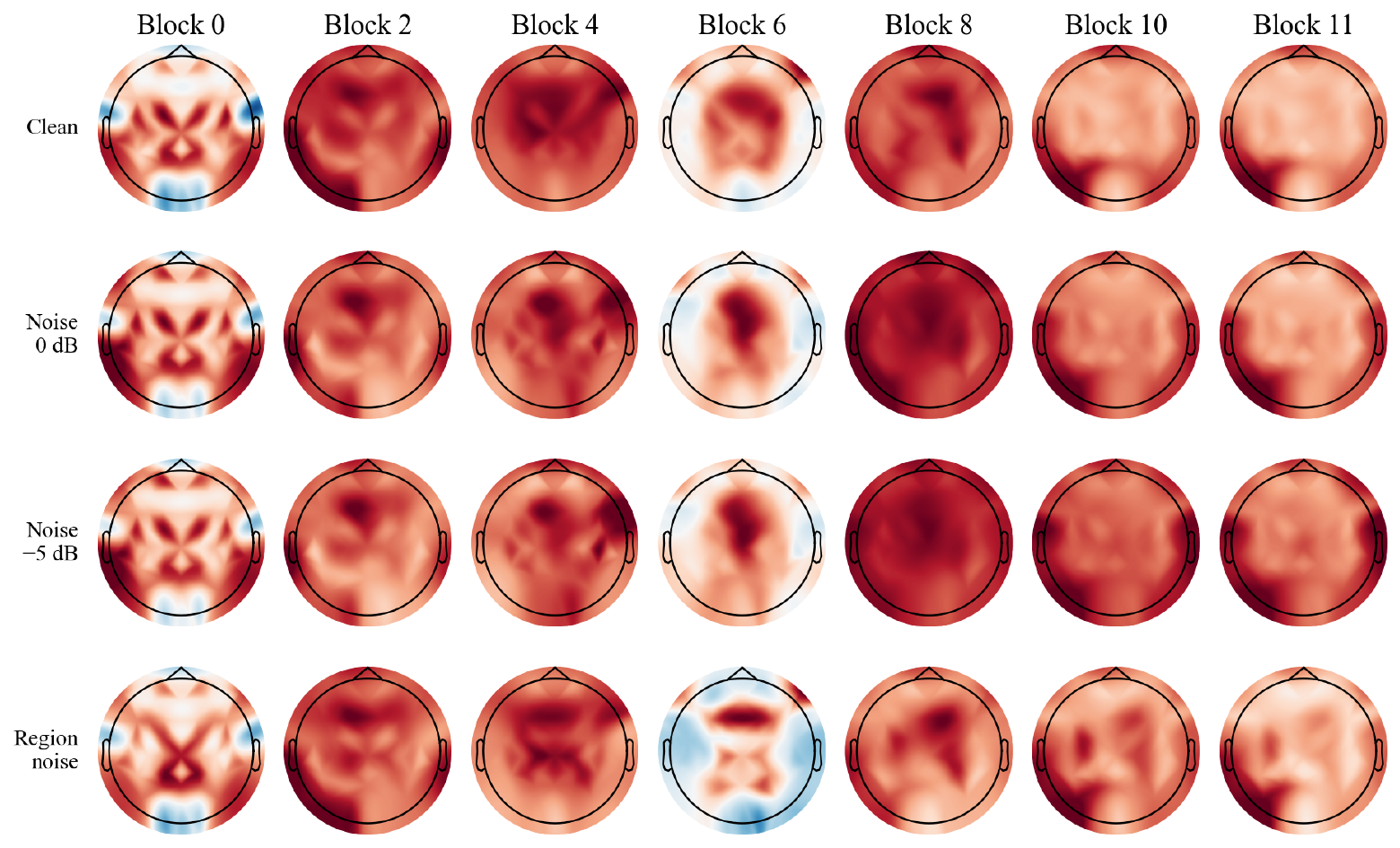}
\caption{Motor-Imagery-KU}
\label{fig:attn_perturb_neurorqv_imagery}
\end{subfigure}
\caption{NeuroRVQ per-block attention under perturbation across datasets. Blocks ordered from input (left) to output (right) \textit{- Plot 1/2}.}
\end{figure}

\begin{figure}[!h]
\centering
\begin{subfigure}[b]{\textwidth}
\centering
\includegraphics[width=\textwidth]{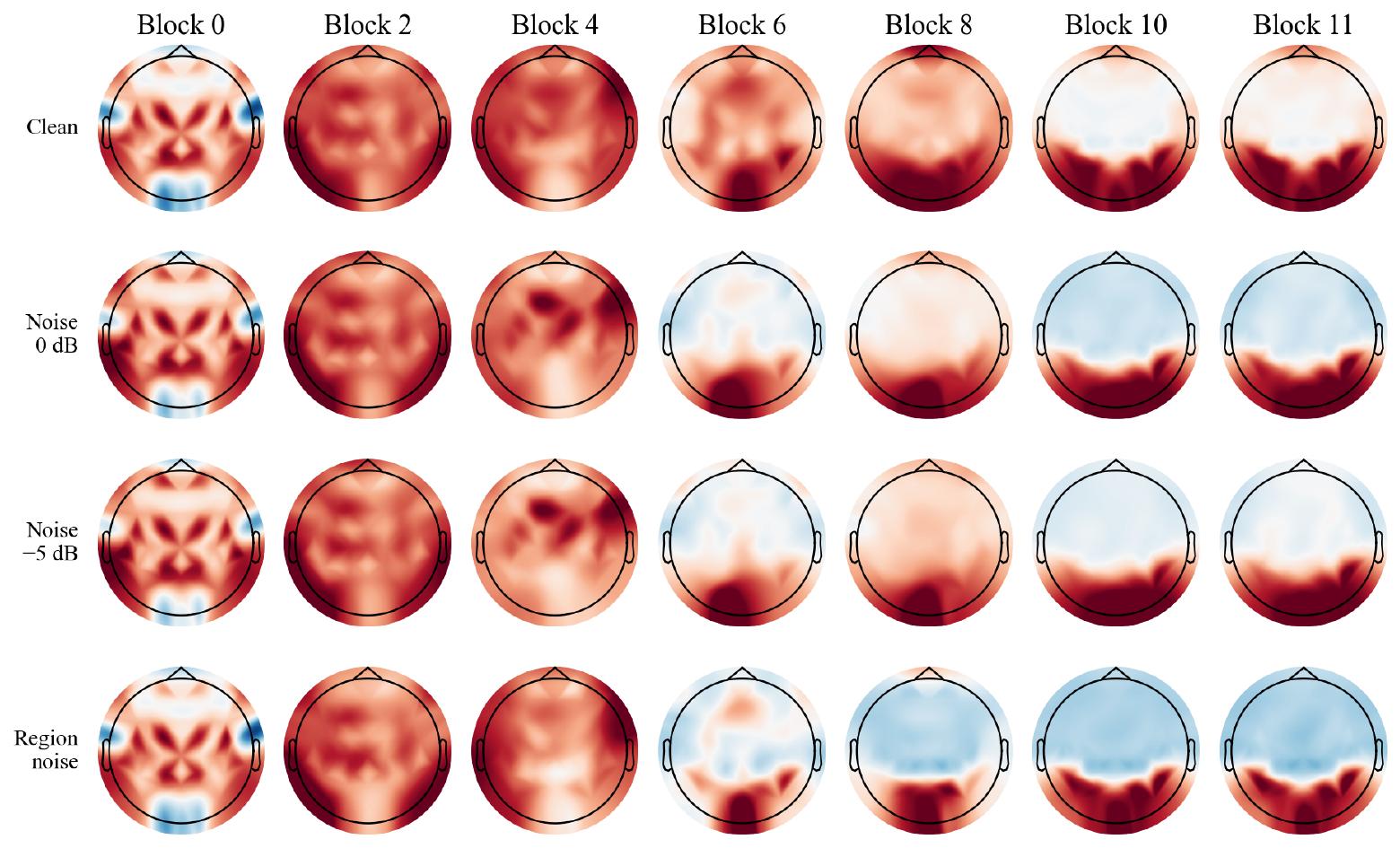}
\caption{ERP}
\label{fig:attn_perturb_neurorqv_erp}
\end{subfigure}
\vspace{0.5em}
\begin{subfigure}[b]{\textwidth}
\centering
\includegraphics[width=\textwidth]{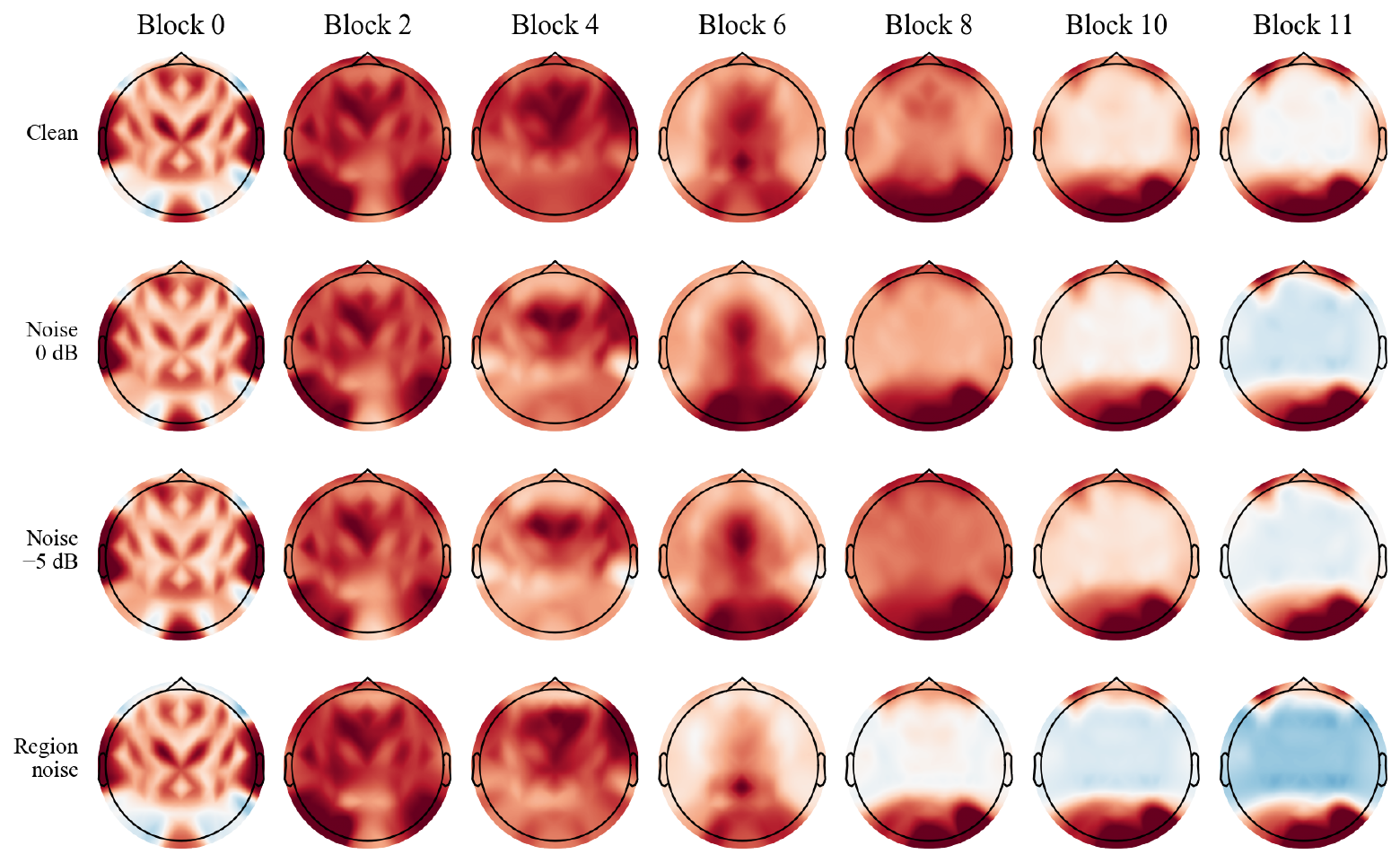}
\caption{Eyes}
\label{fig:attn_perturb_neurorqv_eyes}
\end{subfigure}
\caption{NeuroRVQ per-block attention under perturbation across datasets. Blocks ordered from input (left) to output (right) \textit{- Plot 2/2}.}
\end{figure}

Figures~\ref{fig:attn_perturb_reve_movement}--\ref{fig:attn_perturb_reve_eyes} and Figures~\ref{fig:attn_perturb_neurorqv_movement}--\ref{fig:attn_perturb_neurorqv_eyes} show per-block attention topographic maps for REVE and NeuroRVQ under the same perturbation conditions used in Section~\ref{sec:lrp_perturbation}, excluding noise substitution. Each column corresponds to one transformer block, ordered from input (left) to output (right), and each row to one perturbation condition. REVE's attention patterns shift progressively across blocks under noise, with attention near-uniform at $-5$\,dB. NeuroRVQ's per-block attention is more stable under noise. The spatial patterns at $0$\,dB closely resemble clean attention across block six specifically. Later blocks also remain stable  across most tasks. Region noise leaves both models close to the clean condition.

\newpage

\newpage

\newpage

\newpage

\newpage
\FloatBarrier
\subsection{Block-wise Probing}
\label{app:probing}

To locate where task-relevant information emerges, we train a single layer probe at the output of every transformer block. A single linear layer maps from the block's representation to the number of classes. Probes are trained for up to 20 epochs with Adam, cosine annealing, and early stopping (patience 5 epochs). Probes are trained on the same ten folds as the model checkpoints, and we report the mean balanced accuracy $\pm$ std across folds for each transformer block. We index blocks from $0$ to $L-1$: index $0$ refers to the first block output. Relative depth is $i/(L-1)$, with $L=12$ for NeuroRVQ and $L=22$ for REVE. We use two pooling strategies:
\emph{(i)} \textbf{mean pooling}: averages all tokens to a single vector \emph{(ii)} \textbf{flatten pooling}: concatenates all tokens into a single vector, applies RMSNorm, and fits the linear probe on the full flattened dimension.

\subsubsection{Flatten Probing}
\label{app:probing_concat}

\begin{figure}[h]
\centering
\includegraphics[width=\textwidth]{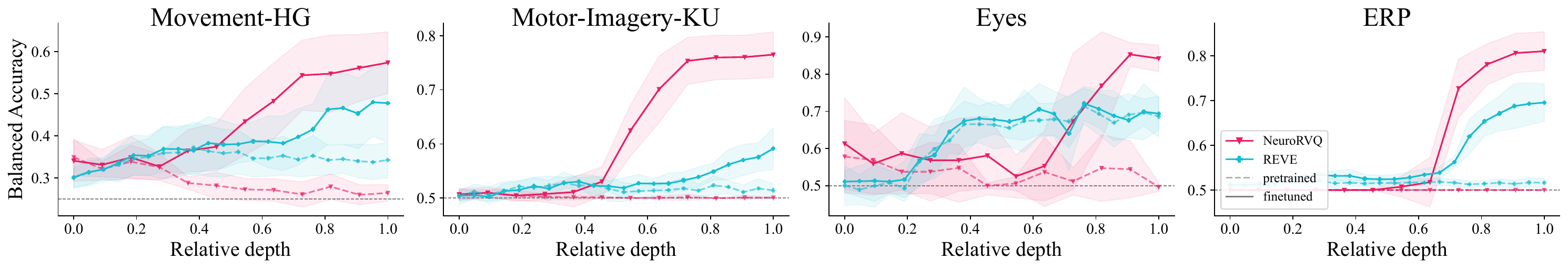}
\caption{Linear probing balanced accuracy (mean pooling) by relative block depth. Pre-trained (dashed) and fine-tuned (solid) curves shown for NeuroRVQ and REVE.}
\label{fig:probing_mean_appendix}
\end{figure}

\begin{figure}[h]
\centering
\includegraphics[width=\textwidth]{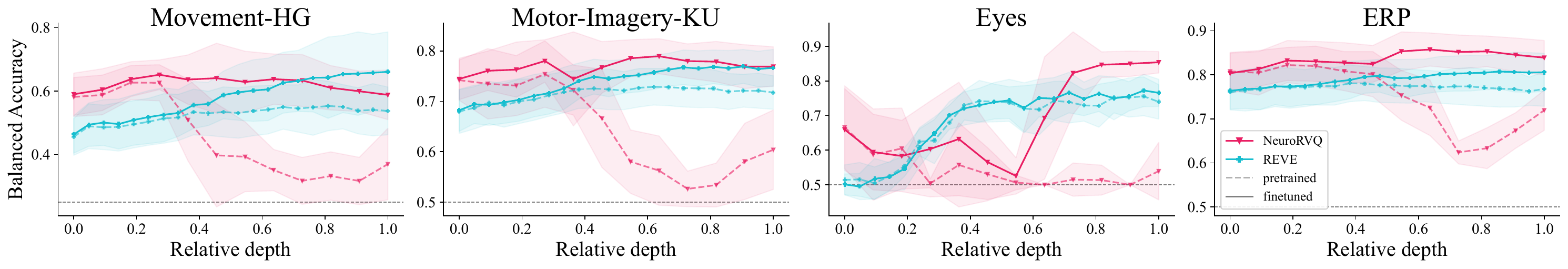}
\caption{Linear probing balanced accuracy (flatten pooling) by relative block depth. Pre-trained (dashed) and fine-tuned (solid) curves shown for NeuroRVQ and REVE. Flatten pooling reveals pre-trained task information that mean pooling misses (Figure~\ref{fig:probing}).}
\label{fig:probing_appendix}
\end{figure}

Section~\ref{sec:probing} reports linear probing with mean pooling. Figure~\ref{fig:probing_mean_appendix} shows the pre-trained and fine-tuned mean-pooled curves overlaid. Figure~\ref{fig:probing_appendix} shows the same analysis with flatten pooling, which preserves positional structure by concatenating all token representations before the probe.

Flatten pooling reveals task information that mean pooling misses in NeuroRVQ's pre-trained representations. Comparing each pooling method's best block, accuracy rises from $0.500$ to $0.822$ on ERP, from $0.509$ to $0.754$ on Motor-Imagery-KU, and from $0.349$ to $0.627$ on Movement-HG. REVE also benefits from flatten pooling. Fine-tuning reduces the mean--flatten gap for NeuroRVQ, but does not close it for REVE. At the final block, flatten pooling still leads by $18.3$, $17.5$ and $11.0$ percentage points on Movement-HG, Motor-Imagery-KU and ERP.

NeuroRVQ's pre-trained flatten probes peak at index $2$ on Movement-HG and ERP (the third block), index $3$ on Motor-Imagery-KU (the fourth block), and index $0$ on Eyes (the first block). Accuracy then declines towards the output. REVE's peaks occur later, at indices $17$, $13$, $8$ and $20$ on Movement-HG, Motor-Imagery-KU, ERP and Eyes. Across these four tasks, the mean relative decline from each task's peak to its final block is $23.1$\% for NeuroRVQ and $2.0$\% for REVE. Fine-tuning reduces the decline for NeuroRVQ but does not remove it on every task: its Movement-HG flatten probe peaks at $0.651$ and ends at $0.588$.

One possible explanation is that later pre-trained blocks specialise for the pre-training objective over downstream task features, as observed in NLP transformers~\citep{voita2019bottomup}. However, the probing results alone do not establish this mechanism. Probes for the pre-training task, subject identity and recording session could test which information persists at each depth and how fine-tuning changes it~\citep{merchant2020happens, saha2020variabilityreview}.

\subsection{Depth Truncation}
\label{app:exit_block}

\begin{figure}[h]
\centering
\includegraphics[width=\textwidth]{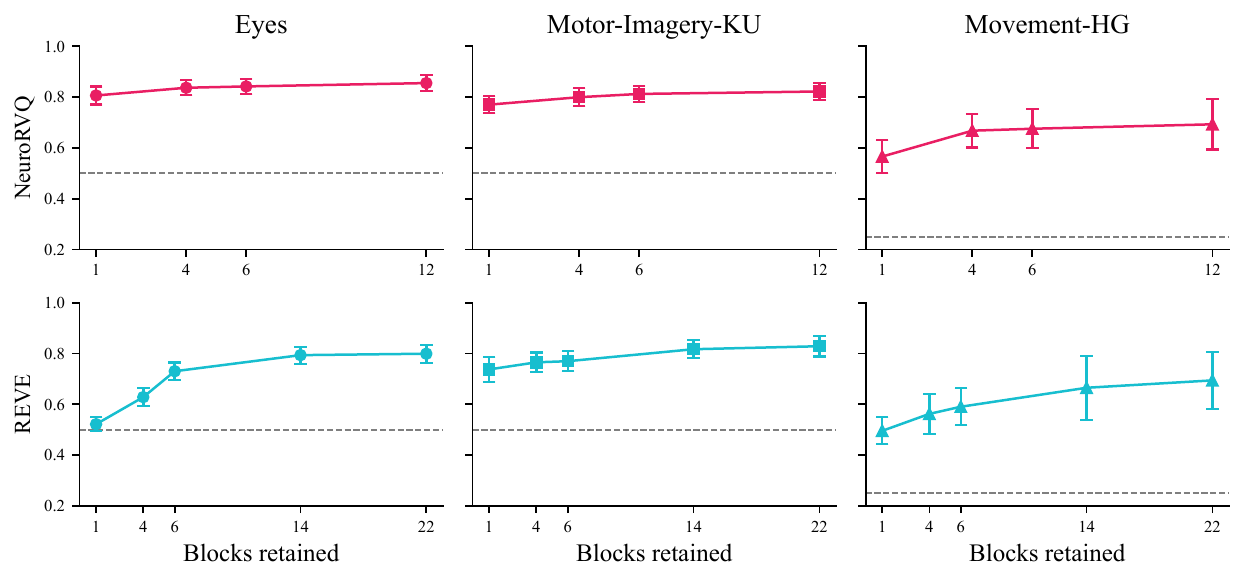}
\caption{Depth truncation, where $k$ counts the transformer blocks retained. For each $k$, we discard later blocks, attach a flatten-pooling head, and fine-tune end-to-end. Balanced accuracy $\pm$ std is shown as a function of $k$. NeuroRVQ approaches full-depth performance by block 6, and REVE by block 14.}
\label{fig:exit_block}
\end{figure}

The probing results show that task-relevant representations are already present in early to middle blocks (Appendix \ref{app:probing_concat}). To test whether downstream performance reflects this, we run a depth truncation ablation. We attach a flatten-pooling classification head (see Appendix~\ref{app:pooling_strategy}) after each of several intermediate blocks and fine-tune the resulting truncated stack. Figure~\ref{fig:exit_block} shows the results. NeuroRVQ approaches full-depth performance by block 6, and REVE by block 14. One hypothesis is that the later layers contribute little to these downstream tasks and instead serve the pre-training objective, though this ablation does not establish this on its own.

\section{Pooling Strategy}
\label{app:pooling_strategy}

We compare four readouts for LaBraM, NeuroRVQ, REVE, BrainOmni and CBraMod on Movement-HG, Motor-Imagery-KU, ERP and Eyes. Each readout is tested under head-only adaptation and full fine-tuning on the same ten folds. 

Mean pooling averages all spatial and temporal tokens. Attention pooling uses a learned query to weigh the tokens before summing them. Flatten pooling concatenates all tokens. Mean-tile repeats the token mean to match the flattened input size and uses the same head as flatten. It therefore matches the head's parameter count without retaining individual tokens. Table~\ref{tab:model_architectures_pooling} reports these counts. For some models, the heads used in this ablation differ from the original heads used in the rest of this study.

\begin{table}[!h]
\caption{Head-only trainable parameters for the four readouts on Eyes (binary, 64 channels). Flatten and mean-tile have the same head structure and parameter count.}
\label{tab:model_architectures_pooling}
\centering
\small
\begin{tabular}{lrrrrr}
\toprule
Pooling & REVE & NeuroRVQ & LaBraM & BrainOmni & CBraMod \\
\midrule
Mean & 2{,}050 & 2{,}401 & 201 & 1{,}026 & 201 \\
Attention & 2{,}562 & 3{,}201 & 401 & 1{,}538 & 401 \\
\addlinespace
Mean-tile & 393{,}218 & 614{,}401 & 153{,}601 & 393{,}218 & 51{,}201 \\
Flatten & 393{,}218 & 614{,}401 & 153{,}601 & 393{,}218 & 51{,}201 \\
\bottomrule
\end{tabular}
\end{table}

Table~\ref{tab:pooling_ablation} reports test balanced accuracy. Under head-only adaptation, flatten pooling gives the highest average for all five models and exceeds mean pooling significantly in each case. Mean-tile improves several models but underperforms compared to flatten, so head size alone does not explain the gain. Attention pooling also underperforms flatten. Training the attention readout during pre-training might help, but this remains a hypothesis.

Under full fine-tuning, the mean-to-flatten gain is smaller for most models. NeuroRVQ changes little ($0.799$ to $0.802$), while REVE improves from $0.762$ to $0.787$ and LaBraM from $0.765$ to $0.786$. The effect remains larger for BrainOmni ($0.638$ to $0.736$) and CBraMod ($0.615$ to $0.759$).

\begin{table}[h]
\centering
\caption{Pooling ablation, for head-only and full fine-tuning. Test balanced accuracy (mean over 10 folds). Within each regime, models are ordered by their best Average. The best per column is \textbf{bold}, second best \textit{underlined}. $^{\ast}$ marks an Average significantly better than Mean pooling for that model (Wilcoxon, Holm-corrected).}
\label{tab:pooling_ablation}
\scriptsize
\setlength{\tabcolsep}{3pt}
\begin{tabular*}{\textwidth}{@{\extracolsep{\fill}}lllccccc}
\toprule
Setting & Model & Pooling & Movement-HG & Motor-Imagery-KU & ERP & Eyes & Average \\
\midrule
\multirow{20}{*}{Head-Only} & NeuroRVQ & Mean & $0.271$ & $0.501$ & $0.542$ & $0.736$ & $0.512$ \\
 &  & Mean-tile & $0.301$ & $0.500$ & $0.516$ & $0.691$ & $0.502$ \\
 &  & Attn & $0.278$ & $0.500$ & $0.535$ & $0.704$ & $0.504$ \\
 &  & Flatten & {\boldmath $0.564$} & \underline{$0.739$} & {\boldmath $0.770$} & {\boldmath $0.827$} & {\boldmath $0.725$}$^{\ast}$ \\
\addlinespace
 & REVE & Mean & $0.309$ & $0.508$ & $0.509$ & $0.743$ & $0.517$ \\
 &  & Mean-tile & $0.359$ & $0.528$ & $0.516$ & $0.783$ & $0.546$$^{\ast}$ \\
 &  & Attn & $0.295$ & $0.520$ & $0.504$ & $0.643$ & $0.491$ \\
 &  & Flatten & \underline{$0.544$} & {\boldmath $0.761$} & \underline{$0.749$} & $0.790$ & \underline{$0.711$}$^{\ast}$ \\
\addlinespace
 & BrainOmni & Mean & $0.322$ & $0.515$ & $0.500$ & $0.651$ & $0.497$ \\
 &  & Mean-tile & $0.371$ & $0.555$ & $0.500$ & $0.754$ & $0.545$$^{\ast}$ \\
 &  & Attn & $0.319$ & $0.515$ & $0.500$ & $0.627$ & $0.490$ \\
 &  & Flatten & $0.533$ & $0.707$ & $0.636$ & $0.809$ & $0.672$$^{\ast}$ \\
\addlinespace
 & LaBraM & Mean & $0.300$ & $0.504$ & $0.500$ & $0.717$ & $0.505$ \\
 &  & Mean-tile & $0.335$ & $0.515$ & $0.500$ & $0.750$ & $0.525$$^{\ast}$ \\
 &  & Attn & $0.296$ & $0.504$ & $0.500$ & $0.599$ & $0.475$ \\
 &  & Flatten & $0.501$ & $0.716$ & $0.568$ & $0.813$ & $0.649$$^{\ast}$ \\
\addlinespace
 & CBraMod & Mean & $0.276$ & $0.505$ & $0.500$ & $0.579$ & $0.465$ \\
 &  & Mean-tile & $0.324$ & $0.509$ & $0.500$ & $0.785$ & $0.530$$^{\ast}$ \\
 &  & Attn & $0.272$ & $0.501$ & $0.500$ & $0.609$ & $0.471$ \\
 &  & Flatten & $0.387$ & $0.653$ & $0.500$ & \underline{$0.820$} & $0.590$$^{\ast}$ \\
\midrule
\multirow{20}{*}{Fine-tuned} & NeuroRVQ & Mean & $0.689$ & $0.809$ & $0.838$ & \underline{$0.861$} & \underline{$0.799$} \\
 &  & Mean-tile & $0.520$ & $0.797$ & \underline{$0.854$} & $0.808$ & $0.745$ \\
 &  & Attn & $0.621$ & $0.805$ & {\boldmath $0.856$} & \underline{$0.861$} & $0.786$ \\
 &  & Flatten & \underline{$0.693$} & \underline{$0.822$} & $0.836$ & $0.855$ & {\boldmath $0.802$} \\
\addlinespace
 & REVE & Mean & $0.626$ & $0.799$ & $0.823$ & $0.799$ & $0.762$ \\
 &  & Mean-tile & $0.608$ & $0.799$ & $0.825$ & $0.796$ & $0.757$ \\
 &  & Attn & $0.626$ & $0.796$ & $0.817$ & $0.780$ & $0.755$ \\
 &  & Flatten & {\boldmath $0.694$} & {\boldmath $0.829$} & $0.826$ & $0.799$ & $0.787$$^{\ast}$ \\
\addlinespace
 & LaBraM & Mean & $0.620$ & $0.775$ & $0.832$ & $0.833$ & $0.765$ \\
 &  & Mean-tile & $0.582$ & $0.744$ & $0.825$ & $0.771$ & $0.730$ \\
 &  & Attn & $0.597$ & $0.780$ & $0.816$ & $0.770$ & $0.741$ \\
 &  & Flatten & {\boldmath $0.694$} & $0.805$ & $0.831$ & $0.814$ & $0.786$$^{\ast}$ \\
\addlinespace
 & CBraMod & Mean & $0.381$ & $0.544$ & $0.757$ & $0.776$ & $0.615$ \\
 &  & Mean-tile & $0.405$ & $0.576$ & $0.779$ & $0.792$ & $0.638$$^{\ast}$ \\
 &  & Attn & $0.402$ & $0.629$ & $0.762$ & $0.817$ & $0.652$$^{\ast}$ \\
 &  & Flatten & $0.613$ & $0.786$ & $0.797$ & $0.840$ & $0.759$$^{\ast}$ \\
\addlinespace
 & BrainOmni & Mean & $0.430$ & $0.572$ & $0.704$ & $0.846$ & $0.638$ \\
 &  & Mean-tile & $0.567$ & $0.545$ & $0.719$ & $0.850$ & $0.670$ \\
 &  & Attn & $0.438$ & $0.544$ & $0.684$ & $0.845$ & $0.628$ \\
 &  & Flatten & $0.607$ & $0.749$ & $0.726$ & {\boldmath $0.864$} & $0.736$$^{\ast}$ \\
\bottomrule
\end{tabular*}
\end{table}

\newpage

\end{document}